%% file: main.tex
\documentclass[runningheads]{llncs}

\usepackage{eccv} 

\usepackage{eccvabbrv}
\usepackage{wrapfig}
\usepackage{multirow}
\usepackage{soul}
\usepackage{graphicx}
\usepackage{booktabs}
\usepackage{hyperref}

\usepackage[accsupp]{axessibility}  

\input{preamble}

\usepackage{hyperref}

\usepackage{orcidlink}

\newcommand\blfootnote[1]{%
  \begingroup
  \renewcommand\thefootnote{}\footnote{#1}%
  \addtocounter{footnote}{-1}%
  \endgroup
}

\begin{document}

\title{Robust Onion: Peeling Open Vocab Object Detectors Under Noise } 

\titlerunning{Robust Onion}

\author{Priyank Pathak*, Mukilan Karuppasamy*$^\dagger$, Aaditya Baranwal, \\
Shruti Vyas, and Yogesh S Rawat}

\authorrunning{Pathak et al.}

\institute{UCF Institute of Artificial Intelligence, University of Central Florida (UCF)\\
\texttt{\{priyank,aaditya.baranwal,shruti,yogesh\}@ucf.edu, mukilan.nitt@gmail.com}\\[1ex]
\textbf{Project Page}: \url{https://ucf-crcv.github.io/RobustOnion/}\\ 
\textbf{Github}: \url{https://github.com/ppriyank/RobustOnion}
\blfootnote{*Equal contribution \hspace{10pt} $^\dagger$ work done as intern}
}
\maketitle

\input{sec/abstract}

\input{sec/intro}

\input{sec/related}

\input{sec/setup}

\input{sec/model}

\input{sec/deeper}

\input{sec/dataset}

\input{sec/lang}

\input{sec/predicted}

\input{sec/conclusion}

\input{sec/acknowledgements}

\bibliographystyle{splncs04}
\bibliography{main}

\input{Supplementary/all_files}

\end{document}

%% file: preamble.tex
\usepackage[dvipsnames]{xcolor}
\usepackage{colortbl}

\usepackage{makecell}
\usepackage{multirow}
\usepackage{booktabs}
\usepackage{hhline}

\definecolor{RedOrange}{RGB}{255,69,0}
\definecolor{mygreen}{rgb}{0.0, 0.5, 0.0}
\definecolor{mygray}{gray}{.9}
\definecolor{mypink}{RGB}{255, 105, 180}
\definecolor{blue2}{HTML}{0066CC}
\definecolor{orange2}{HTML}{CC6600}
\definecolor{mygray}{gray}{.9}
\definecolor{Gray}{gray}{0.8}
\definecolor{LGray}{gray}{0.9}
\definecolor{LGray2}{gray}{0.92}
\definecolor{ao(english)}{rgb}{0.0, 0.6, 0.0}

\sethlcolor{LGray2}

\newcommand{\bluetext}[1]{{\color{blue2} #1}}
\newcommand{\orangetext}[1]{{\color{RedOrange} #1}}
\newcommand{\orangetexttwo}[1]{{\color{orange2} #1}}
\newcommand{\pinktext}[1]{{\color{magenta} #1}}

\newcommand{\figno}[1]{{\color{blue}#1}}

\newcommand{\insight}[1]{\hl{#1}}
\newcommand{\SUPP}{\textit{Supplementary}}

\newcommand{\takeaway}[0]{{\textcolor{RubineRed}{\textbf{[Takeaways \& Model Design] }}}}

\newcommand{\SV}[1]{}
\newcommand{\PP}[1]{}

\newcommand{\TIM}[1]
{{\cellcolor{ao(english)!10}#1}}
\newcommand{\SIM}[1]
{{\cellcolor{ao(english)!25}#1}}
\newcommand{\BIM}[1]
{{\cellcolor{ao(english)!40}#1}}

\newcommand{\TD}[1]
{{\cellcolor{red!10}#1}}
\newcommand{\SD}[1]
{{\cellcolor{red!20}#1}}
\newcommand{\BD}[1]
{{\cellcolor{red!35}#1}}


%% file: sec/abstract.tex
\vspace{-8pt}
\begin{figure}[!h]
\centering
\includegraphics[width=\linewidth]{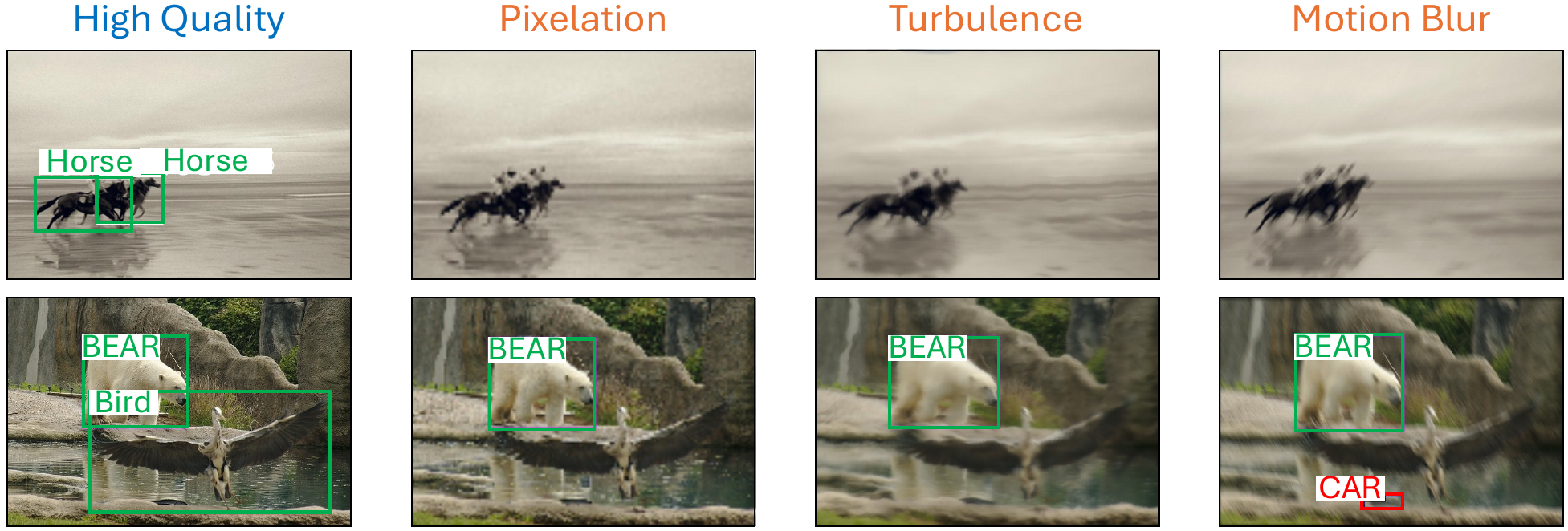}
  \caption{
  \textbf{Effect of Noise:} GLIP\cite{li2022grounded} (above) \& MM-GDINO\cite{zhao2024open} (bottom) performance on  COCO~\cite{lin2015microsoft} for noises like turbulence, pixelation, 
  and motion blur.
}
\vspace{-12pt}
\label{fig:motivate}
\end{figure}

\begin{abstract}
The impact of real-world noise on \textbf{O}pen \textbf{V}ocabulary \textbf{O}bject \textbf{D}etectors (OV-ODs) remains poorly understood due to their architectural complexity.
We present our comprehensive analysis, \textbf{Robust Onion}, an empirical study that uses controlled
synthetic visual degradations to \textit{peel} OV-ODs layer-by-layer, revealing \textbf{how, why}, and \textbf{where} robustness degrades, systematically analyzing \textit{feature collapse}.
Our findings reveal that models with similar vision backbones exhibit comparable robustness, driven by similar feature collapse at similar layers, while factors such as pretraining strategy, architectural nuances, and caption supervision contribute little. Robustness is primarily governed by the image domain rather than annotations, explaining the similar robustness impact on COCO and LVIS, and why datasets like ODinW-13 can give an impression of inflated robustness due to large, isolated objects.
Finally, we validate our insights by \textit{improving robustness} on real-world BDD-100K, WiderFace, and VisDRONE via our lightweight plug-and-play \textbf{NN \& TK0} approach, using 
96$\times$ fewer trainable parameters than end-to-end training.
We also \textbf{explain the prior works'} robustness observations.
\keywords{Explainability \and Interpretability \and Object Detectors \and Noises 
}
\end{abstract}

%% file: sec/intro.tex
\section{Introduction}

Vision Language Models (VLMs) have shown strong generalization in tasks like image-to-text retrieval~\cite{saha2024improvedzeroshotclassificationadapting, Pathak_2025_ICCV}, open-vocabulary classification~\cite{abdelhamed2025seeenhancingzeroshotimage}, image captioning~\cite{cheng2025caparenabenchmarkinganalyzingdetailed}, visual-question-answering~\cite{huynh2025visualquestionansweringearly} \etc. The ability to adapt without fine-tuning makes VLMs highly lucrative for applications where \textit{zero-shot} is not just a convenience but a necessity. VLM based
\textbf{O}pen \textbf{V}ocabulary \textbf{O}bject \textbf{D}etectors (\textbf{OV-ODs}) have gained attention for their easy adaptability in security~\cite{he2024instruct}, medical imaging~\cite{yu2025umit}, environmental monitoring~\cite{xue2024reo}, and self-driving cars~\cite{Tian2024DriveVLMTC}.

\begin{figure}[!t]
\centering
\begin{subfigure}[t]{0.34\textwidth}
\centering
\includegraphics[height=2.6cm]{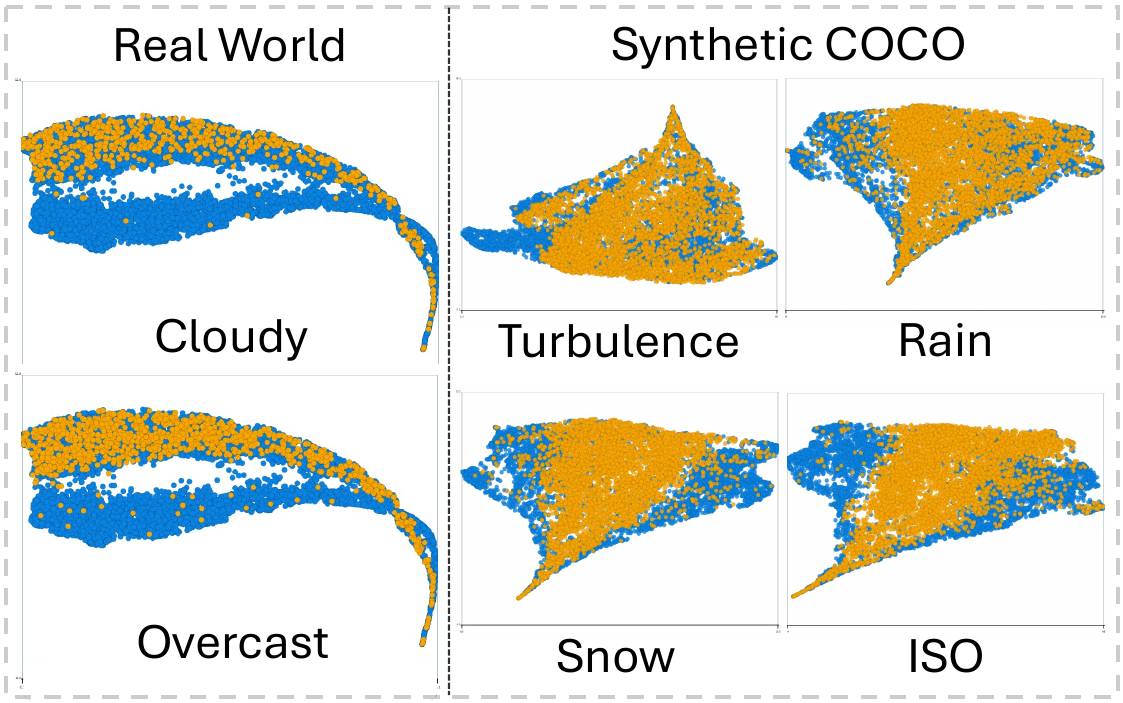}
\caption{\textbf{Observable} Collapse}
\label{fig:observable}
\end{subfigure}
\hfill
\begin{subfigure}[t]{0.34\textwidth}
\centering
\includegraphics[height=2.6cm]{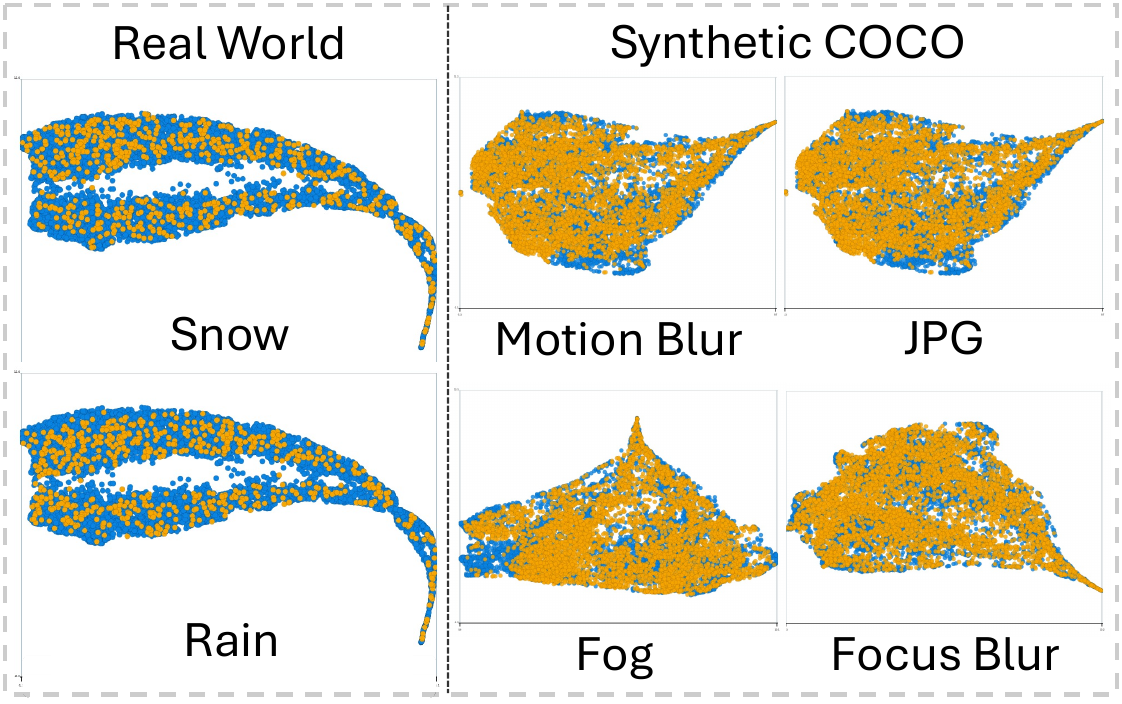}
\caption{\textbf{Minimum} collapse}
\label{fig:minimal}
\end{subfigure}
\hfill
\begin{subfigure}[t]{0.24\textwidth}
\centering
\includegraphics[height=2.6cm]{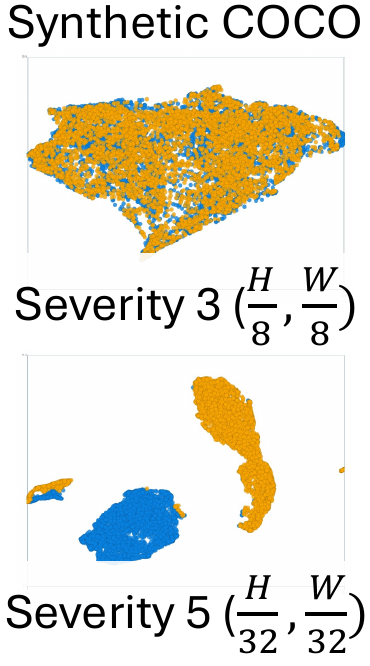}
\caption{Pixelation Severity}
\label{fig:pix_sev}
\end{subfigure}
\hfill

\caption{ 
\textbf{Real-World \& Synthetic:}
GLIP-T synthetic COCO  
\orangetexttwo{noisy features} collapse against \bluetext{clean image} \textit{aligns} with BDD-100K real-world collapse. 
BDD-100K doesnt have explicit clean images, hence all noisy categories are in \bluetext{blue}, except the \orangetexttwo{highlighted}.
}
\label{fig:noise_motivate}
\end{figure}

Real-world deployment of OV-ODs requires a critical understanding of their robustness against visual distortions and noise.
However, their enormous complexity from various moving parts like vision backbone, text backbones, fusion network, box predictors, alignment networks \textit{etc.}, makes them one of the most obscure deep learning models. 
Additionally, studying and isolating the true impact of low quality (LQ) noise requires matching high quality (HQ) counterpart images; and such \textit{well-annotated `noise-paired' real-world datasets} are nearly impossible to find. 
As a result, despite their growing use, the effect of out-of-distribution noise on VLM based OV-ODs remains largely underexplored.

Addressing the gap in robustness against real-world distortions (\eg BDD-100K~\cite{bdd100k}), our novel analysis,
\textbf{Robust Onion} broadly categorizes noise-induced \orangetexttwo{\textbf{feature (variance) collapse}}~\cite{ling2023dive, chai2023recognizability, pathak2025lrfm, bao2025mint} 
into two categories:
\textbf{1. Observable} (\figno{\cref{fig:observable}}): \orangetexttwo{noisy features} form distinct clusters separate from \bluetext{clean HQ features} (\eg  cloudy, overcast),
\textbf{2. Minimal} (\figno{\cref{fig:minimal}}): little or no observable collapse (\eg snow, rain).
By carefully tuning synthetic 
noises, we mimic these feature collapses, providing a \textit{practical proxy} for real-world 
degradations (\eg 
turbulence for observable, motion blur for minimal). 
Increasing noise severity can change the collapse from minimal to observable (\figno{\cref{fig:pix_sev}}).
Robust Onion then empirically \textit{peels} each component of OV-OD under these controlled noises.

To answer: \textit{``How do visual distortions impact complex models such as OV-ODs?"} and \textit{``What are the most effective directions to improve their robustness?"},
We present an interpretability-driven analysis of robustness in OV-ODs. 
Rather than simply evaluating performance, our goal is to explain why failures occur, where they originate in the architecture, and how robustness can be improved.

Our analysis revolves around five key questions:
\textbf{(1)} Do bells and whistles like pretraining, finetuning, and architecture modules impact the robustness? 
\textbf{(2)} Which part of the model architecture (Backbone, FPN, RPN, etc) induces the robustness?
\textbf{(3)} Are larger models inherently more robust, or are other factors decisive?
\textbf{(4)} Is robustness solely determined by the model, or do input images play a role? 
\textbf{(5)} Does language prompting improve robustness under visual noise?

Our analysis frames robustness as an explainable phenomenon rather than a black box outcome. 
We peel the layers of OV-ODs 
to reveal the following:

\begin{itemize}
    \item \textbf{Vision Backbone drives robustness:} Similar backbones exhibit comparable robustness due to resembling feature collapses at \textit{similar depths},
    regardless of bells and whistles of pretraining or overall architecture.

    \item \textbf{Layer-wise robustness:} Shallow layers are adversely affected by the noise. 
    Further, \textit{cross-exchanging the backbone layer} can improve robustness.
    
    \item \textbf{Dataset bias:} ODinW-13 can give a false impression of robustness because it features large and isolated objects.
    
    \item \textbf{Domain over annotation:} Detecting \textit{on} what (domain) matters more than detecting \textit{what} (annotation), thus similar robustness of COCO \& LVIS.

    \item \textbf{Minimal language influence:} Once the visual features are degraded, language/captions contribute little to recover lost robustness.

\end{itemize}

\noindent Each analysis includes \takeaway highlighting key actionable insights for designing \textbf{robust real-world OV-OD}.
We conclude by validating our analysis on 
real-world autonomous driving datasets like BDD-100K, DAWN, 
Foggy Cityscapes, and Virtual KITTI 2, WiderFace, and VisDRONE via our \textbf{NN \& TK0} approach, a lightweight technique to induce robustness in object detectors (open vocabulary). 
\textbf{NN \& TK0} uses 96$\times$ fewer trainable parameters than end-to-end training with similar robustness. 
Our analysis also explains several of the previously published robustness-related observations.


%% file: sec/related.tex
\section{Related Work}
\noindent \textbf{VLMs and OV-ODs:}
VLM generalizability~\cite{radford2021learning, minderer2022s} has progressed into object detection~\cite{shen2024groundvlp, gu2021open, zareian2021open}. 
Advances in large-scale multimodal training, have further enhanced open-vocabulary object detectors (OV-OD)~\cite{alayrac2022flamingo, tsimpoukelli2021multimodal, chen2022visualgpt, minderer2023scaling, zhao2024taming}. 
Versatility of OV-ODs~\cite{li2025zeroshot, deng2024zeroshotgeneralizableincrementallearning} makes them ideal for real-world (\eg surveillance~\cite{davila2023mevid}, industrial-inspection~\cite{synspill_2025}, satellite imagery~\cite{patil2017classification},
autonomous-driving~\cite{Tian2024DriveVLMTC}, \textit{etc}; thus understanding their limitations~\cite{bianchi2024devil, zhang2024visionlanguagemodelsvisiontasks, bao2025mint} against noise is crucial.

\vspace{3pt}
\noindent \textbf{Robustness against  Noise:} Weather (\eg rain, fog, snow) \& common artifacts (\eg jpg compression) pose significant challenges in object detection~\cite{mao2023coco, qin2022denet, chhipa2024openvocabularyobjectdetectorsrobustness, zhang2024enhancing, yoo2024and}. 
These distortions lose discriminative features,  fundamentally affecting almost all models~\cite{Shermeyer_2019_CVPR_Workshops}.
Robustness methods often use synthetic noises to improve on real-world tasks, like person re-id~\cite{pathak2025coarse}, self-driving in fog/rain~\cite{gupta10483822}.
Despite the prevalence of noises, their impact on VLMs remains largely unexplored~\cite{cheng2019low, 8600370}, with most recently LR0.FM~\cite{pathak2025lrfm} analyzing robustness of VLMs in image classification.
On the contrary, OV-ODs are far more complex than simple VLMs. Here, we present a comprehensive analysis of SOTA OV-ODs, revealing bottlenecks and critical factors in their robustness. More works in ~\cref{tab:all_prev_works}.

%% file: sec/setup.tex
\section{Analysis Setup}

\begin{wrapfigure}{r}{0.48\textwidth}
\vspace{-14pt}
\centering
\includegraphics[width=\linewidth]{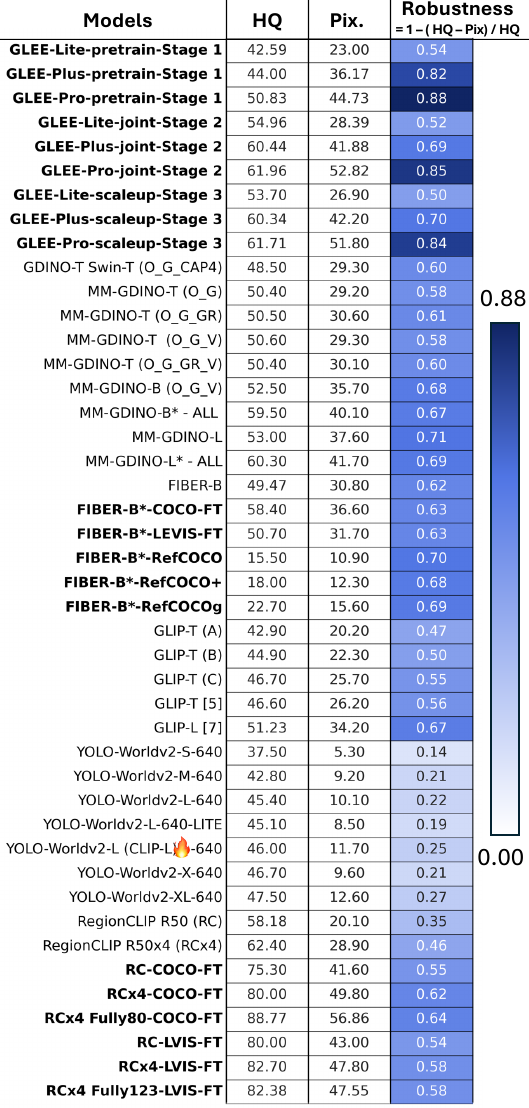}
\caption{
\textbf{Models vs Pixelated COCO (mAP):} Shade $\propto$ robustness. 
Fine-tuned  (COCO, LVIS, RefCOCO) in \textbf{bold}.
%
}
\label{fig:model_ranking}
\vspace{-35pt}
\end{wrapfigure}

\noindent \textbf{Models}: 
We analyze 6 publicly available OV-ODs: RegionCLIP~\cite{zhong2022regionclip} (RC, RCx4), GLIP~\cite{li2022grounded}, FIBER~\cite{fiber2022}, MM-Grounding-DINO~\cite{zhao2024open} (MM-GDINO), GLEE~\cite{Wu_2024_CVPR}, and YOLO-World~\cite{yoo2024and} (YOLO). 
\figno{\cref{fig:model_ranking}} shows robustness of all models. 
CNN-based ones (YOLO \& RegionCLIP) are not as robust as transformer ones (unless fine-tuned), hence, \textbf{we mainly focus of our analysis on transformers}.

\vspace{3pt}
\noindent \textbf{Datasets:}  
Robustness is evaluated on 3 benchmarks: COCO~\cite{lin2015microsoft} (val2017), LVIS~\cite{gupta2019lvis} (miniVal) and ODinW-13~\cite{li2022grounded} (set of 13 datasets). COCO (80 categories) and LVIS (1,203 categories) have \textit{same images, but different annotations}.
Language analysis uses RefCOCO/+/g~\cite{Kazemzadeh2014ReferItGameRT,yu2016modeling,mao2016generation}, and  
Flickr30k~\cite{7410660}. 
Model-related insights are validated on \textbf{real-world} driving datasets: BDD-100K~\cite{bdd100k}, DAWN~\cite{kenk2020dawn},   Foggy Cityscapes~\cite{SDV18, Cordts2016Cityscapes}, Virtual KITTI 2~\cite{cabon2020virtual},  WiderFace~\cite{yang2016wider}, and VisDRONE~\cite{zhu2021detection}.

\begin{figure*}[!t]
\centering    \includegraphics[width=\linewidth]{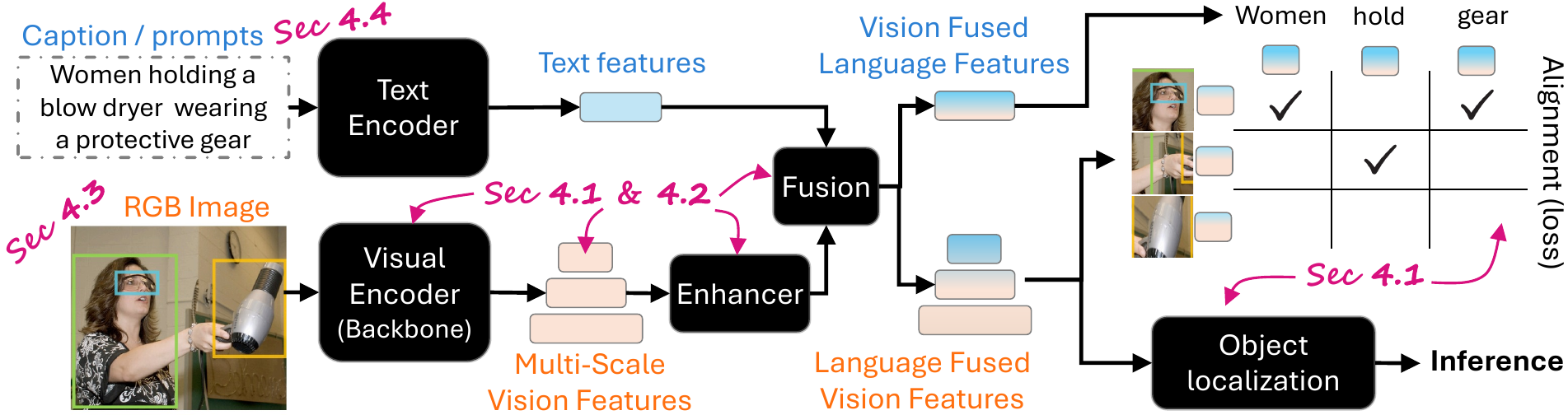}
\caption{\textbf{OV-OD Overview:} Architecture (black), may include additional components and losses. 
Fusion of 
\bluetext{text features} with \orangetexttwo{multi-scale vision features} via self-attention, cross-exchanges text-vision modality information.
The role of each component in robustness against noise is described in \pinktext{listed sections}.
The vision feature enhancer (neck) is commonly referred as FPN / pixel decoder. 
Image modified from GLIP~\cite{li2022grounded}. 
}
\label{fig:General_framework}
\end{figure*}

\vspace{3pt}
\noindent \textbf{Framework:}  
\figno{\cref{fig:General_framework}} shows the general framework of OV-ODs. 
Image and captions are processed through a pyramidal multi-scale vision encoder (\eg ResNet, Swin Transformer) and a text encoder 
(\eg  BERT, CLIP),
respectively. Vision features are enhanced via FPN (or pixel-decoder) followed by cross-self-attention fusing text and vision embedding. These fused features predict bounding boxes, confidence scores, and class labels.

\vspace{3pt}
\noindent \textbf{Noises:}  
Analyzing noise requires measuring the drop in performance \textit{relative} to clean features \ie analysis of LQ-HQ image pairs (absent in real-world). 
Instead, we sample one controlled synthetic noise from each feature collapse category (\figno{\cref{fig:noise_motivate}}): Turbulence (\textit{observable} collapse), and Motion Blur (\textit{minimal} collapse). 
We also use Pixelation (\eg compression, distant objects) for severity (intensity) analysis. 
Pixelation is simulated~\cite{pathak2025lrfm} by downsampling the image and upscaling it back via bicubic interpolation\footnote{
Severity `s':
$(H,\:W) \rightarrow (\frac{H}{2^s},\frac{W}{2^s}) \rightarrow (H,\:W)$.
Pixelated images $\ne$ low resolution (\eg $(\frac{H}{2^3}, \frac{W}{2^3})$ can still be high resolution $(256,256)$).
}.
Turbulence (\eg hot air) is simulated via pre-trained GAN~\cite{Mao_2021_ICCV}. 
Motion Blur~\cite{10483822} simulates motion (\eg videos). 
These noises are only applied to the input image, leaving \textit{textual captions unchanged}.

\begin{figure*}[!t]
\centering
\includegraphics[width=\textwidth]{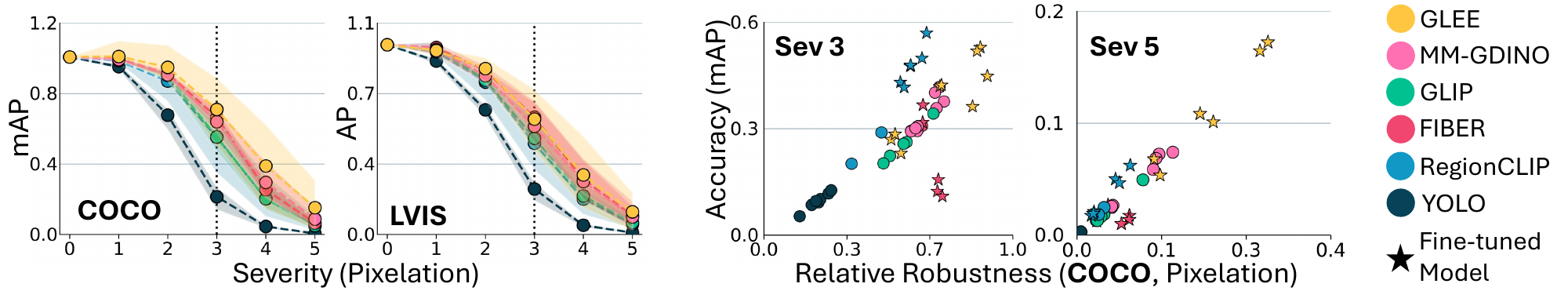}
\caption{ \figno{\textit{(Left)}} 
\textbf{Performance decline w/Severity:}
Models start dropping performance around severity 3. 
Shade shows all model variants, with solid indicating mean accuracy.
\figno{\textit{(Right)}} 
\textbf{Accuracy and Robustness linearity:}
Accuracy (zero-shot) forms an $\sim$ linear relationship with robustness; preserving the relative ranking of models. 
}
\label{fig:acc_sev}
\label{fig:acc_robustness}
\label{fig:sec_41}
\end{figure*}

\vspace{3pt}
\noindent \textbf{Evaluation:}  
Metrics include AP (LVIS), mAP (COCO), AP\textsubscript{avg} (ODinW-13), and Recall@1 (Flickr30K).
We shall use relative robustness (denoted as `robustness')~\cite{chen2024robustsam, schiappa2024robustness} as the key metric for measuring a model's robustness against noise.  
Relative Robustness = $1\:-\:$(Drop in Accuracy/Accuracy) = $1 - (\textit{Acc}_{Clean} - \textit{Acc}_{Noise})/\textit{Acc}_{Clean}$.
Here, $\textit{Acc}_{Clean}$ and $\textit{Acc}_{Noise}$ denote accuracy on original and noisy images. 
Relative Robustness is independent of absolute performance, enabling cross-model, and cross-dataset comparisons. 
Higher severity (4, 5) risks random predictions ($Acc_{Noise} \simeq 0$, \figno{\cref{fig:acc_sev} (left)})
We also observe a linear relationship between absolute accuracy and relative robustness (\figno{\cref{fig:acc_robustness} (right)}), with outliers being mostly fine-tuned models, \eg RegionCLIP ({\textcolor{cyan}{$\bigstar$}}, accuracy $\uparrow$ \& robustness $\uparrow$), FIBER-B ({\textcolor{magenta}{$\bigstar$}} accuracy $\downarrow$ \& constant robustness). 


\section{Analysis and Explainability
(Where, Why, What)}
In our analysis visuals, y-axis will represent relative robustness (unless indicated otherwise); insights in \insight{highlight}; 
fine-tuned model as $\bigstar$;  \takeaway describes practical robust OV-OD design using insights.

%% file: sec/model.tex
\subsection{WHERE: Model-Based Analysis}
\label{sec:model_based_analysis}

\begin{figure}[!t]
\centering
\begin{subfigure}[t]{0.35\textwidth}
\centering
\includegraphics[width=0.92\linewidth]{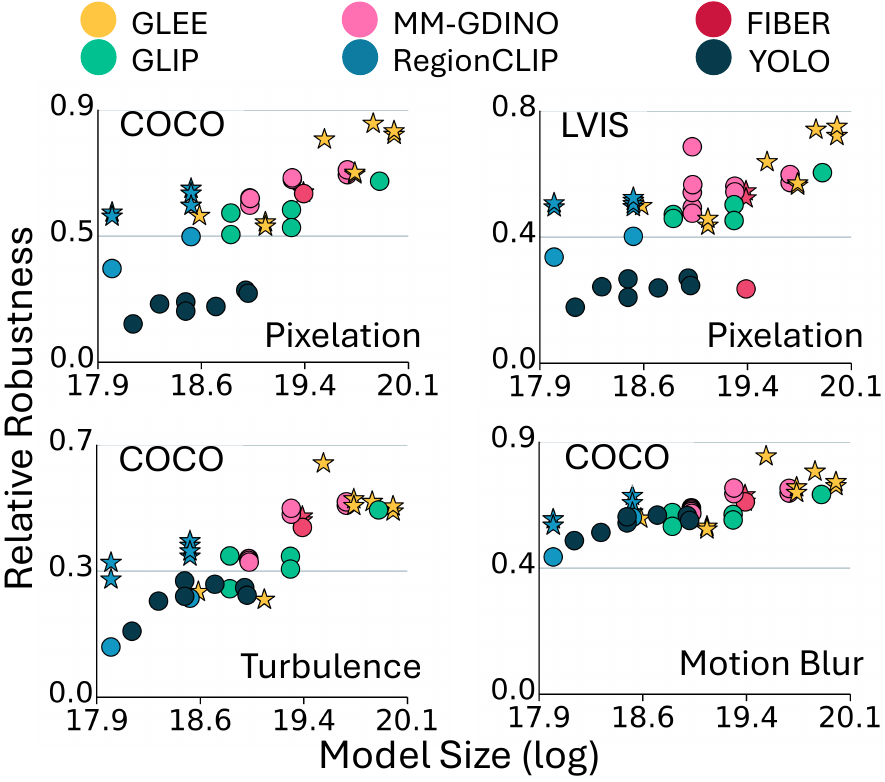}
\end{subfigure}
\hfill
\begin{subfigure}[t]{0.62\textwidth}
\centering
\includegraphics[width=1\linewidth]{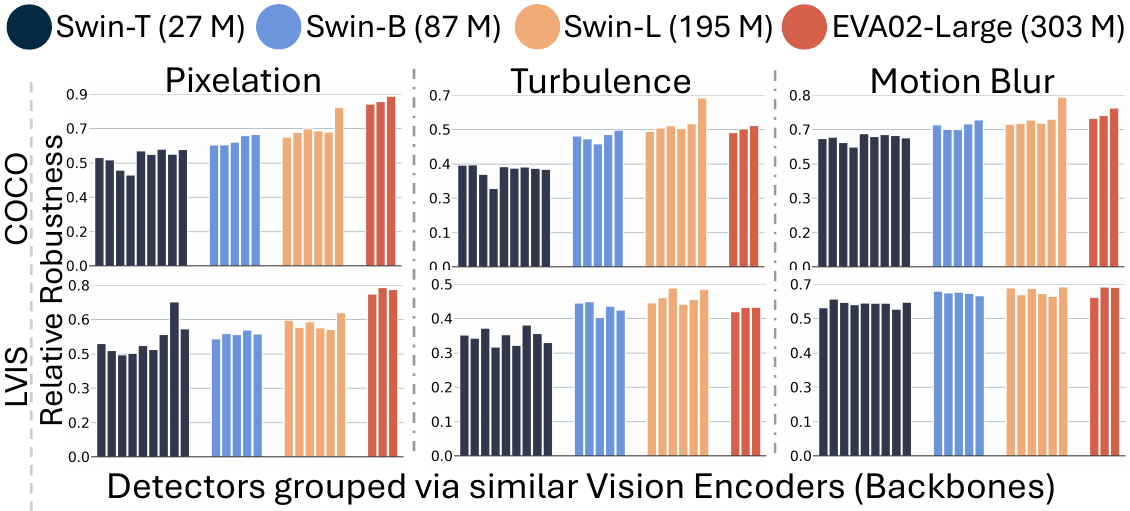}
\end{subfigure}

\caption{
\figno{\textit{(Left)}} 
\textbf{Robustness vs Size:}
Larger models are more robust (+ve  correlation with size);
GLEE transformers $>$ GLEE ResNet robustness.
\figno{\textit{(Right)}} 
\textbf{Similar vision backbones:}
Performance remains relatively consistent across models with similar backbones and depths. 
EVA-02 (303M) $\simeq$ Swin-L (195M) robustness, both 24-blocks.
}
\label{fig:model_size}
\label{fig:backbones}
\end{figure}

\figno{\cref{fig:model_size} left)}
shows a \insight{strong positive correlation between robustness and model size} (entire detector + text backbone) with pearson correlation on COCO / LVIS: 0.68 / 0.66 for pixelation, 
0.78 / 0.72 for turbulence, and 
0.77 / 0.70 for motion blur\footnote{Pearson correlation $<0.3$ is none / weak and moderate for $[0.3,0.7]$}.  
Large Transformer detectors consistently outperform CNNs ($<19.4$ size). 

\figno{\cref{fig:backbones} right)} groups detectors on similar vision backbones (irrespective of other architectural components like enhancers, fusion, decoders, text backbones \etc), revealing:
1) Models with \insight{\textbf{similar backbones show similar robustness}} despite different architectures ($\frac{3}{23}$ exceptions).
2) Larger backbones are almost always robust (\eg EVA-02). However, 
in general (turbulence \& motion blur), deeper transformers are slightly more robust than shallow ones, while \insight{size \textbf{not} playing a crucial role}, \ie ResNet $<$ Swin-T (12 blocks, \textit{27M}) $<$ Swin-B (12 blocks, \textit{87M}) $\le$ Swin-L (24 blocks, \textit{195M})  $\simeq$ EVA-02 (24 blocks, \textit{303M}), where $\simeq$ indicates similar robustness.
Similar trend observed on other noises like Pixel drop, ISO, Salt-pepper, JPG compression, and Fog (\textit{Supplementary}).

\begin{figure}[!t]
\centering
\begin{subfigure}[t]{0.49\textwidth}
\centering
\includegraphics[width=\linewidth]{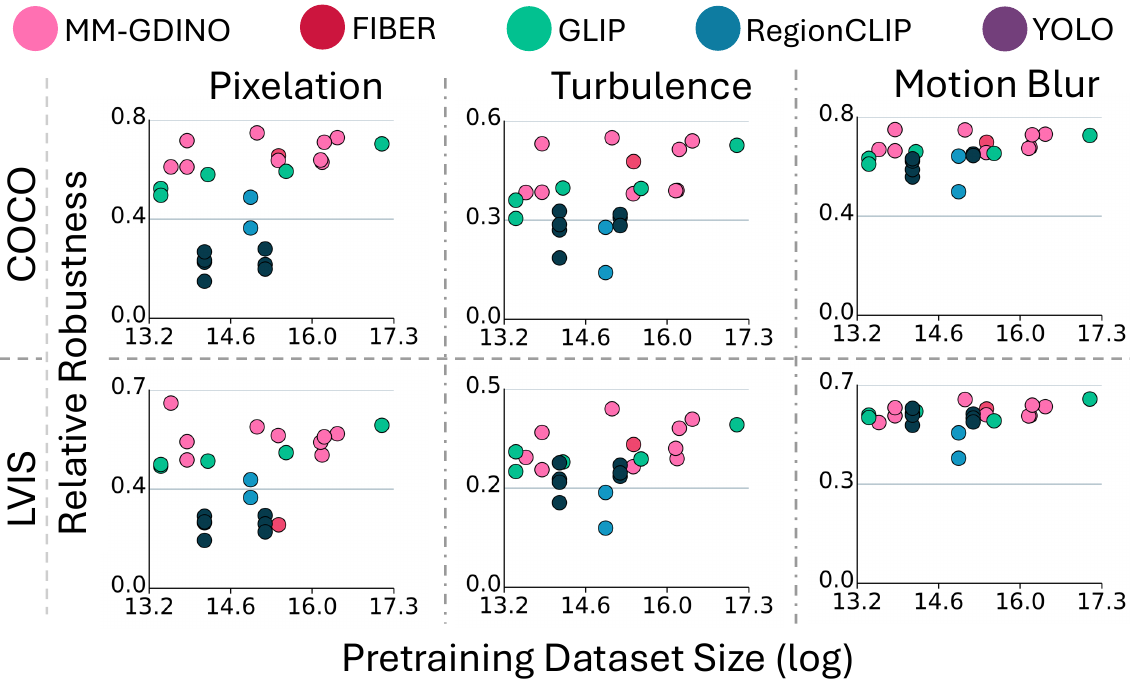}
\end{subfigure}
\hfill
\begin{subfigure}[t]
{0.48\textwidth}
\centering
\includegraphics[width=\linewidth]{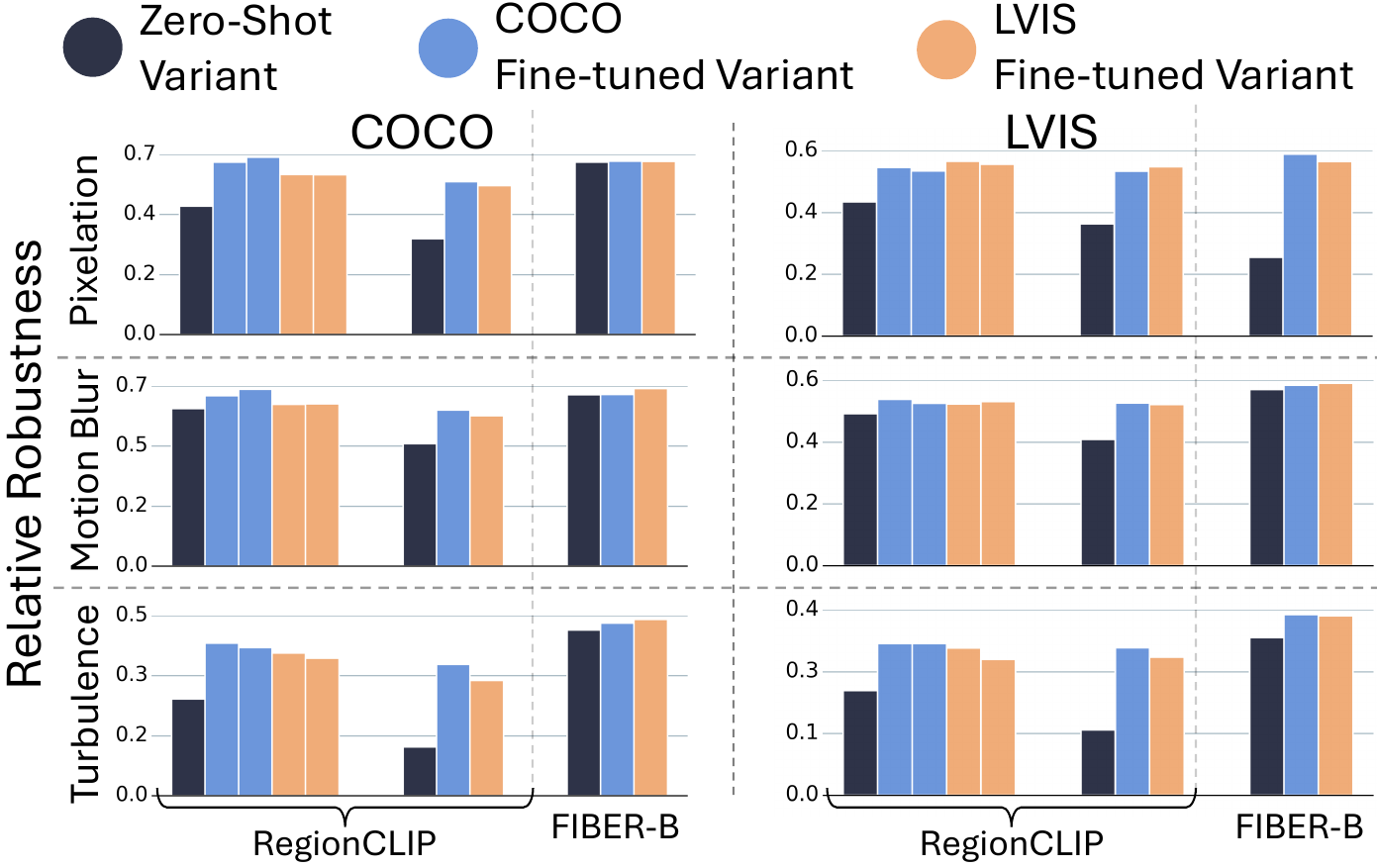}
\end{subfigure}
\caption{
\figno{\textit{(Left)}} 
\textbf{Zero-shot Pretraining}:
GLIP (green) consistent  robustness regardless of pretraining size (no clear correlation).
\figno{\textit{(Right)}}  
\textbf{Fine-tuning Impact}: 
Identical impact of COCO and LVIS finetuning on COCO with synthetic noise evaluation. 
}
\label{fig:fine_tune_analysis}
\label{fig:pretraining}
\end{figure}

\figno{\cref{fig:pretraining} left)} 
shows \insight{pre-training dataset size weakly affects the zero-shot robustness}; indirectly meaning different pre-training datasets (\& their count).
For example, GLIP shows steady robustness across different pretraining sizes (green dots). 
Pearson's correlation on COCO / LVIS: 0.35 / 0.23 for pixelation, 0.41 / 0.39 for turbulence, and 0.42 / 0.22 for motion blur.

\figno{\cref{fig:fine_tune_analysis} right)} 
shows clean COCO and LVIS fine-tuning improves (noisy COCO / LVIS) RegionCLIP (ResNet) robustness over zero-shot, while FIBER-B (transformer) is minimally impacted. 
Overall, \insight{fine-tuning is \textbf{not} a universal robustness boost}~\cite{chen2024robustsam}.
Interestingly, COCO and LVIS (same images, different annotations) have a similar impact on robustness, \ie \insight{impact of fine-tuning is mostly governed by \textbf{domain of images}, with annotation playing a minimal role}. 
This importance of image domain (over annotation) will re-emerge in \figno{\cref{sec:dataset_robustness}}.

\vspace{3pt}
\noindent \takeaway 
We observe that robustness is driven primarily by backbone architecture and depth. 
Large models like Swin-L and EVA-02-L are more robust than smaller ResNets, even under identical training and frameworks (GLEE transformer $>$ GLEE ResNet).
Models sharing similar backbones exhibit similar robustness, indicating that backbone choice matters far more than \textit{auxiliary modules} (\eg MM-GDINO/GLEE decoder, neck/FPN network, \etc), and \textit{extensive pre-training} (\eg MM-GDINO pre-trained on 9 datasets, and GLEE on 18 in three stages of pre-training, and GLIP via two stages \etc). 
Explicit noise-robust training is likely to have a stronger impact on robustness than simple pretraining and fine-tuning.
Given resource constraint environments, Swin-B can be an amazing alternative to EVA-02 (4x bigger), and Swin-L (2x bigger), with \textit{size playing a limited role} for sufficiently large backbones.  

%% file: sec/deeper.tex
\subsection{WHY: Insight into Transformers: GLIP, MM-GDINO \& GLEE}
\label{sec:feats_viz}

\begin{figure*}[!t]
\centering
\includegraphics[width=\linewidth]{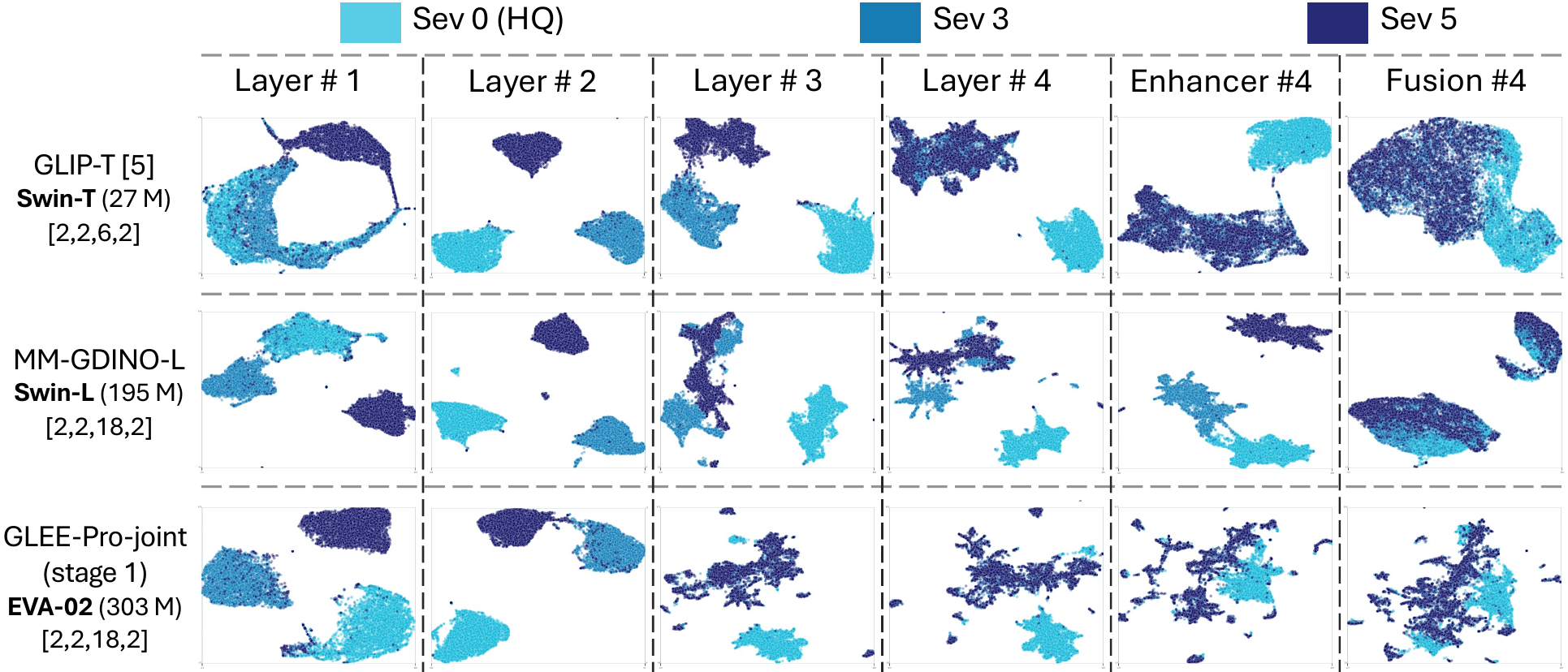}
\caption{
\textbf{Pixelation UMAP:} 
Vision backbone, enhancer, and language fused features shown for sev 5 (dark) and sev 0 (lighter).
n-th layer features represented by `\#n', \eg layer \#4 is 24-th block in Swin-L \& EVA-02, and 12-th in Swin-T (blocks in each layer shown on left).
Feature overlap between sev 5 and sev 0 (similar) implies robustness. 
}
\label{fig:pipeline_flow}
\end{figure*}

\cref{fig:backbones} showed 
consistent robustness for Swin-L in GLIP, MM-GDINO, and GLEE, with a maximum $\Delta$ in robustness of 0.15/0.08 in COCO/LVIS across all noises. 
This narrows the analysis to the vision backbone.
\figno{\cref{fig:pipeline_flow}} illustrates the UMAP~\cite{mcinnes2018umap} of the last 4 layers of the backbone, last layer for the feature enhancer, and fusion network. 
Ideally, \hl{robust models shouldn't distinguish between severities}, \ie sev 5 noisy features identical to sev 0 (HQ), \ie \textit{robustness $\approx$ features overlap}.
\SUPP{} details further on t-SNE plots and other noises.

\vspace{3pt}
\noindent \textbf{1) Backbone (Layers)}:
Deeper layers (\#3,\#4) have a higher features overlap across severities that the shallow ones (\#1,\#2), \ie \insight{shallow layers are more vulnerable to noise}. Additionally, 2\textsuperscript{nd} and 4\textsuperscript{nd}   block (\#1,\#2)  for all models have similar feature overlap (despite architectural differences), \ie \insight{\textbf{same depth} have similar feature collapse}. This helps explain why similar-depth backbones exhibit similar robustness (Swin-L $\simeq$ EVA-02-L, both have 24-block depth).

\vspace{3pt}
\noindent \textbf{2) Enhancer:} 
Enhancers are simple conv block on top of vision backbone (\cref{fig:General_framework}); produces 
 similar feature overlap across severities as that of backbone features (layers \#4).
 This indicates 
 \insight{a minimal contribution to robustness}. 

\vspace{3pt}
\noindent \textbf{3) Fusion:} 
Fusion cross-exchanges multi-scale vision features ($192\times\!192, 96\times\!96, 48\times\!48, 24\times\!24$) 
with language.
Although language does not significantly affect robustness (\cref{sec:languge_useless}), \insight{\textbf{cross exchanging information across all vision layers induces robustness}}, as evidenced by the significant overlap of features.

\vspace{3pt}
\noindent \takeaway 
Robustness in OV-ODs, which typically \textit{consist of 3-5 transformers, can be reliably enhanced by simply focusing on the vision backbone}.
Techniques that enhance robustness, particularly via shallow layer modifications~\cite{pathak2025lrfm}, can be viable, as early layer features are more vulnerable to noise.
Encouraging cross-layer information exchange within the vision backbone, such as self-attention or non-local~\cite{wang2018non} modules, can further enhance robustness.
Although, this insight is derived from synthetic noise, \textbf{we show its real-world impact} in \figno{\cref{sec:future_work}}.
Layers at \textbf{similar depth tend to exhibit similar feature collapse}, which partly explains why similar backbones have similar robustness.

%% file: sec/dataset.tex
\subsection{WHAT: Robustness as a function of Dataset}
\label{sec:dataset_robustness}

\begin{figure}[!t]
\centering
\includegraphics[width=\linewidth]{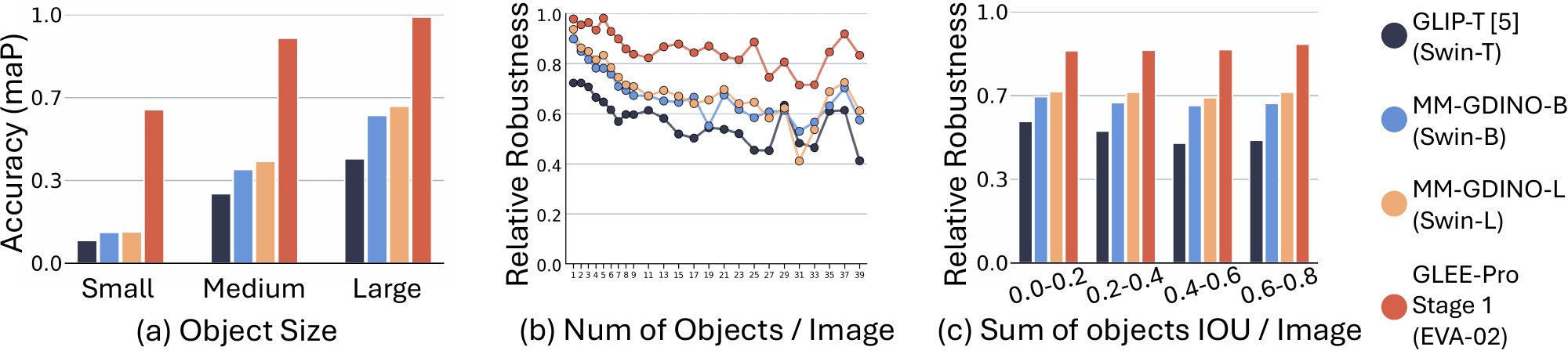}
\caption{ 
\figno{\textit{(a)}}  \textbf{Object size:}
Larger objects are more robust.
\figno{\textit{(b)}} 
\textbf{Object Count:}
Models are very robust when they have to detect $\mathord{<}3$ objects.
Jumping accuracy after $\mathord{>}25$ objects is likely due to very few samples in that range for averaging to smooth out. 
\figno{\textit{(c)}} 
\textbf{Occlusion:}
Robustness pretty much is unaffected with degrees of overlap between objects. 
All experiments on COCO for pixelation. Other noises in \SUPP.
}
\label{fig:occlusion}
\label{fig:no_of_objects}
\label{fig:object_size}
\end{figure}

\figno{\cref{fig:object_size} a)} illustrates \insight{larger ($\ge96\times96$) object detection is more robust to noise than the smaller ones ($\le32\times32$)}. 
\figno{\cref{fig:no_of_objects} b)} illustrates that almost \insight{all \textbf{detectors are highly robust when there is only one object to detect}}. 
As the number of objects/image goes above 3, robustness starts to see a drop, eventually saturating around 10 or more objects/image.
The jitter in robustness after 25 objects likely stems from the small sample size in that bin.
\figno{\cref{fig:occlusion} c)} illustrate \insight{the IOU of overlapping objects  \textit{i.e.} occlusion, has hardly any impact on robustness} (\textit{counter-intuitive}).  
Cluttering (or business) in an image is a function of both the number and  occlusion of objects.
Hence, it's safe to say, despite  occlusion, cluttered image with few objects will still be relatively easier to detect.

\begin{figure}[!t]
\centering
\begin{subfigure}[t]{0.48\textwidth}
\centering
\includegraphics[width=\linewidth]{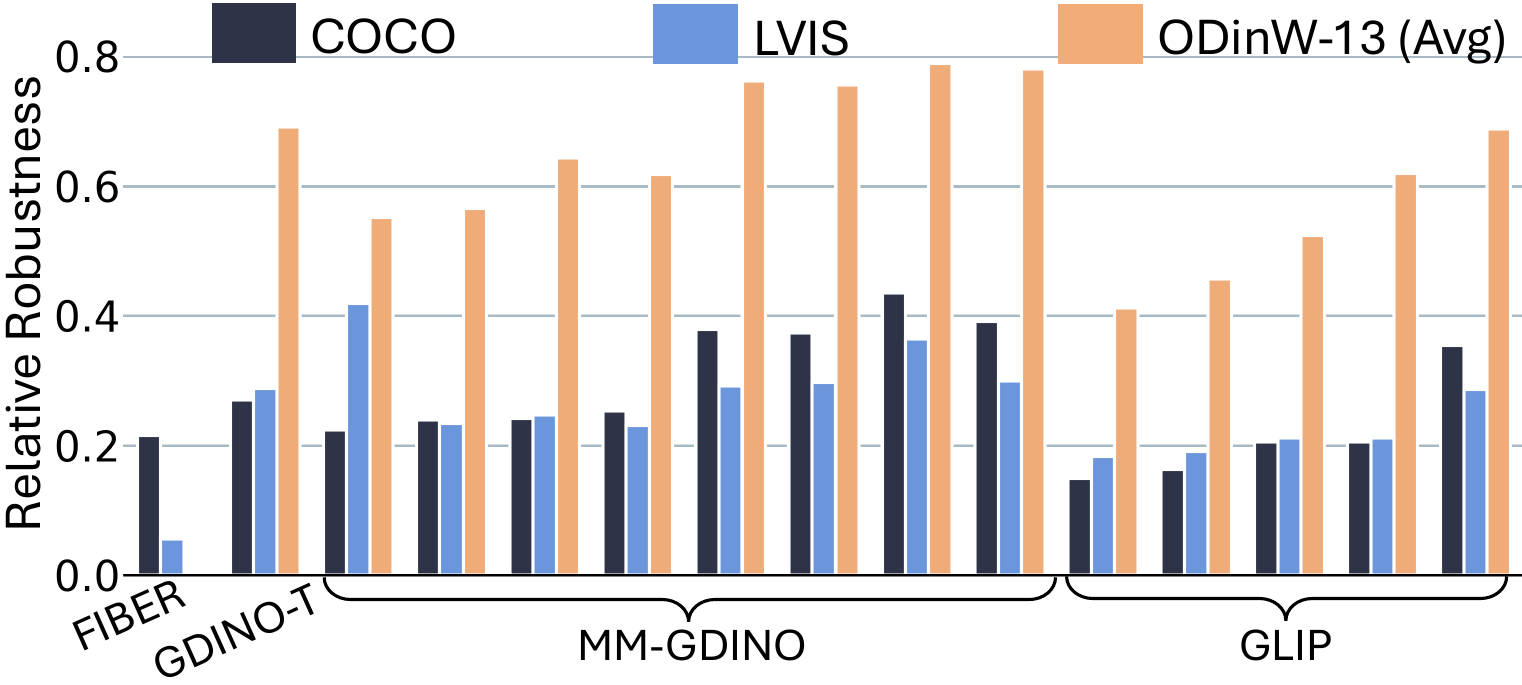}
\label{fig:dataset_eval}
\end{subfigure}
\hfill
\begin{subfigure}[t]{0.48\textwidth}
\centering
\includegraphics[width=\linewidth]{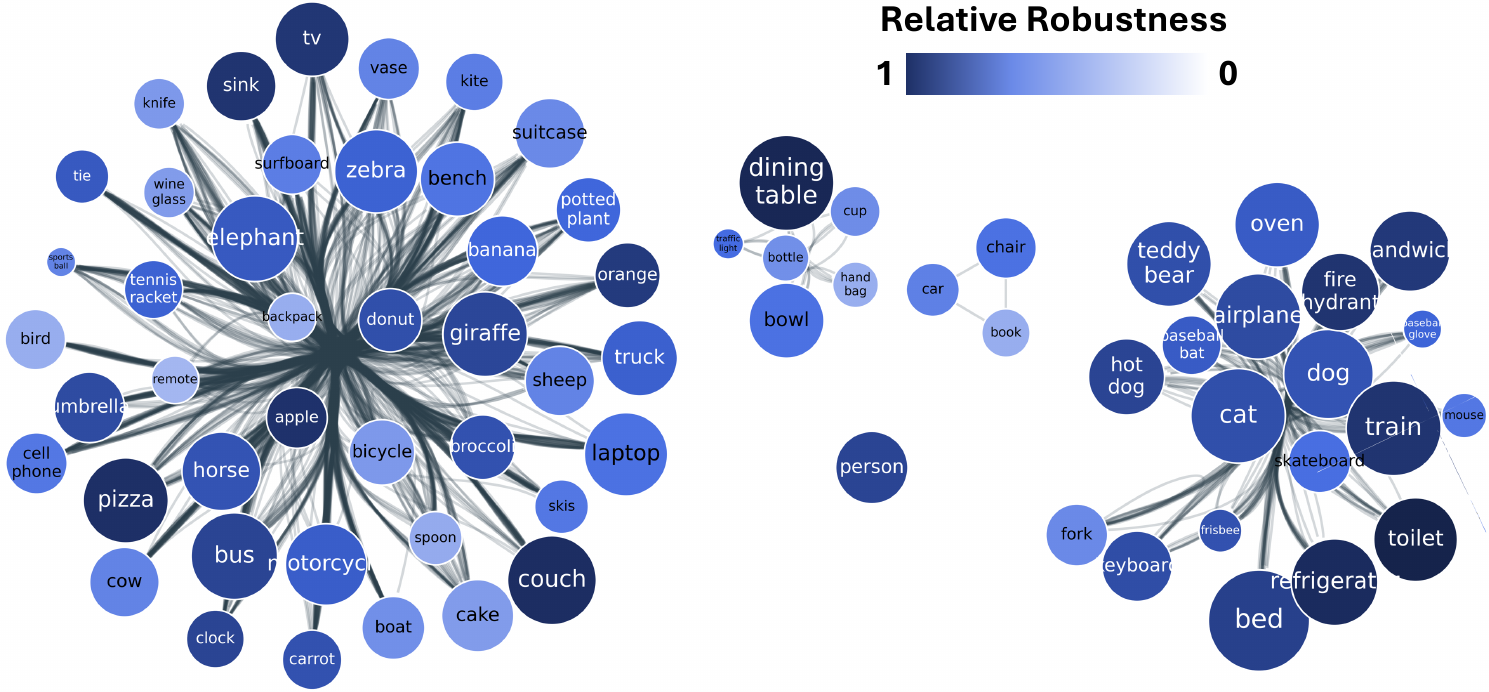}
\label{fig:categoes}
\end{subfigure}
\hfill

\caption{
\figno{\textit{(Left)}} \textbf{Dataset Dependency:} 
ODinW-13 is more immune to noise than COCO \& LVIS (both with comparable robustness) on sev 4 pixelation.
\figno{\textit{(Right)}} \textbf{Class-wise robustness:} 
Certain COCO classes (grouped by bins of log frequency) are more robust (shade of blue) for GLIP-T. Moderate correlation with object size (dot size).
}
\label{fig:dataset_specific}
\end{figure}

\figno{\cref{fig:dataset_specific}} (left) shows ODinW-13 maintains high robustness scores of $\simeq0.6$ across models, nearly twice that of LVIS and COCO (both exhibiting similar robustness). 
This supports \textit{~\cref{fig:fine_tune_analysis} (right)}, \insight{robustness relies more on the image domain (same images for COCO \& LVIS, different for ODinW-13) rather than annotation}.
Differences between ODinW-13 and COCO/LVIS include 
\textit{i) Object Size:} 
Only $\approx 10\%$ of ODinW-13 objects are very small ($\le32\times32$) versus $\simeq42\%$ in COCO and $\simeq58\%$ in LVIS.
\textit{ii) Object Density:} 
ODinW-13 has $\simeq50\%$ images with single object, compared to $\simeq12\%$ for COCO and $\simeq38\%$ for LVIS.

\figno{\cref{fig:dataset_specific} (right)} shows robustness varying by class of object, moderately reflected by average object size (size of dot). For pixelation, robustness correlates with mean class size for COCO / ODinW-13 as 0.52 / 0.45.
On COCO, categories like \textit{parking meter}, \textit{stop sign}, and \textit{toilet} are easiest to detect, while for ODinW-13 classes like \textit{lobster}, \textit{jellyfish}, and \textit{hand} are easiest \textit{(\SUPP)}.
In contrast, robustness shows almost no correlation with class frequency ($\simeq 0.02$ for COCO, $\simeq 0.021$ for ODinW-13), suggesting \insight{\textbf{how often a class appears does not drive robustness}}.
This explains why LVIS's long-tail distribution, adding rare classes to COCO, exhibits similar robustness to COCO.
A possible explanation for the disproportionate robustness of certain objects is that they usually appear alone (\eg single traffic signal), and may be easier to detect.

\vspace{3pt}
\noindent \takeaway 
Models are inherently robust 
for large, singular objects. 
Conversely, \textbf{datasets like ODinW-13 can overstate robustness}, giving false impression of high robustness
than reflecting true measure. 
Robustness seems to largely depend on diverse image domains rather than annotation  (detecting \textit{on} what is more important than \textit{what}, COCO $\simeq$ LVIS). 

%% file: sec/lang.tex
\subsection{WHERE: Expressiveness of Captions and Prompt Engineering}
\label{sec:languge_useless}

\begin{figure}[!t]
\centering
\begin{subfigure}[t]{0.33\textwidth}
\centering
\includegraphics[width=\linewidth]{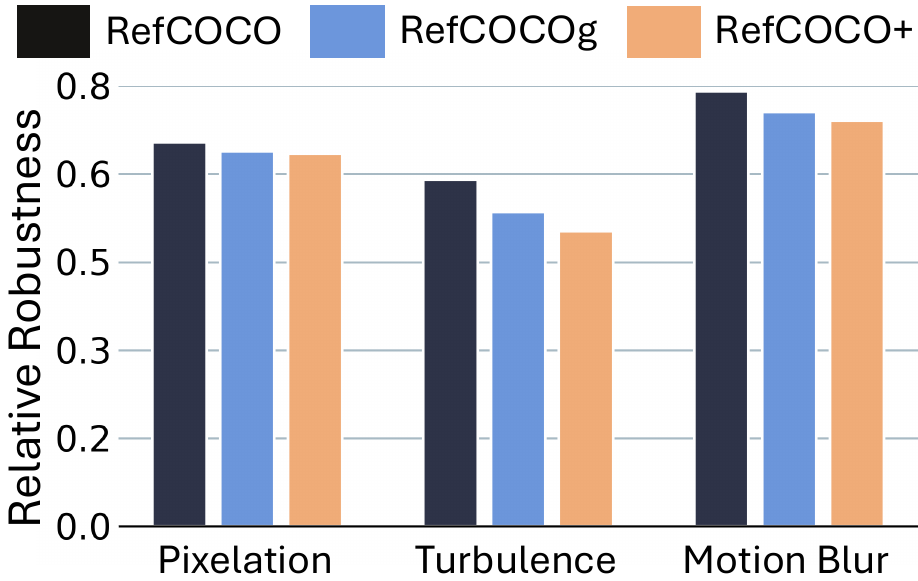}
\end{subfigure}
\hfill
\begin{subfigure}[t]{0.63\textwidth}
\centering
\includegraphics[width=\linewidth]{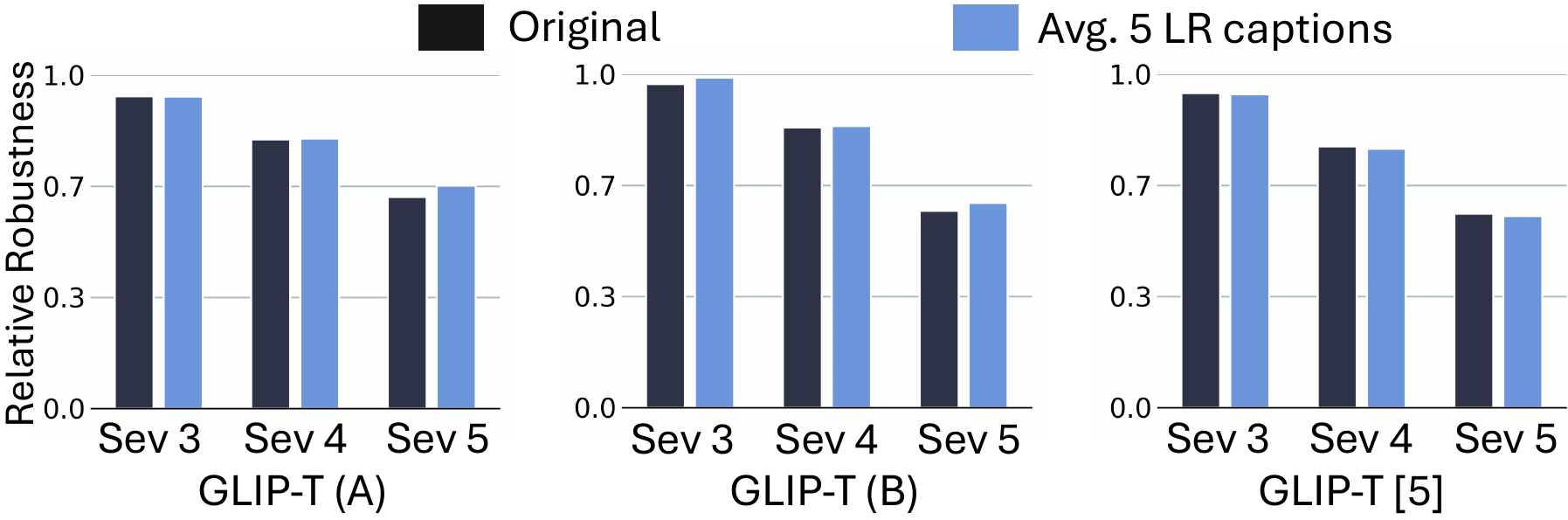}
\end{subfigure}
\caption{
\figno{\textit{(Left)}} 
\textbf{Train Captions:} 
REC fine-tuned FIBER evaluated on COCO.
Despite \textit{training on captions} with different degrees of expressiveness (RefCOCOg is most descriptive), robustness varies slightly, indicating limited impact on robustness. 
\figno{\textit{(Right)}} 
\textbf{Prompt Engineering:} 
\textit{Evaluation} on Flickr30k with \textit{test captions} modified with textual context of pixelation (light) has minimal impact on robustness of GLIP variants.
}
\label{fig:dirty_captions}
\label{fig:language_model}
\label{fig:language_prob}
\end{figure}

\figno{\cref{fig:language_model} left) } shows that fine-tuning on the REC datasets (RefCOCO, RefCOCO+, and RefCOCOg) results in similar robustness with a minor differences in turbulence. 
RefCOCO contains simple expressions, while RefCOCO+ has appearance-based prompts, and RefCOCOg includes more elaborate and detailed language; all based COCO with differently phrased captions.
Empirically, this implies \insight{the descriptive nature (expressiveness) of captions or text prompts used in training has seemingly weak/limited impact on robustness}. 
This yet another instance of annotation paying negligible influence on robustness.

\figno{\cref{fig:dirty_captions} right)} evaluates GLIP variants on Flickr30k on 2 sets of captions:
1) \textit{Original (dark)} captions from the dataset , and 
2) \textit{LLM-modified (light)} captions obtained from Granite LLM~\cite{mishra2024granite}, generating five augmented captions infused with context of pixelation/low-resolution. For example, ``\textit{A boy smiles in front of a stony wall in a city}'' becomes ``\textit{In a low-resolution cityscape, a boy's smile is captured against a rough stone wall}''. 
Evaluation occurs on the average of vision embedding fused with 5 LR captions;  exhibiting consistent robustness across severities across models. Thus, it's safe to say \insight{prompt engineering to introduce noise-awareness in 
test captions has marginal impact}.

\begin{wrapfigure}{r}{0.5\textwidth}
\centering
\includegraphics[width=\linewidth]{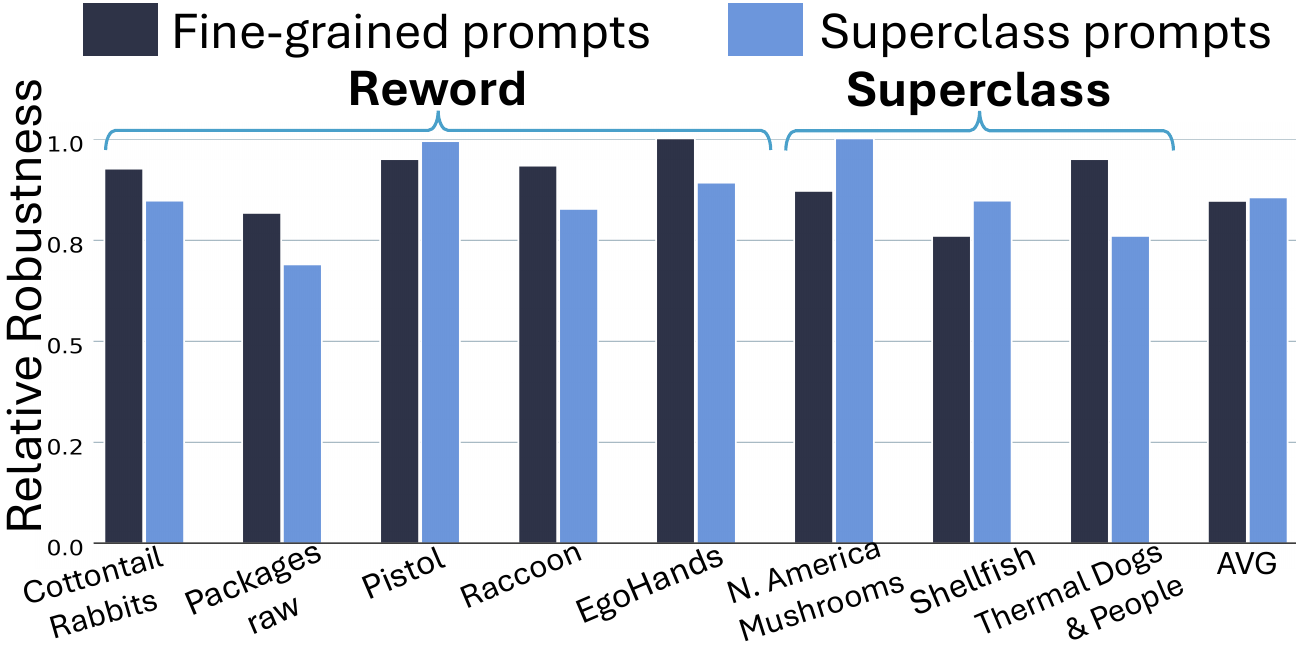}
\caption{\textbf{Fine-grained vs Superclass:} 
Similar robustness on ODinW-13 datasets with pixelation (GLIP-T). Datasets ($\frac{5}{13}$)  with accuracy $\simeq0$ (random prediction) not shown, 1-class datasets `reworded' \&  multiple classes clubbed as `superclass'. 
}
\label{fig:odinw_superclass}
\end{wrapfigure}

ODinW-13 is \textit{evaluated} for robustness  using descriptive fine-grained prompts and coarse-grained superclass prompts as shown in \figno{\cref{fig:odinw_superclass}}.
Superclass prompts groups multiple fine-grained classes under one parent superclass (and reworded if the dataset has only 1 class).  
This shows a small difference in robustness, with mean $\Delta$ of $\simeq 0.14$ for pixelation, $\simeq 0.17$ for turbulence, $\simeq 0.12$ for motion blur, and negligible $\Delta$ overall.
Lower performance on coarse-grained superclass prompts can explain lower robustness (linearity between robustness \&  accuracy \cref{fig:acc_robustness} (right)).
On average, \insight{superclass annotations have limited impact on robustness}. 

\noindent
\takeaway
The above findings suggest that expressive captions (during fine-tuning), and prompts engineered with the context of degradations (during evaluation) do not significantly improve robustness.
Combining these, with previously observed results like the marginal impact of text backbone, (\cref{fig:backbones} (right)), and role of annotations (\cref{sec:dataset_robustness}), its safe to say that \textbf{language expressiveness and distortion-awareness
has minimal influence on `visual' robustness}.  
To our knowledge, we are among the first to demonstrate this limited role of language, hinting that future robustness efforts targeting vision modality is likely to yeild more success.  
Another potential direction is to \textit{re-train} entire models with prompts injected with noise context, rather than at evaluation. 
This would require a new dataset with noise-aware captions and annotations for caption–image alignment and is \textit{left as future work}.

%% file: sec/predicted.tex
\section{HOW: Empirical Validation of Our Analysis}

\label{sec:future_work}
Next, we focus on the empirical validation of our findings and try to explain observations from previously published works.

\input{sec/valid_model_design}

\input{sec/valid_prev_work}

\input{sec/valid_dataset}

%% file: sec/valid_model_design.tex
\subsection{Validating Model Design on Real-world datasets}

\cref{sec:feats_viz} \& \cref{fig:backbones}  (right) in our analysis has highlighted \textbf{four} key model designs: 

\vspace{4pt}
\noindent \textbf{1)} \textit{Visual backbone} is the primary determinant of robustness. 
Improving it w/o modifying other components (neck, fusion) should suffice.

\noindent \textbf{2)} \textit{Shallow layers} are most adversely affected by visual noise.

\noindent \textbf{3)} \textit{Cross-exchanging} information across backbone layers improves feature overlap across severities, \ie features from pixelated images behave like HQ images. 

\noindent \textbf{4)} \textit{Language Features} 
minimal impact on visual robustness (need not be trained).

\vspace{4pt}
\noindent \textbf{Dataset:} 
Design choices are validated on  \textit{real-world driving dataset} BDD-100K~\cite{bdd100k} (AP50); learning traffic related classes like pedestrian, cars, and bikers.
\textit{Zero-shot} noise robustness is evaluated on \textit{noisy weather} driving datasets, including DAWN~\cite{kenk2020dawn}, Foggy Cityscapes~\cite{SDV18, Cordts2016Cityscapes} (amodal annotation, fog intensity 0.02), and Virtual KITTI 2~\cite{cabon2020virtual}, all evaluated using mAP.
COCO is used to measure the change in original zero-shot performance after adapting the model.
We also show results on WiderFace~\cite{yang2016wider} (trained on Flickr30k), and VisDRONE~\cite{zhu2021detection} (trained on UAVDT~\cite{Uavdt}).

\input{sec/ext_results}

\vspace{4pt}
\noindent \textbf{Architecture:}
\textit{Visual backbone and Shallow layers}
We extend LR-TK0~\cite{pathak2025lrfm} method to hierarchical transformers (\eg Swin) for object detection.
LR-TK0 originally added trainable  tokens on top of a frozen transformer; trained via distillation for low-resolution classification.
Here, we remove the expensive teacher student design and retain only the cost efficient trainable tokens, applied \textit{only to the visual backbone while keeping the entire OV-OD frozen}.
At every layer, \textit{particularly shallow layers}, we insert a fixed $32\times32$ set of trainable spatial tokens. These tokens are interpolated to match the layer spatial resolution and added to the frozen feature maps.
Interpolating a small number of spatial tokens provides two advantages: \textit{1) Flexible hierarchy}: Original LR-TK0 attaches a fixed number of prompts to every ViT layer and cannot handle varying $H \times W$, while interpolation adapts to different spatial resolutions across layers. 
\textit{2) Lower Overhead}: For a $600\times600$ Swin-T input, our $32\times32$ tokens add only 5.7\% parameters, compared to 22.5\% for the fixed token design.
We denote call extension as \textbf{TK0}.  

\vspace{4pt}
\noindent \textit{Cross-exchanging layers}
For cross-exchanging information across layers, we introduce a simple trainable non-local block~\cite{wang2018non}, implemented as a single head self attention module. 
The query, key, and value are formed by concatenating spatial tokens from all layers.
The self attention operation enables tokens to share spatial information across layers. We denote this method as \textbf{NN}.
\SUPP{} has a visual representation of this framework.

\vspace{4pt}
\noindent \textbf{Implementation}
GLIP-T is trained using its default training configuration, on the BDD-100K training set (698,630 image text captions).
Baselines include \textit{`E2E'
}, which trains the entire GLIP model end-to-end (including text backbone), and \textit{`Fuse'}, which freezes the model and trains only the fusion transformer. 
The fusion network is the default OV-OD design of cross-exchanging information across vision backbone layers. 
However, both baselines modify pretrained weights and risk degrading zero shot performance of VLM.

\vspace{4pt}
\noindent \textbf{Results \& Discussion}
\figno{\cref{tab:all_resutls}}
shows zero-shot performance of BDD-100K trained GLIP-T on real-world driving datasets.
\figno{\cref{tab:realworld_dataset}} shows an additional zero-shot evaluation on two challenging real-world benchmarks: \textit{WiderFace}, which contains unconstrained face images with significant blur, scale variation, and occlusion, and \textit{VisDrone}, which consists of aerial images captured by drones under various weather, lighting, and environmental conditions. 

\noindent \textit{-Baselines:}
\textit{E2E} \& \textit{Fuse} perform poorly on vanilla COCO dataset, highlighting the importance of preserving pretrained weights.
\textit{Fuse}, which keeps the language and vision backbones frozen and  \insight{\textbf{trains only cross-layer fusion}, achieves robustness comparable to end-to-end training}, with overall $\Delta$ of 1.3\% on BDD-100K, 0.5\% on DAWN, 0.3\% on Foggy Cityscapes, and 1.8\% on Virtual KITTI 2, despite improving accuracy by about 33\%, 2\%, 5\%, and 9\% respectively.

\noindent \textit{-TK0 and NN:}
Training our modified TK0 improves robustness while causing minimal drop in zero shot COCO performance since the OV-OD remains entreily frozen. 
However, its robustness is still below E2E and Fuse. A similar trend is observed for NN.
While TK0 lacks cross exchange, NN is a 1-head 1-layer self-attention while fusion network is an entire multi-layered multi-head transformer decoder. 
Combining both insights through \textit{`NN} \& \textit{TK0'} yields improvements comparable to the baselines, and even surpasses them on DAWN and Foggy Cityscapes; suggest fusing spatial vision features, while modifying improving robustness on shallow layers of vision backbone can improves robustness.

\noindent \textit{-Compute Cost:}
For GLIP-T (Swin-T), E2E trains the entire OV-OD model, including the vision backbone, fusion decoder, and language transformer, totaling about \hl{231.76 M trainable parameters}.
Compared to E2E, \textit{Fuse} trains only the \hl{fusion transformer with 91.85 M parameters, making it about $2.5\times$ cheaper}, gaining almost similar robustness. 
\textit{TK0} operates only on the vision backbone with \hl{\textbf{1.57 M} trainable parameters, about \textbf{147.4$\times$} lighter. \textit{NN} introduces just \textbf{0.84 M }parameters, about \textbf{275.7$\times$ } fewer compared to end-to-end}.
In totality, \insight{insights from our analysis (\textit{`NN} \& \textit{TK0’})
achieve similar robustness like end-to-end finetuning using \textbf{2.41 M parameters},  \textbf{96$\times$} lesser number of parameters}.


\input{sec/valid_observe_table}

Robustness advancements rarely extend to VLMs-based OV-ODs, primarily because of the complexity  (\eg 3-5 transformer backbones, GLEE trained on 64 GPUs across 18 datasets, \etc), limiting progress to smaller CNN models such as Faster R-CNN or simpler tasks. 
Insights from our Robust Onion can enable cost-efficient extensions that improve robustness under real-world noise.

%% file: sec/ext_results.tex
\begin{table}[!t]
\centering
\caption{
\textbf{Robust Onion Empirical Validation:} 
BDD-100K trained GLIP-T zero shot evaluation on DAWN, Foggy Cityscapes (Fog City), Virtual KITTI 2, and COCO. 
E2E and Fuse modify pretrained weights (baselines).
PC=Partly Cloudy; O=Overcast; S=Snow; R=Rain; F=Fog; Sa=Sand; All=Overall across all categories. 
Green/red indicates performance $\uparrow$/$\downarrow$ relative to Zero-shot (0-shot). Our contribution in orange. 
}
\label{tab:all_resutls}
\renewcommand{\arraystretch}{1.2}
\setlength\tabcolsep{2pt}
\resizebox{\textwidth}{!}{
\begin{tabular}{c | cccccc | ccccc|c|cccc|c}
\specialrule{1pt}{0pt}{0pt}
\rowcolor{mygray} 
& \multicolumn{6}{c|}{\textbf{Training} }  & 
\multicolumn{11}{c}{\textbf{Zero-Shot Evaluation} }
\\
[-.4pt]\hhline{~ ------ ----- - ---- -} 
\rowcolor{mygray} 
& \multicolumn{6}{c|}{BDD-100K }  & 
\multicolumn{5}{c|}{DAWN}  & Fog & \multicolumn{4}{c|}{Virtual KITTI 2}  &  \\ 
[-.4pt]\hhline{~ ------ ----- ~ ---- } 
\rowcolor{mygray} 
\multirow{-3}{*}{Model} & PC & S & R & F & O  & All & 
F & R & S & Sa  & All & 
City & F & O & R & All & \multirow{-2}{*}{COCO } \\ 
\hline 
0-shot & 
43.4 & 43.7 & 41.7 & 46.6 & 47.4 & 31.4 & 32.6 & 40.5 & 38.5 & 49.8 & 33.8 & 21.4 & 15.6 & 11.2 & 11.5 & 20.9 & 46.6 \\
\hline 
\hline 
& 
71.3 & 72.7 & 71.0 & 69.5 & 77.4 & 66.4 & 31.6 & 36.4 & 38.5 & 45.2 & 36.0 & 26.9 & 16.7 & 12.0 & 12.6 & 31.6 & 3.0 \\ 

\multirow{-2}{*}{E2E} &  
 \BIM{ 27.9}  & \BIM{ 29.0}  & \BIM{ 29.3}  & \BIM{ 22.9}  & \BIM{ 30.0}  & \BIM{ 35.0}  & \TD{ -1.0}  & \SD{ -4.1}  & \TD{ 0.0}  & \SD{ -4.6}  & \TIM{ 2.2}  & \SIM{ 5.5}  & \TIM{ 1.1}  & \TIM{ 0.8}  & \TIM{ 1.1}  & \SIM{ 10.7}  & \BD{ -43.6} 
\\
\hline 

& 
69.4 & 72.1 & 70.3 & 67.7 & 76.6 & 65.1 & 30.0 & 35.9 & 36.3 & 42.5 & 35.6 & 26.6 & 17.2 & 12.3 & 12.7 & 29.8 & 3.8 \\ 
\multirow{-2}{*}{Fuse}   &  \BIM{ 26.0}  & \BIM{ 28.4}  & \BIM{ 28.6}  & \BIM{ 21.1}  & \BIM{ 29.2}  & \BIM{ 33.7}  & \SD{ -2.6}  & \SD{ -4.6}  & \SD{ -2.2}  & \SD{ -7.3}  & \TIM{ 1.8}  & \SIM{ 5.2}  & \TIM{ 1.6}  & \TIM{ 1.1}  & \TIM{ 1.2}  & \SIM{ 8.9}  & \BD{ -42.8} 
\\

\hline 
\hline 
\cellcolor{orange!20} 
 & 
61.5 & 63.0 & 61.0 & 59.2 & 68.4 & 45.1 & 35.3 & 42.3 & 40.6 & 50.1 & 32.5 & 25.5 & 16.9 & 12.2 & 12.6 & 26.1 & 43.2 \\ 
\cellcolor{orange!20} 
\multirow{-2}{*}{TK0} & 
\SIM{ 18.1}  & \SIM{ 19.3}  & \SIM{ 19.3}  & \SIM{ 12.6}  & \BIM{ 21.0}  & \SIM{ 13.7}  & \TIM{ 2.7}  & \TIM{ 1.8}  & \TIM{ 2.1}  & \TIM{ 0.3}  & \TD{ -1.3}  & \TIM{ 4.1}  & \TIM{ 1.3}  & \TIM{ 1.0}  & \TIM{ 1.1}  & \SIM{ 5.2}  & \SD{ -3.4} 
\\
\hline 
\cellcolor{orange!20} 
 & 
61.8 & 64.0 & 61.6 & 63.0 & 68.5 & 45.0 & 33.3 & 41.6 & 38.3 & 49.7 & 33.4 & 25.8 & 16.9 & 12.0 & 12.4 & 28.7 & 22.1 \\ 
\cellcolor{orange!20} 
\multirow{-2}{*}{NN}  & 
\SIM{ 18.4}  & \BIM{ 20.3}  & \SIM{ 19.9}  & \SIM{ 16.4}  & \BIM{ 21.1}  & \SIM{ 13.6}  & \TIM{ 0.7}  & \TIM{ 1.1}  & \TD{ -0.2}  & \TD{ -0.1}  & \TD{ -0.4}  & \TIM{ 4.4}  & \TIM{ 1.3}  & \TIM{ 0.8}  & \TIM{ 0.9}  & \SIM{ 7.8}  & \BD{ -24.5}
\\
\hline 
\cellcolor{orange!20} 
NN \&   & 
66.8 & 68.6 & 66.1 & 65.5 & 73.1 & 55.2 & 34.0 & 41.3 & 39.8 & 49.5 & 36.1 & 28.2 & 17.3 & 12.4 & 13.1 & 29.9 & 33.4 \\ 
\cellcolor{orange!20} 
TK0 & 
\BIM{ 23.4}  & \BIM{ 24.9}  & \BIM{ 24.4}  & \SIM{ 18.9}  & \BIM{ 25.7}  & \BIM{ 23.8}  & \TIM{ 1.4}  & \TIM{ 0.8}  & \TIM{ 1.3}  & \TD{ -0.3}  & \TIM{ 2.3}  & \SIM{ 6.8}  & \TIM{ 1.7}  & \TIM{ 1.2}  & \TIM{ 1.6}  & \SIM{ 9.0}  & \BD{ -13.2}
\\ 
\specialrule{1pt}{0pt}{0pt}
\end{tabular}
`}
\end{table}

\begin{table}[!t]
\centering
\caption{\textbf{More Real World Dataset:} X (Train) $\rightarrow$ Y (zero-shot evaluation). WiderFace has blurry faces of people; UAVDT \& VisDRONE are drone-captured satellite images of streets.  Our Contribution in orange. Best performance in \textbf{bold}.
}
\label{tab:realworld_dataset}
\renewcommand{\arraystretch}{1.2}
\setlength\tabcolsep{10pt}
\resizebox{\linewidth}{!}{%
\begin{tabular}{c|c|cc||cc}
\specialrule{1pt}{0pt}{0pt}
\rowcolor{mygray} 
Method & \# of Trainable &\multicolumn{2}{c||}{Flickr30k $\rightarrow $\textbf{WiderFace}} & 
\multicolumn{2}{c}{UAVDT $\rightarrow$ \textbf{VisDRONE}} \\
\cline{3-6}
\rowcolor{mygray} 
GLIP-T & Parameters (M) & AP $\uparrow$ & AP-50 $\uparrow$& AP$\uparrow$ & AP-50$\uparrow$ \\ 
\hline 
Zero-Shot & - & 11.98 & 26.49 & \textbf{20.44} & 27.09 \\
\hline 
E2E & 231.76 & 10.86 & 24.69 & 18.68 & 30.56 \\ 
Fuse & 91.85 & 10.55 & 24.92 & 20.07 & 29.89 \\ 
\hline 
 \rowcolor{orange!20} 
NN \& TK0 & 2.41 & \textbf{13.23} & \textbf{28.71} & 19.34 & \textbf{30.97} \\ 
\bottomrule
\end{tabular}}
\end{table}

%% file: sec/valid_observe_table.tex
\begin{table}[!t]
\centering
\caption{\textbf{Explanation of previous works with our findings}. Reference (Refer) refers to the relevant section/figures in our work, hosting our explanation.}
\label{tab:all_prev_works}
\renewcommand{\arraystretch}{1.3}
\setlength\tabcolsep{3pt}
\resizebox{\textwidth}{!}{



\begin{tabular}{
>{\centering\arraybackslash}m{2cm}|
>{\centering\arraybackslash}m{4.5cm}|
>{\centering\arraybackslash}m{5.5cm}|
>{\centering\arraybackslash}m{1.2cm}
}
\specialrule{1pt}{0pt}{0pt}

\rowcolor{mygray} 
\multicolumn{2}{c|}{\textbf{Previous Work Observation}} & \multicolumn{1}{c|}{\textbf{Our Explanation}} & \textbf{Refer} \\ 
\hline 
\multirow{2}{*}{Mao \etal~\cite{cocoo}} 
& 
High-accuracy detectors are often more robust.
& Robustness is proportional to clean accuracy.
& \cref{fig:acc_robustness}\\
[-.4pt]\hhline{~---}
& Robustness improvements are due to the vision backbone, rather than the neck or detection head. 
& 
Simple Conv / MLP blocks (neck \& head) have a limited impact on feature collapse. Cross-layer feature exchange (\textit{`fusion'}) between vision features improves this overlap. OV-OD has it by design.
& \cref{sec:feats_viz} \& Enhancer in \cref{fig:pipeline_flow} \\ 
\hline 
Yao \etal~\cite{Yao2024DetCLIPv3TV} 
& Simply replacing the vision backbone from Swin-T with Swin-L improved robustness.
& 
Vision backbone depth impacts robustness; Swin-L (24 blocks) > Swin-T (12 blocks).
& \cref{sec:model_based_analysis} \& 
\cref{fig:backbones} (right) 
\\
\hline
Zhou \etal~\cite{zhou2022understandingrobustnessvisiontransformers} \& Pathak \etal~\cite{pathak2025lrfm} 
& Vision model shallow-layer features exhibit greater degradation than deeper-layer features under noise.
& Shallow layers have significant feature collapse (distinct non-overlapping clusters  of noisy-clean feats), compared to deeper ones.
& Layer \#1 and \#2 in \cref{fig:pipeline_flow}
\\ 
\hline 
Yutaro and Mayu~\cite{Yamada_2022_CVPR}
& Robustness contribution from the Swin architecture is more significant than the pretraining strategy \& robustness techniques.
& Backbone seems to have a stronger influence on robustness than the choice of pretraining strategy and datasets, and other bells and whistles. 
& \cref{fig:pretraining} (left) \& \cref{fig:backbones} (right)  \\
\hline 
Chhipa \etal~\cite{chhipa2024openvocabularyobjectdetectorsrobustness}
& GroundingDINO is more robust than Owl-ViT overall.
& GroundingDINO clean accuracy is better than Owl-ViT;  (robustness $\propto$ clean accuracy). 
& \cref{fig:acc_robustness} (right) \\ 
\hline 
Liu \etal~\cite{10.1007/s11263-024-02096-6} 
& 
DETR and Deformable DETR have similar robustness, violating the accuracy-robustness linearity. 
& Both models share the same vision backbone, which likely leads to similar robustness.
& \cref{fig:backbones} (right)
\\
\hline 
Pathak~\cite{pathak2025coarse} & 
External datasets improve the target dataset's robustness (\textit{`task-agnostic robustness'}).   & 
Detecting `\textit{on} what' is more important than `\textit{what}' (annotations' minimal role; images domain makes the difference). &
\cref{sec:model_based_analysis}, 
\cref{sec:dataset_robustness}\\
\hline 
Bhojanapalli \etal~\cite{Bhojanapalli_2021_ICCV} & 
Self-attention is more important than MLP for robustness. & Features exchange (self-attention) reduces feature collapse, while Conv, MLP (feature transformation) do not affect feature collapse.  &
\cref{sec:feats_viz}

\\
\specialrule{1pt}{0pt}{0pt}
\end{tabular}
}
\end{table}

%% file: sec/valid_prev_work.tex
\subsection{Validating insights on Previous Works \& Dataset Bias}

\figno{\cref{tab:all_prev_works}} shows how Robust Onion analysis aligns well with several prior findings, while providing a unified explanation for some of them. 
We also show dataset bias for ODinW-13 (pixelated) in\figno{~\cref{fig:emperical_dataset}}. 
FIBER-B can detect 3-4 objects in pixelated images when the number of objects is small (a \& b) despite heavy occlusion~\cref{fig:occlusion} (mid and right). 
However, when the number of objects are large, the model clubs all of them under one large box.

%% file: sec/valid_dataset.tex
\begin{figure}[!t]
\centering
\includegraphics[width=\linewidth]{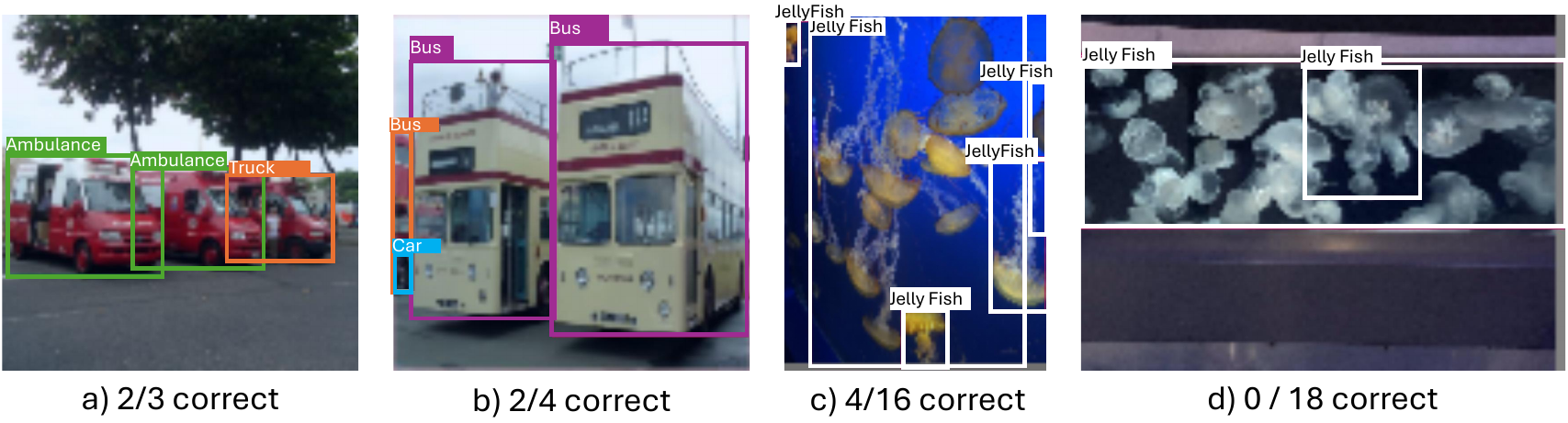}
\caption{ 
\textbf{Empirical Dataset Bias:} Models detect small \# objects despite heavy occlusion (a,b); fail for large \# objects (c, d). FIBER-B on pixelated OdinW-13. 
}
\label{fig:emperical_dataset}
\end{figure}

%% file: sec/conclusion.tex
\section{Conclusion}

\textbf{Robust Onion} provides a detailed analysis of robustness against visual distortions in OV-ODs, systematically \textit{`peeling'} each component, to assess their individual impact. 
Analyzing detectors under realistic, yet underexplored, visual distortions (mirroring real-world noise feature collapse) reveals:
1) Vision backbone dominates robustness, outweighing architectural variations or pre-training choices.
2) Shallow layers are most noise-sensitive.
3) Cross-exchange of features within the vision backbone improves robustness.
4) Prompt/caption expressiveness has minimal effect on robustness during both inference and fine-tuning. To further validate our analysis, we test cross-exchanging backbone features to show improvement on real-world autonomous driving datasets. 
We additionally verify and explain some of the robustness-related observations as reported in previous works. 
Robust Onion
provides a clear, actionable roadmap for advancing the robustness of next-generation OV-OD.

%% file: sec/acknowledgements.tex
\section*{Acknowledgements}
The authors thank Steven Dick (UCF High-Performance Computing) and Shyam Marjit (Centre for Data Science, Indian Institute of Science) for their help and contributions to this project.

%% file: Supplementary/all_files.tex

\title{Robust Onion: Peeling Open Vocab Object
Detectors Under Noise \\\textit{Supplementary}
}

\titlerunning{Robust Onion}

\author{Priyank Pathak*\and
 Mukilan Karuppasamy*\and
 Aaditya Baranwal \and \\ Shruti Vyas\and Yogesh S Rawat}

\authorrunning{Pathak et al.}

\institute{UCF Institute of Artificial Intelligence, University of Central Florida (UCF)
\\
\email{\{priyank,aaditya.baranwal,shruti,yogesh\}@ucf.edu} \hspace{1pt} \email{mukilan.nitt@gmail.com}\\
\blfootnote{*Equal contribution}
}

\maketitle

\input{Supplementary/table_of_content}  


\input{Supplementary/motivate}

\input{Supplementary/model_zoo}

\input{Supplementary/dataset_zoo}

\input{Supplementary/figures_3_9}

\input{Supplementary/figures_9_all}

\input{Supplementary/dataset_analysis}


\input{Supplementary/proposed}

\input{Supplementary/ethical_implications}

%% file: Supplementary/table_of_content.tex
\section*{Table of Content}
\begin{enumerate}
\item \figno{\Cref{sec:rgb_ex}} highlights some RGB examples, and predictions on various synthetic and real-world noise examples.  
\item 
\figno{\Cref{sec:model_zoo}}, and 
\figno{\Cref{sec:dataset_zoo}} has details for various datasets and models used in our analysis, 
\item \figno{\Cref{sec:examples}} have variants of various analysis shown in the main submission, but generalized for all severities, noises, and the LVIS dataset. 
\item \figno{\Cref{sec:dataset_analysis}} have statistics for 
COCO, LVIS, and ODinW-13, as referenced in the main submission ``Section 4.3 WHAT: Robustness as a function of Dataset"
\item 
\figno{\Cref{sec:validation_viz}} has figures for TKO and NN for architectural modification as used in verifying our claim of cross-exchanging features across model layers in helping robustness".
\item \figno{\Cref{sec:ethical}} tries to list some of the ethical concerns and limitations our work. 
\end{enumerate}

%% file: Supplementary/motivate.tex
\section{Visual Examples}
\label{sec:rgb_ex}
\begin{figure}[!h]
\centering
\includegraphics[width=0.95\textwidth]{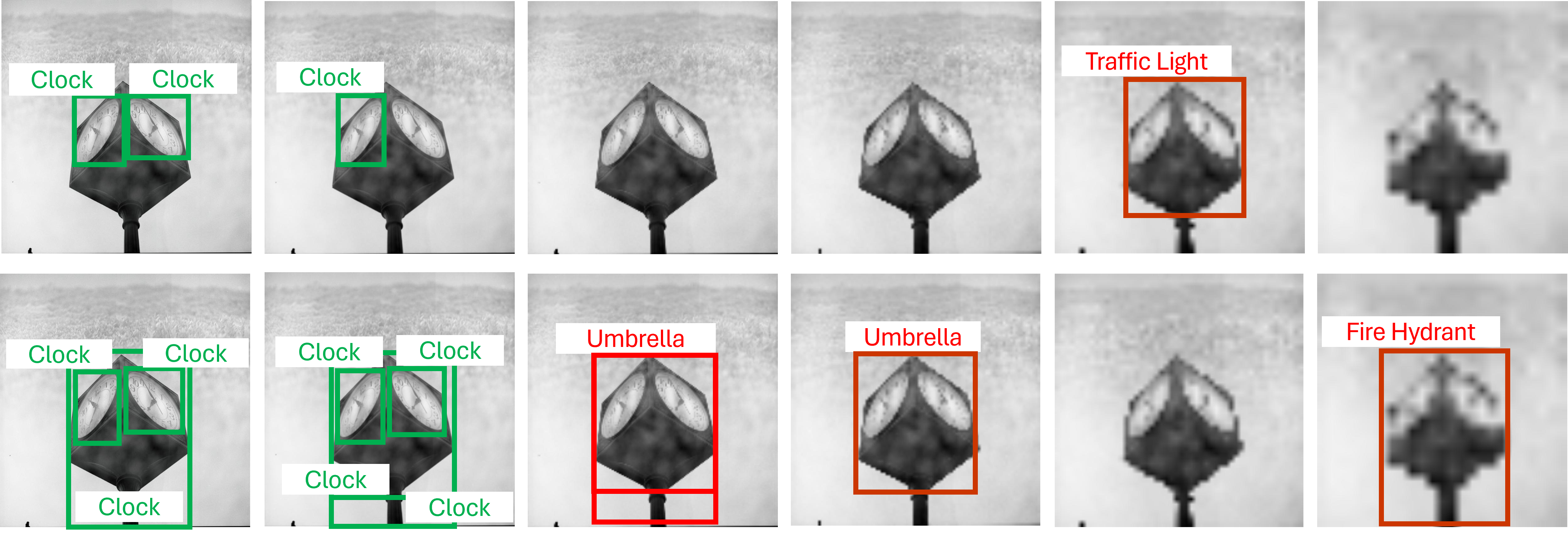}
\caption{
\textbf{Progressive Pixelation:} GLIP~\cite{li2022grounded} (top) and MM-GDINO~\cite{zhao2024open} (bottom); performance degrades on COCO image $(888\times924)$ from left (clean) to right (pixelated) via downsampling by $\frac{1}{2}$.
}
\label{fig:teaser}
\end{figure}

\begin{figure}[!h]
\centering
\includegraphics[width=\textwidth]{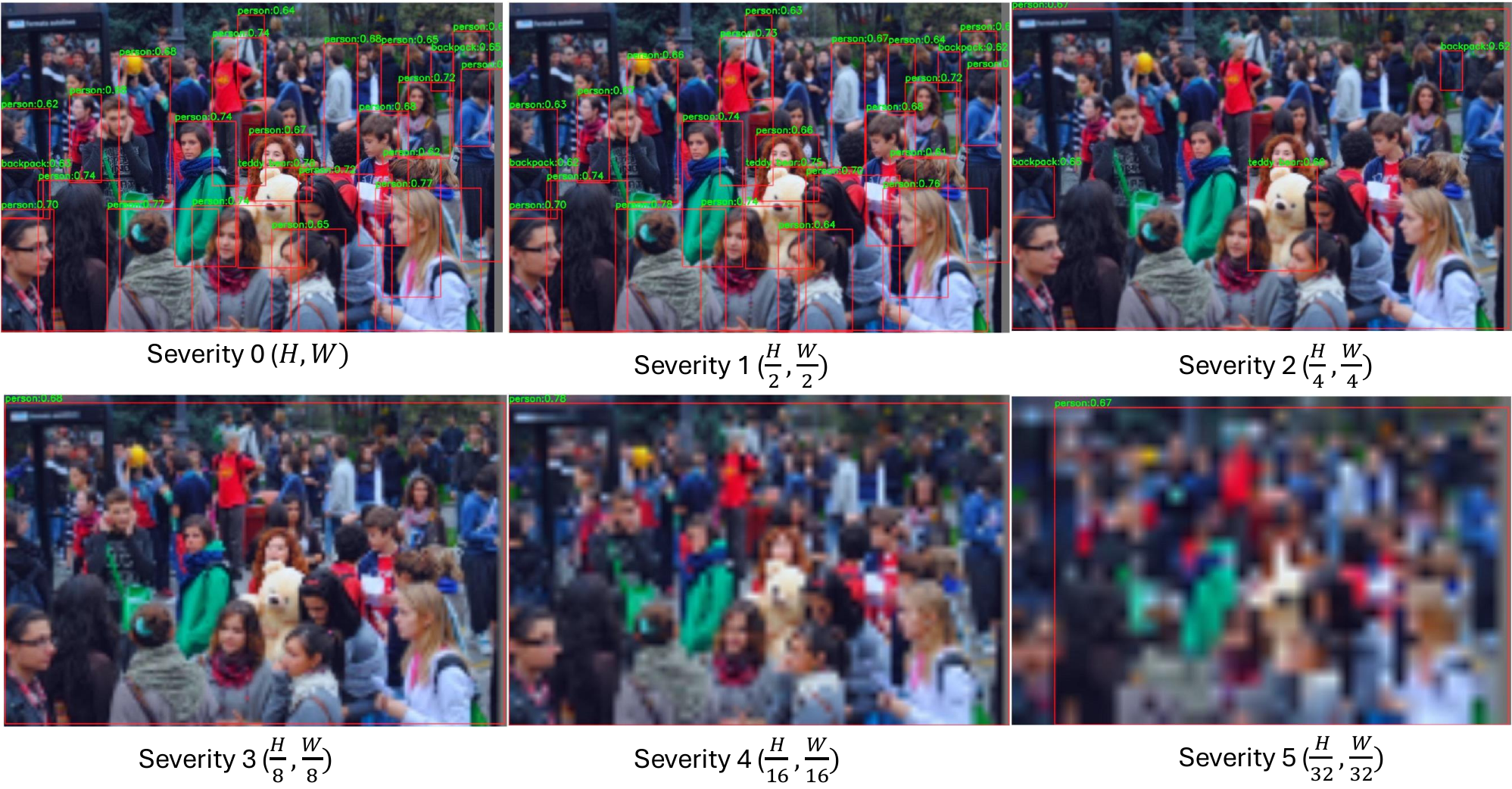}
\caption{\textbf{Models progressive degradation with pixelation.}}
\label{fig:motivation2}
\end{figure}

\begin{figure}[!h]
\centering
\includegraphics[width=\textwidth]{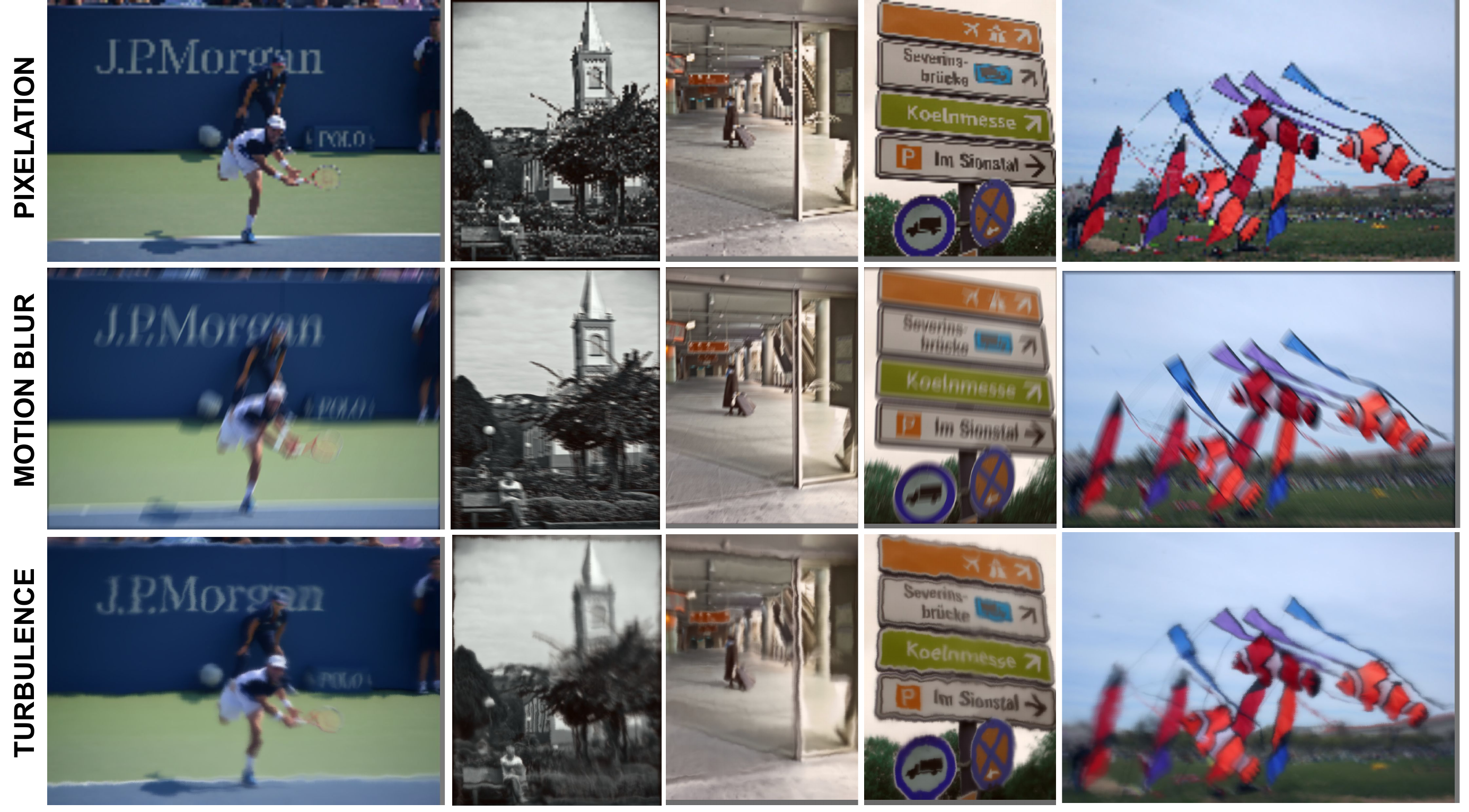}
\caption{\textbf{Visualizations for synthetic noise} Samples of noise perturbations}
\label{fig:real_world_sample}
\end{figure}

\begin{figure}[!h]
\centering
\includegraphics[width=\textwidth]{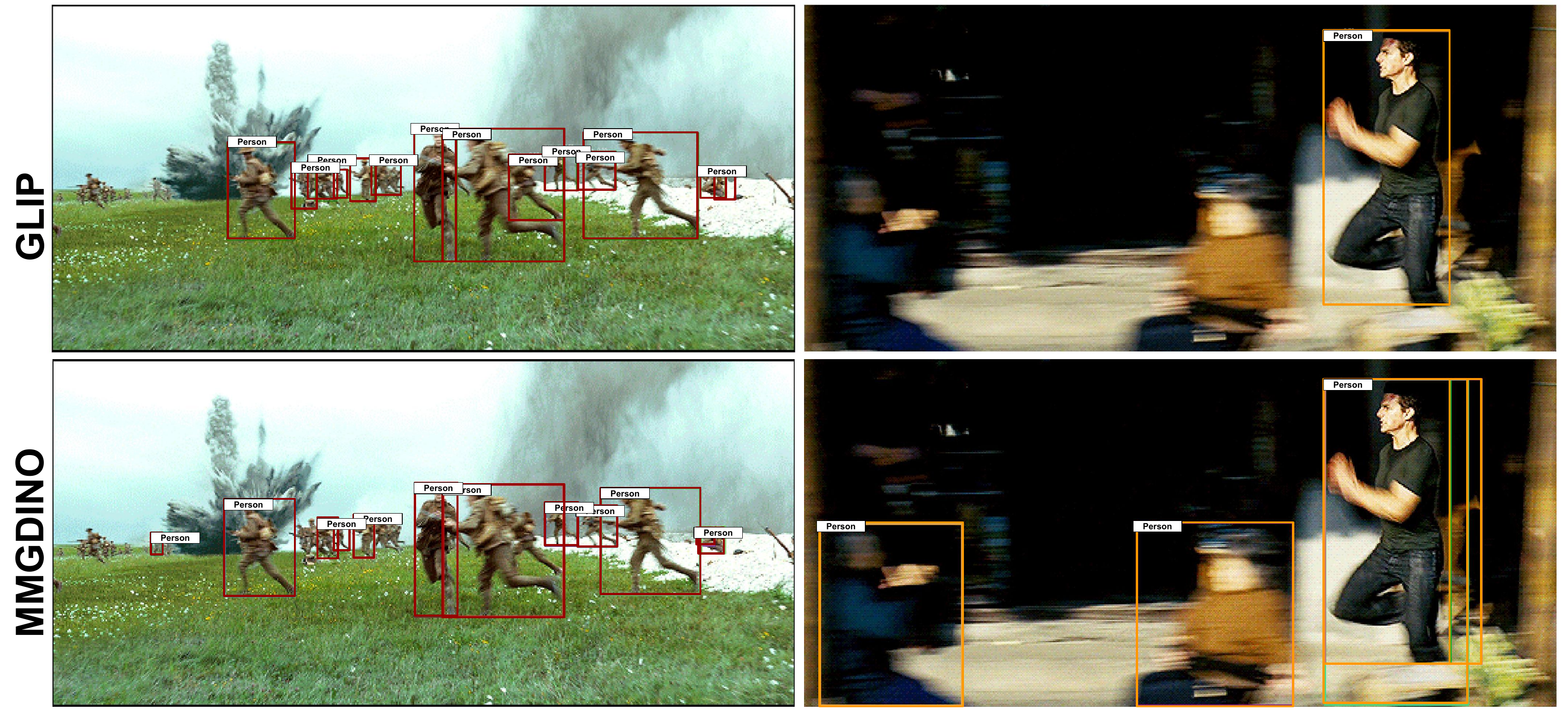}
\caption{\textbf{Sample detection results on real-world images collected from the internet without synthetic perturbations.}}
\label{fig:run}
\end{figure}

Image perturbations significantly affects the performance of the detection models. As, we increase the severity, models often misclassify objects (Figure \ref{fig:teaser}) and fail to preserve accurate bounding box predictions (Figure \ref{fig:motivation2}). Some samples of the perturbations are shown in \Cref{fig:real_world_sample}. Some sample detections on images without synthetic perturbations is shown in \Cref{fig:run}. This shows the model fails to detect accurately, even on the most prominent class (person).

\begin{figure}[!h]
\centering
\includegraphics[width=0.8\textwidth]{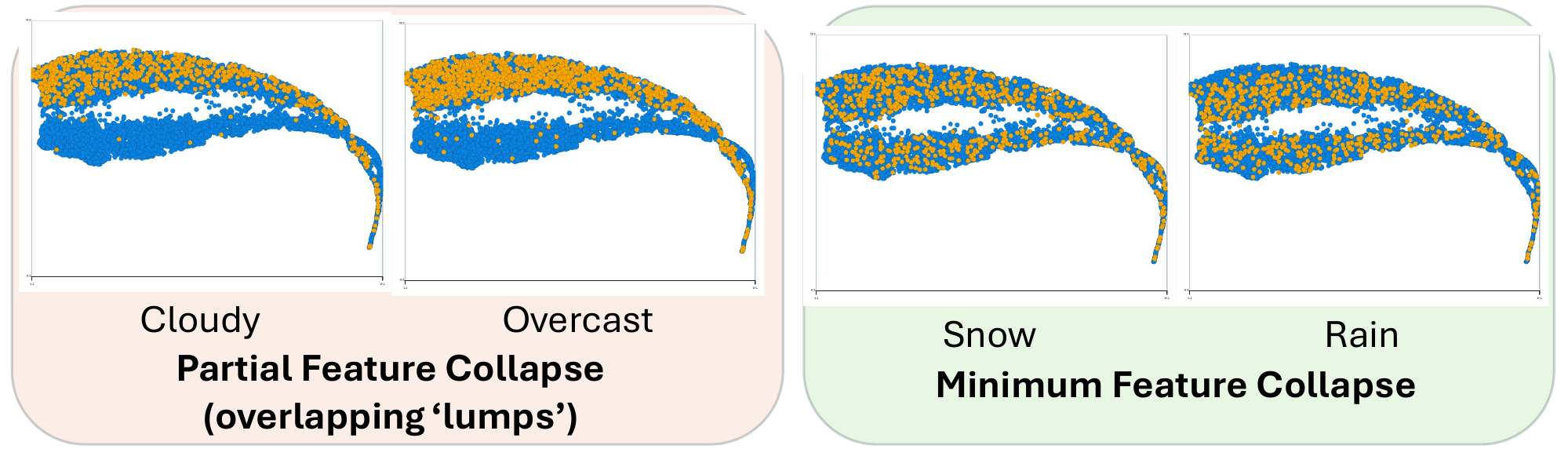}
\caption{
\textbf{Real World} BDD-100K all Feature Collapses.
  } 
\label{fig:all_feature_collapse}
\end{figure}

\begin{figure}[!h]
\centering
\includegraphics[width=0.8\textwidth]{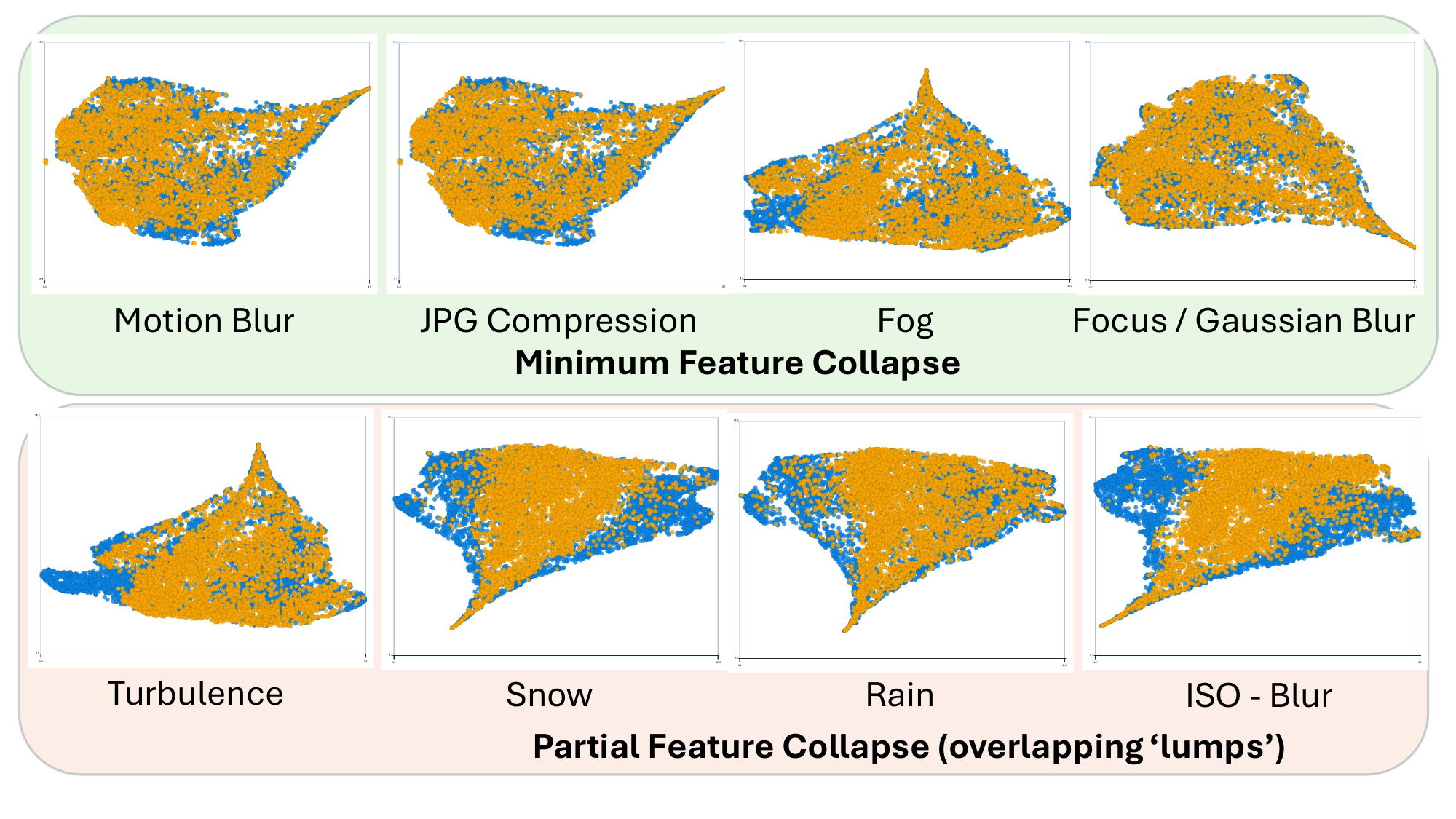}
\caption{
\textbf{Synthetic Noises mimicking Real World} all Feature Collapses.
}
\label{fig:all_feature_collapse2}
\end{figure}

\Cref{fig:motivation2} is the continuation of \Cref{fig:teaser}, indicating how as severity of pixelation is increased, GLIP detection ability degrades. 
Two visible degradations are: 1) The model can't detect multiple objects; instead, it clubs them all under one detection at a lower severity 
2) The model loses the capability of detecting small objects. 

\Cref{fig:all_feature_collapse} shows the feature collapse existing in the current BDDK dataset. Cloudy and Overcast images in the dataset show more separation (stronger perturbation), whereas snow and rain subsets show less collapse (weaker perturbation). We replicate the similar feature collapse using synthetic perturbations as shown in \Cref{fig:all_feature_collapse2}.

%% file: Supplementary/model_zoo.tex
\section{Model Zoo}
\label{sec:model_zoo}
\begin{table*}[!h]
\centering
\caption{
\textbf{Benchmark Models (6 Models and 45 Backbones):} Pre-training is image-text pairs from datasets like Object365~\cite{shao2019objects365}, OpenImages~\cite{krasin2017openimages}, GoldG~\cite{kamath2021mdetr}, CC~\cite{sharma2018conceptual}, \emph{etc}. Visual Backbone uses Swin-Transformer~\cite{liu2021swintransformerhierarchicalvision} (mostly) and ResNets~\cite{he2015deepresiduallearningimage}.
}
\centering
\renewcommand{\arraystretch}{1}    \setlength\tabcolsep{1.2pt}
\scalebox{0.8}
{
\begin{tabular}{l|c|c|c|c}
\toprule
Models & \multicolumn{2}{c|}{\# Backbones and Size (in Million)} & \multicolumn{2}{c}{Pretraining Datasets and Size (in Million)}\\ \midrule

\textbf{RegionCLIP} & 8 ResNets (RN50 \& RN50x4) & 65-114 & CC3M, COCO Caption & 3-3.1\\\midrule

\multirow{2}{*}{\textbf{FIBER}} & 6 Swin-Transformers & \multirow{2}{*}{252} & 
\multirow{2}{*}{COCO, CC, SBU, VG} & \multirow{2}{*}{13}\\
& (Swin-Base) & & \\
\midrule

\multirow{2}{*}{\textbf{GLIP}}
 & 5 Swin-Transformers & \multirow{2}{*}{152-430} & O365, GoldG (Flickr30K+VG+GQA)& 
 \multirow{2}{*}{0.66-27}\\ 
&(4 Swin-Tiny \& 1 Swin-Large)&&CC3M, SBU, CC12M, OI &\\\midrule

 & 9 Swin Trans (5 Swin-Tiny, & {174-343} & 
O365, GoldG, OI, GRIT, & \\ 
\textbf{MM-GDINO}& 2 Swin-Base, 2 Swin-Large) && V3Det,COCO, RefCOCO, & 1.7-13 \\
 & & & RefCOCO+, RefCOCOg \\\midrule

\multirow{3}{*}{
\textbf{YOLO }}
 & 7 YOLOv8(1 YOLOv8-S, & & \\ 
& 1 YOLOv8-M, 3 YOLOv8-L, & {76-168} & {O365, GoldG, CC3M} & {1.4-1.6} \\ 
&  1 YOLOv8-X, 1 YOLOv8-XL)&&\\\midrule

\multirow{6}{*}{\textbf{GLEE}}
 & 3 ResNets (RN50), & 
 \multirow{6}{*}{121-476}
 & Stage-1: O365, OI; & \\  
 & 3 EVA-02 Large, &  & Stage-2: COCO, LVIS, BDD, & Stage-1: 3.6 \\
& 3 Swin Transformers & &
YTVIS19, YTVIS21, OVIS, 
& Stage-2: 0.9\\
& (Swin-Large) && 
RefCOCO, RefCOCO+, RefCOCOg, & Stage-3: 7.3 \\  
&&& 
VG~, VOS~,  
RVOS~, UVO~,  & \\
& & & UVO-dense~; Stage-3: SA1B~, GRIT~ \\ 
\midrule
\bottomrule
\end{tabular}
}
\label{tab:model_zoo}
\end{table*}

In Table~\ref{tab:model_zoo}, we present the details of all the benchmark models considered, including their visual backbones, sizes, and the corresponding pre-trained datasets along with their sizes. RegionCLIP~\cite{zhong2022regionclip}, YOLO~\cite{yoo2024and}, and GLEE~\cite{Wu_2024_CVPR} are the models that have a ResNet-based visual backbone. 
On the other hand, models like FIBER~\cite{fiber2022}, GLIP~\cite{li2022grounded}, MM-GDINO~\cite{zhao2024open}, and GLEE~\cite{Wu_2024_CVPR} leverage Swin Transformers. Notably, the GLEE~\cite{Wu_2024_CVPR} model also uses \texttt{EVA-02 Large} backbone, which is the largest backbone considered in the study and greatly contributes to the higher robustness.

%% file: Supplementary/dataset_zoo.tex
\section{Dataset Descriptions}
\label{sec:dataset_zoo}
We evaluate the zero-shot performance of object detectors on three standard benchmarks to analyze robustness against pixelation:
\textbf{COCO}~\cite{lin2015microsoft} (val2017): Contains 5,000 images with 80 object categories. The validation set includes approximately 36,781 object instances.
\textbf{LVIS}~\cite{gupta2019lvis, kamath2021mdetr} (MiniVal): A long-tail detection dataset comprising 1,203 object categories. The MiniVal set contains 5,000 images with about 62,397 object instances.
\textbf{ODinW-13}~\cite{li2022grounded}: A collection of 13 small out-of-distribution datasets, totaling approximately 3,235 images across diverse domains.
For language-based analysis, we evaluate \textbf{Flickr30k Entities}~\cite{7410660} dataset, which contains 31,783 images and 275,775 bounding boxes. This dataset is commonly employed in pretraining zero-shot models (referred to as ``Gold'') or fine-tuning them (referred to as ``MDETR'' data).

\noindent \textbf{NOTE: } COCO and LVIS share the same image set but differ in their annotations and train/val/test splits. Regarding object categories, LVIS is a superset of COCO, with COCO's 80 categories as a subset of LVIS's 1,203 categories. This relationship allows for interesting cross-dataset comparisons and analyses.



\subsection{Referring Expression (RefCOCO, RefCOCO+,  RefCOCOg)}

Referring Expression Comprehension (REC) is a task, in which, for an image and an expression (\eg \textit{"A red 
colored ferrari"}), the model should detect the region corresponding to the expression. This task was evaluated on RefCOCO, RefCOCO+, and RefCOCOg, which is derived by the detail in the expressions. 

\textit{RefCOCO:} Collected using an interactive two-player game called ReferItGame, where one player described a target object in an image, and the other had to identify it. As a result, the referring expressions are typically short and direct, averaging around 3–4 words. These expressions commonly include both appearance and spatial cues.
Example: \textit{"The red and white checkered table on the left"}.

\textit{RefCOCO+:} Created using the same ReferItGame framework, but with one key restriction: annotators were not allowed to use absolute spatial terms (such as “left,” “right,” “top,” etc.). This restriction forces the referring expressions to rely solely on appearance, attributes, and relative object descriptions, rather than location-based cues.
Example: \textit{"The giraffe with lowered head"}. 

\textit{RefCOCOg:} Unlike RefCOCO and RefCOCO+, it was collected offline (not through a game), by making annotators write longer and more natural, descriptive, and contextual expressions. On avg., expressions are around 8 to 9 words long, often including complex language, object relationships, and scene-level reasoning. 
Example: \textit{"An adult giraffe scratching its back with its horn"}.

\noindent \textbf{NOTE:}
REC dataset have abnormally high robustness scores, as REC fine-tuned models already has low accuracy of FIBER-B REC finetuned models on COCO (and LVIS), which results in a small drop in accuracy on noises (random predictions remain random), giving "abnormally high robustness scores"~\cite{pathak2025lrfm}.

\subsection{External Real-World Datasets: BDD-100K, DAWN, Foggy Cityscapes, and Virtual KITTI 2}
To verify our analysis of the role of cross-exchanging features in improving robustness, we train our model on BDD-100K~\cite{bdd100k}, and evaluate on other driving datasets like Foggy Cityscapes~\cite{SDV18, Cordts2016Cityscapes}, Virtual KITTI 2~\cite{cabon2020virtual}, and DAWN~\cite{kenk2020dawn}. 

\textit{BDD-100K} is a real-world driving dataset consisting 10000 images in the validation set, and 69,863 images in train set. 
The classes used for training and testing are 
['person. car. \textbf{rider}. bus. \textbf{truck}. \textbf{bike}. \textbf{motor}. \textbf{traffic light}. \textbf{traffic sign}']. These classes (in bold) rarely appear in GLIP's vanilla Flickr30k training, and thus need to be made familiar with such classes.
BDD-100K trains model on weather conditions like partly cloudy (PC), Snowy (S), Rainy (R), Foggy (F), Overcast (O), Undefined, and Clear. It also subdivides the images into time categories: undefined, night (luminosity), dawn/dusk, and daytime. Overall (All) evaluates across all categories and weather/time. 

\textit{Foggy Cityscapes}
Fog simulated CityScape dataset with 488 images containing  ['car. \textbf{bicycle}. bus. \textbf{motorcycle}. \textbf{truck}'] classes. We use the Amodal annotation (alternative being modal annotation, generated via Yolo models). 
\textbf{Fog severity} include 0.005, 0.01, 0.02, with 0.02 being the most severe dense fog.

\textit{DAWN}
Dawn dataset has 926 images, with classes ['car. person. \textbf{bicycle. motorcycle. truck}. \textbf{bus}'], categorized under Fog (F), Rain (R), Snow (S), Sand (Sa, like dust), and Overall (All).  

\textit{Virtual KITTI 2}
This is a synthetic dataset with 16740 data points with images categorized as clean (15-deg-left, 15-deg-right, 30-deg-left, 30-deg-right, and clone), Fog (F), Morning, Overcast (O), Rain (R), Sunset, and Overall (All). 
Classes for detection include : ['car. \textbf{truck. van}'].

%% file: Supplementary/figures_3_9.tex
\section{Additional Figure / Details in Main Submission}
\label{sec:examples}

\subsection{Fig 3}
\begin{figure}[!ht]
\centering
\begin{subfigure}[t]{0.47\textwidth}
\centering
\includegraphics[width=\linewidth]{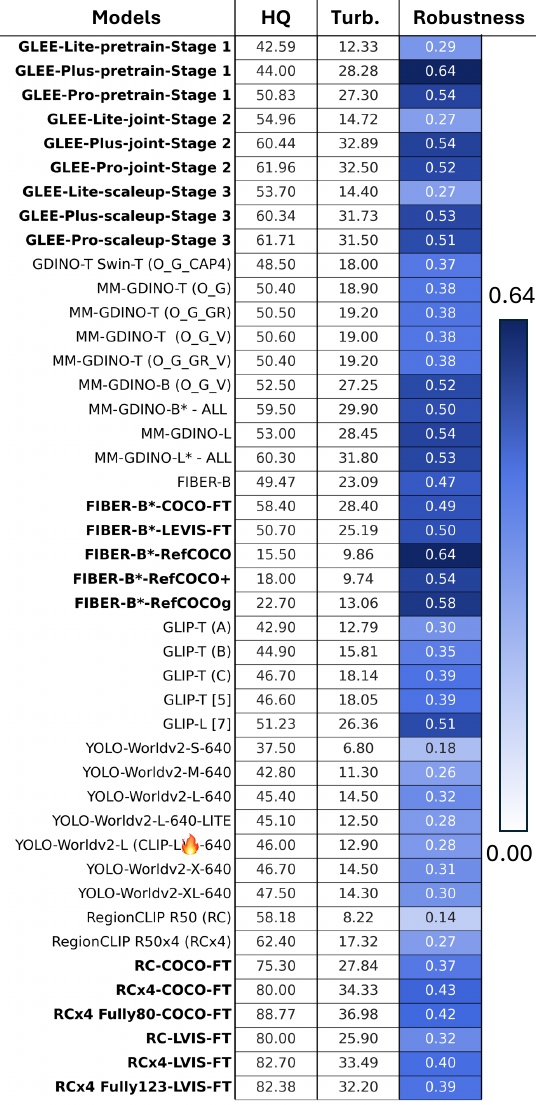}
\caption{
\textbf{Robustness scores for all models under atmospheric turbulence perturbation} 
}
\label{fig:heatmapp_turb}
\end{subfigure}
\hfill
\begin{subfigure}[t]{0.47\textwidth}
\centering
\includegraphics[width=\linewidth]{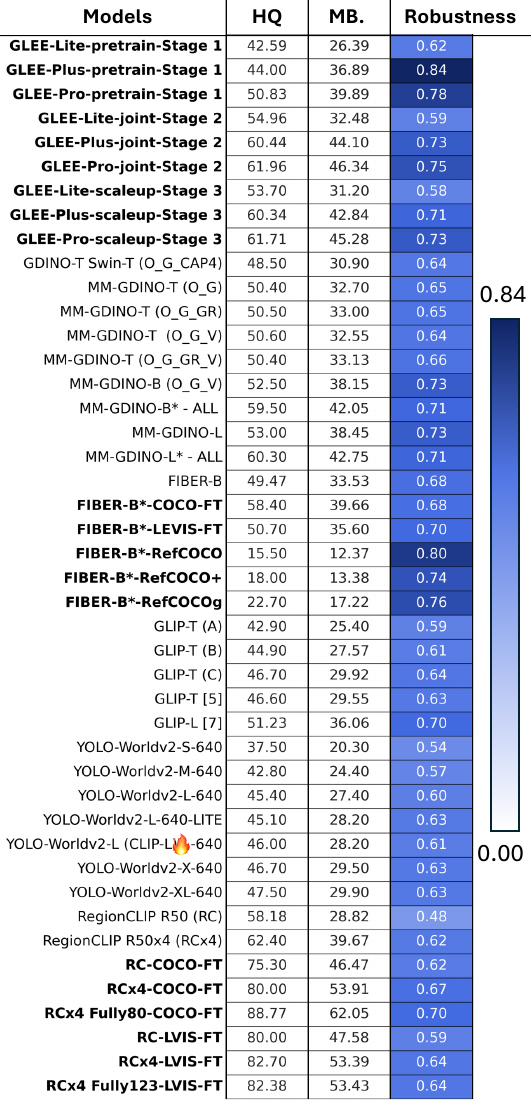}
\caption{
\textbf{Robustness scores for all models under motion blur perturbation} 
}
\label{fig:heatmap_motion}
\end{subfigure}
\caption{
}
\label{fig:heatmap_noise}
\end{figure}

Here we show the robustness scores for all models perturbed with atmospheric turbulence in \Cref{fig:heatmapp_turb} and with motion blur in \Cref{fig:heatmap_motion}.

\subsection{Fig 5(right)}
 Fig 5 (right) showed results for COCO for pixelation; here we show results for Accuracy vs Robustness for LVIS. A similar linear relationship between robustness and accuracy exists here as well. 

\begin{figure*}[!h]
\centering
\includegraphics[width=\linewidth]{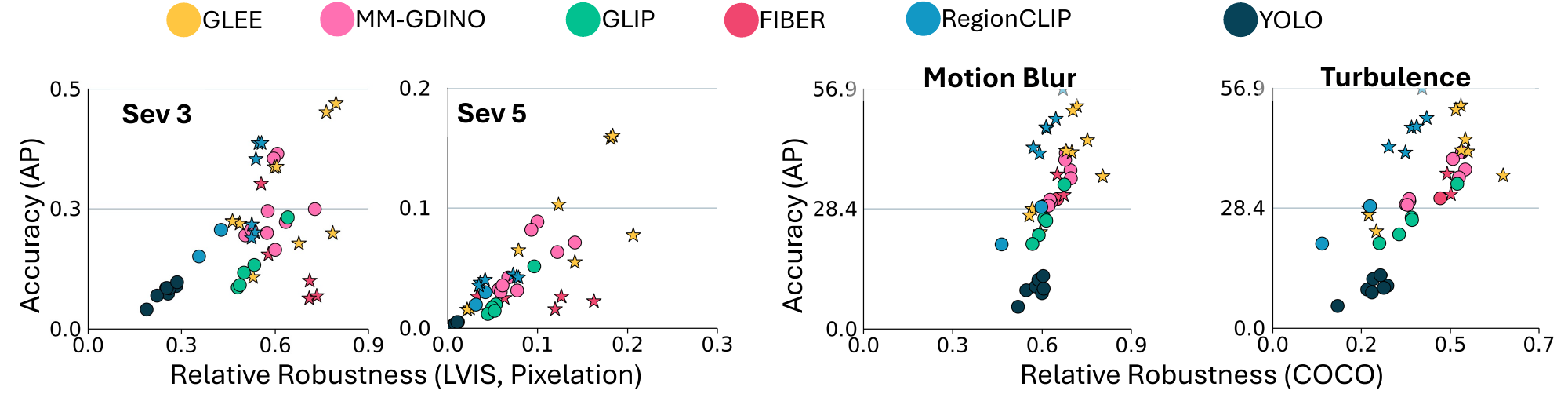}
\caption{
\textbf{Accuracy, Robustness linear relationship}. Extension of Fig 5 (Right).
}
\label{fig:acc_robustness_supp}
\end{figure*}

\subsection{Fig 6 (left)}
\begin{figure*}[!h]
\centering
\begin{subfigure}[t]{0.47\textwidth}
\centering
\includegraphics[width=0.75\linewidth]{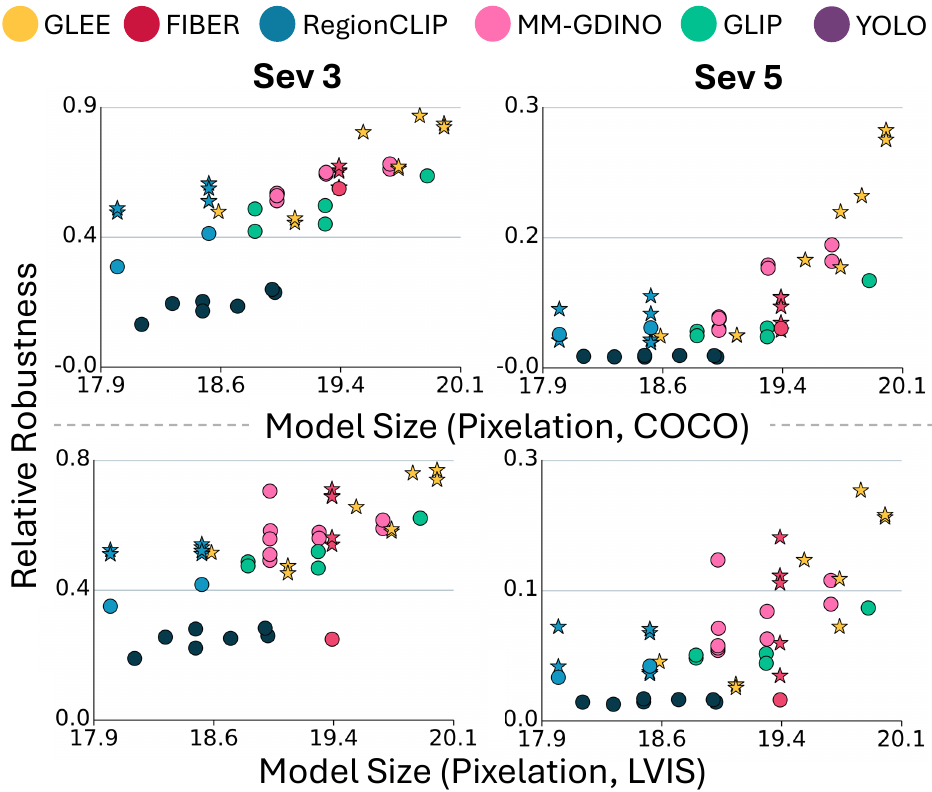}
\caption{
\textbf{Robustness vs model size for pixelation severities}. Extension of Fig 6 (left). 
}
\label{fig:model_size_allsev}
\end{subfigure}
\hfill
\begin{subfigure}[t]{0.47\textwidth}
\centering
\includegraphics[width=\linewidth]{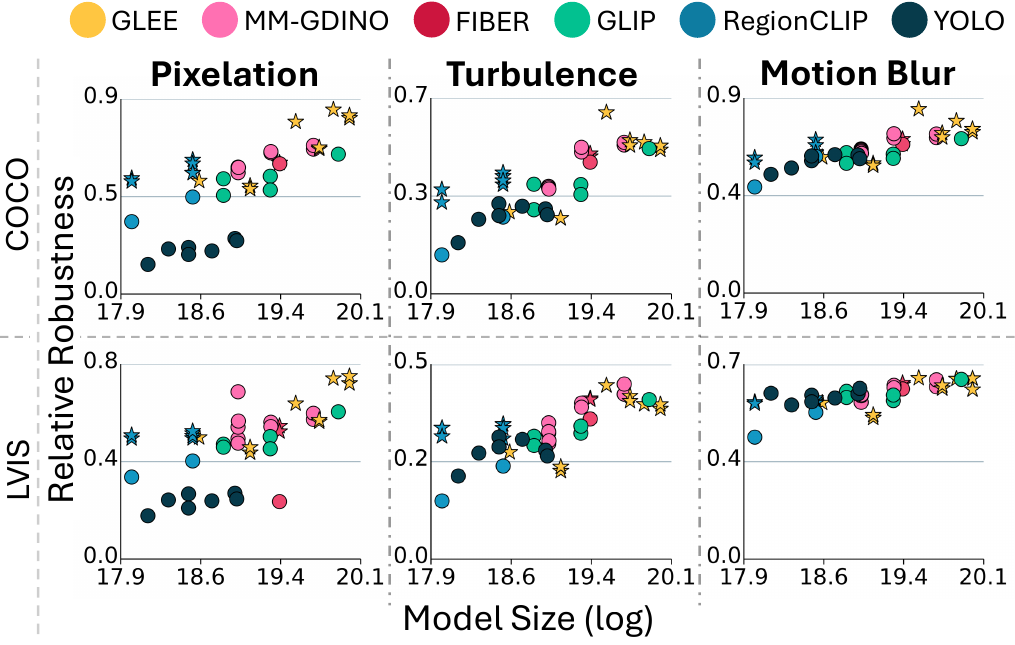}
\caption{
\textbf{Robustness vs model size for real world noises}. Extension of Fig 6 (left). 
}
\label{fig:model_size_noise_allnoise}
\end{subfigure}
\end{figure*}

 Fig 6 (left) showed results for COCO, here we show results for Robustness vs model size for LVIS at sev3 and sev5 in \Cref{fig:model_size_allsev}. We also show the results for turbulence and motion blur in \Cref{fig:model_size_noise_allnoise}

\subsection{Fig 6 (Right)}
\begin{figure*}[!h]
\centering
\includegraphics[width=\linewidth]{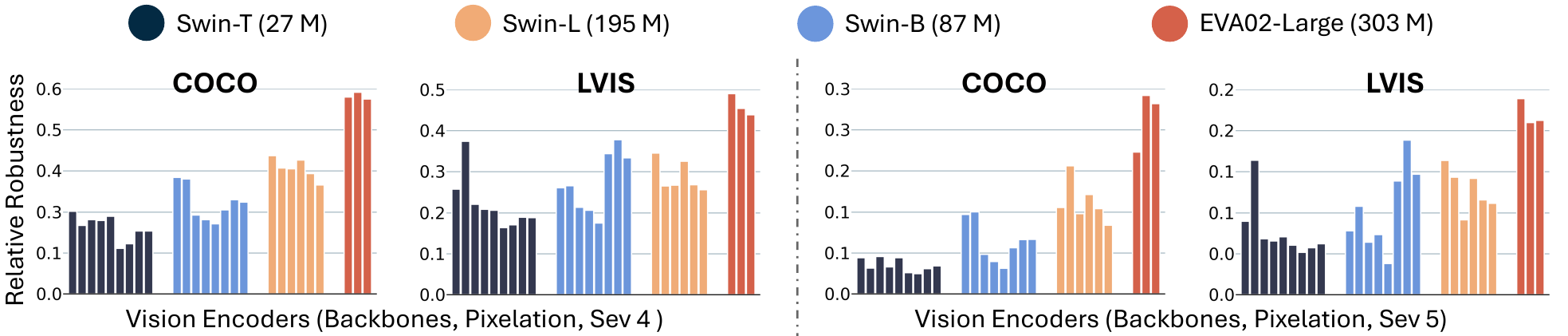}
\caption{
\textbf{Similar Backbone Robustness}. Extension of 
Fig 6(Right), Sev 4 \& Sev 5.
}
\label{fig:backbones_sev5}
\end{figure*}

 Fig 6 (Right) showed results for sev 3 pixelation, here we show results for sev 4 \& 5 in \Cref{fig:backbones_sev5}. 
 Low close-to-random performance, outlier behavior can not draw reliable conclusions, however performance is consistent across backbones.

\begin{figure*}[!h]
\centering
\includegraphics[width=\linewidth]{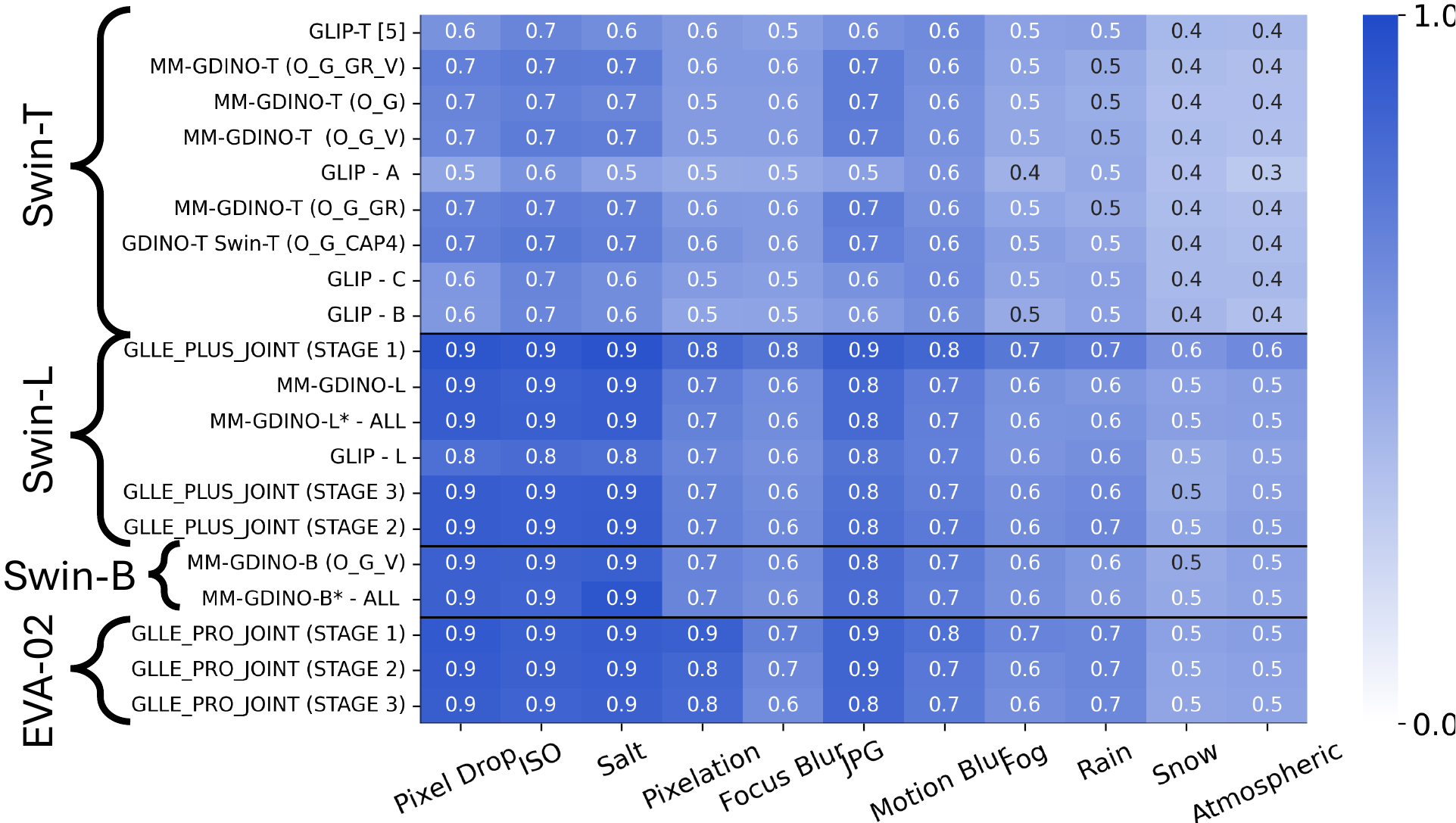}
\caption{
\textbf{All noises for all backbones.}. 
Same details as those of Fig 6 (Right)
}
\label{fig:all_noises}
\end{figure*}

\cref{fig:all_noises} show all kinds of synthetic noises grouped via similarity in backbones.

\subsection{Fig 7 (Left)}

\begin{figure*}[!h]
\centering
\includegraphics[width=0.5\linewidth]{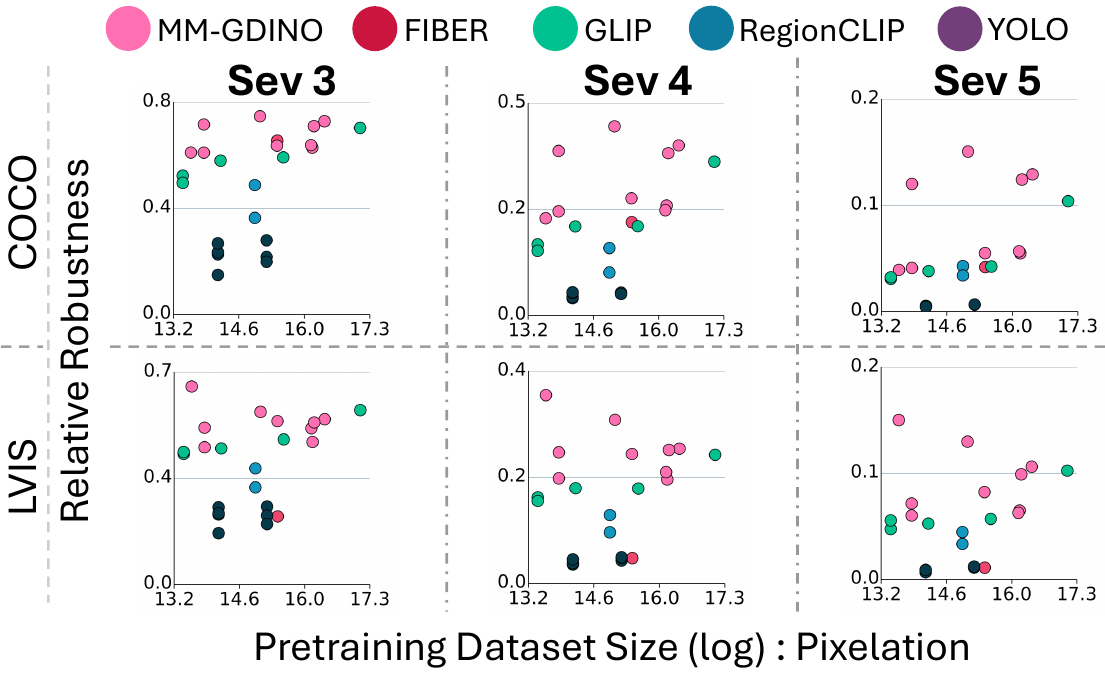}
\caption{
\textbf{Robustness vs Dataset Size for all severity on pixelation}. 
Same details as that of Fig 7 (Left). 
}
\label{fig:pretraining_sev}
\end{figure*}

Fig 7 (Left) showed the results for effect of pretraining dataset size in robustness for all noises. Here we show the same trend for all severity in \Cref{fig:pretraining_sev} 

\subsection{Fig 8}
Fig 8 showed, UMAP plot of features, here we show t-SNE plot for the same. 

For \textbf{GLIP}, backbone features `$\mathcal{B}$' (the last 4 layers of backbone, $\mathcal{B}_1$, $\mathcal{B}_2$, $\mathcal{B}_3$, and $\mathcal{B}_4$) are used for Swin-T transformers with blocks partitions as [2,2,6,2]. These multi-scale features (features from different intermediate backbone layers) are passed through a neck network, which are simple intermediate convolutional layers (channels $\rightarrow$ channels), such as 
$\mathcal{C} (D \rightarrow 256)$ \& $\mathcal{H} (256 \rightarrow 256)$, on these backbone features creating `$\mathcal{N}$' features as $\mathcal{N}_1, \mathcal{N}_2, \mathcal{N}_3, \mathcal{N}_4 \& \mathcal{N}_5$, where 
$\mathcal{N}_3 = \mathcal{H}_1\large(\mathcal{C}_1 \large(\mathcal{B}_4\large)\large)  \parallel 
\mathcal{N}_2$ = $\mathcal{H}_2 \bigl(
\mathcal{C}_2 \large(\mathcal{B}_3\large) +
\mathcal{C}_1\large( \mathcal{B}_4\large) \bigr) 
\parallel 
\mathcal{N}_1$ = $\mathcal{H}_3 \bigl(
\mathcal{C}_3 \large(\mathcal{B}_2\large) +
\mathcal{C}_2 \large(\mathcal{B}_3\large) + \mathcal{C}_1\large( \mathcal{B}_4\large) \bigr) \parallel 
\mathcal{N}_4 = \mathcal{H}_4 \bigl(
\mathcal{C}_1 \large(\mathcal{B}_4\large)
\bigr) \parallel 
\mathcal{N}_5 = \mathcal{H}_5 \bigl(
\mathcal{H}_4 \bigl(
\mathcal{C}_1 \large(\mathcal{B}_4\large)
\bigr) \bigr)$.
The Fusion network induces text context into vision neck features, generating `$\mathcal{F}$' features as $\mathcal{F}_1$, $\mathcal{F}_2$, $\mathcal{F}_3$, $\mathcal{F}_4$, $\mathcal{F}_5$ for $\mathcal{N}_1$, $\mathcal{N}_2$, $\mathcal{N}_3$, $\mathcal{N}_4$, $\mathcal{N}_5$ respectively. 
For plotting, we use 
$\boldsymbol{\mathcal{B}_1, \mathcal{B}_2, \mathcal{B}_3, \mathcal{B}_4, \mathcal{N}_4, \& \mathcal{F}_4}$.

\begin{figure*}[!t]
    \centering
\includegraphics[width=\linewidth]{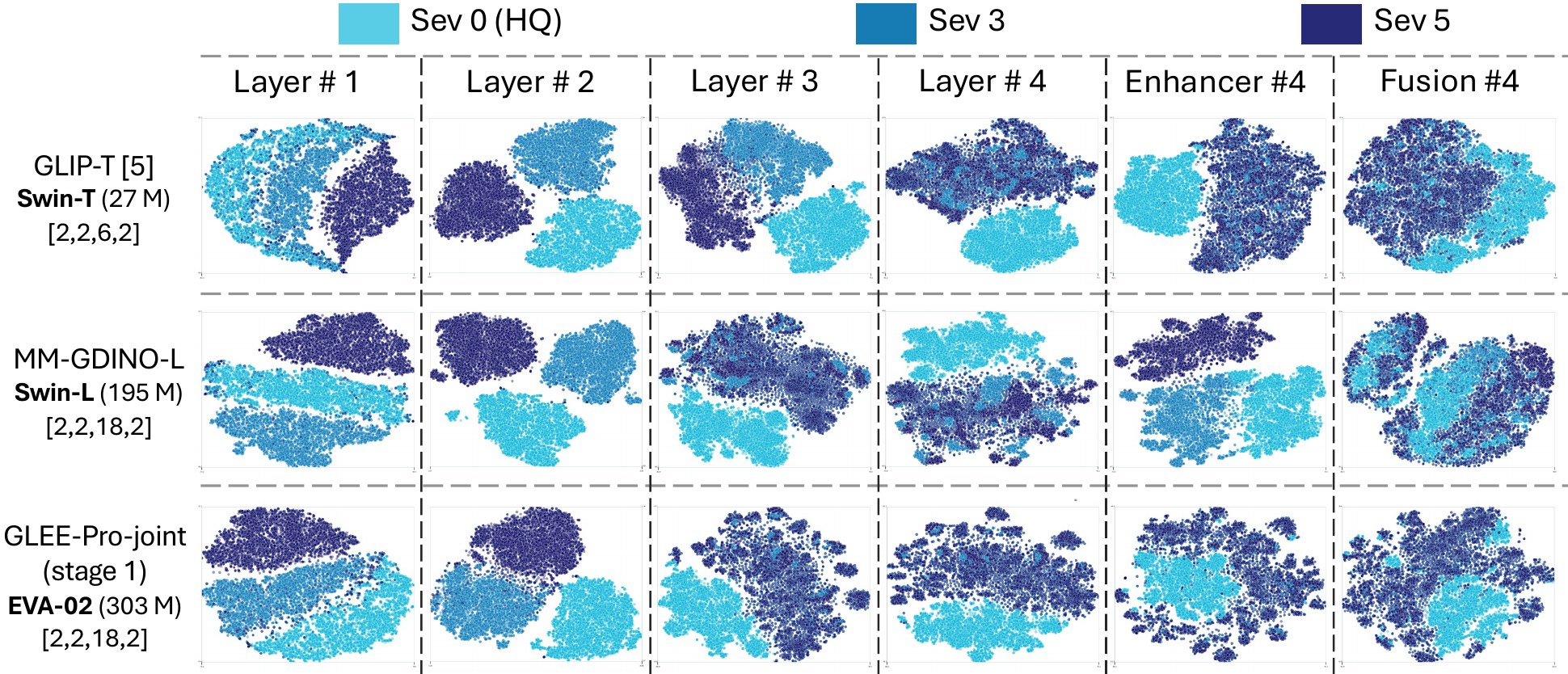}
\caption{
\textbf{Pixelation Features t-SNE:} 
Same details as those of Fig 8. 
}
\label{fig:pipeline_flow_tsne}
\end{figure*}

\vspace{3pt}
For \textbf{MMGDINO} (Swin-Large), backbone features `$\mathcal{B}$' ($\mathcal{B}_1$, $\mathcal{B}_2$, $\mathcal{B}_3$, and $\mathcal{B}_4$) are used with 24 blocks partitioned as [2,2,18,2].
These multi-scale features are passed through a neck network, which produces neck `$\mathcal{N}$' features as $\mathcal{N}_1$, $\mathcal{N}_2$, $\mathcal{N}_3$, $\mathcal{N}_4$, and $\mathcal{N}_5$. 
Here, the neck is a simple convolutional network, 
$\mathcal{N}_1 = \mathcal{C}_1 \large(\mathcal{B}_1\large) \parallel 
\mathcal{N}_2 = \mathcal{C}_2 \large(\mathcal{B}_2\large) \parallel 
\mathcal{N}_3 = \mathcal{C}_3 \large(\mathcal{B}_3\large) \parallel 
\mathcal{N}_4 = \mathcal{C}_4 \large(\mathcal{B}_4\large) \parallel 
\mathcal{N}_5 = \mathcal{C}_5 \large(\mathcal{B}_4\large)$.
This extra neck feature $\mathcal{N}_5$ is termed as \hl{extra\_convs} in the original code. 
The Fusion network consists of an encoder-decoder structure, with early fusion, meaning the encoder fuses the textual feature in the vision feature at both the encoder and decoder stages. To maintain uniformity with GLIP, we plot only encoder fused features as its much closer to the parameter size of GLIP fusion transformers. 
Fused features `$\mathcal{F}$' correspond to neck features as $\mathcal{N}_1 \rightarrow\mathcal{F}_1$, 
$\mathcal{N}_2 \rightarrow\mathcal{F}_2$,
$\mathcal{N}_3 \rightarrow\mathcal{F}_3$,
$\mathcal{N}_4 \rightarrow\mathcal{F}_4$, and 
$\mathcal{N}_5 \rightarrow\mathcal{F}_5$.
For plotting, we use 
$\boldsymbol{\mathcal{B}_1, \mathcal{B}_2, \mathcal{B}_3, \mathcal{B}_4, \mathcal{N}_5, \& \mathcal{F}_5}$.

\begin{figure*}[!ht]
\centering
\includegraphics[width=\linewidth]{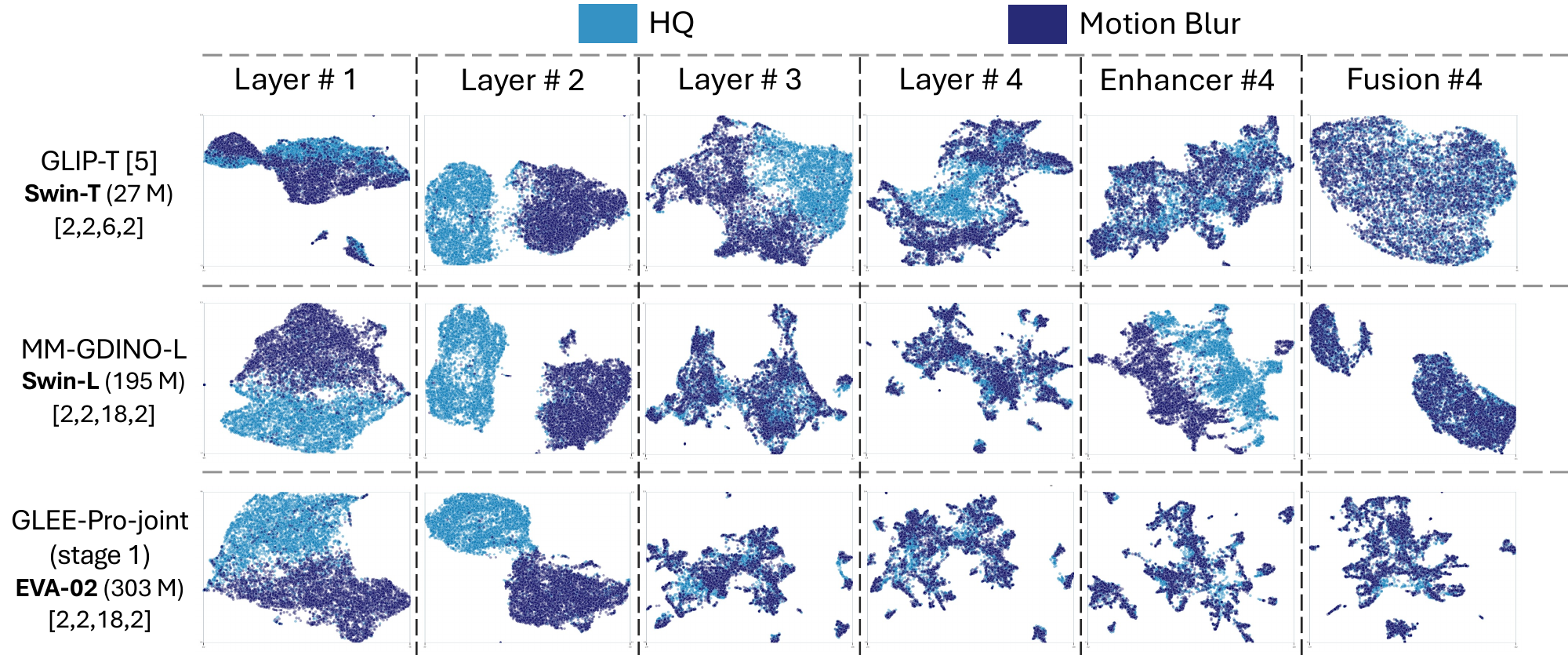}
\includegraphics[width=\linewidth]
{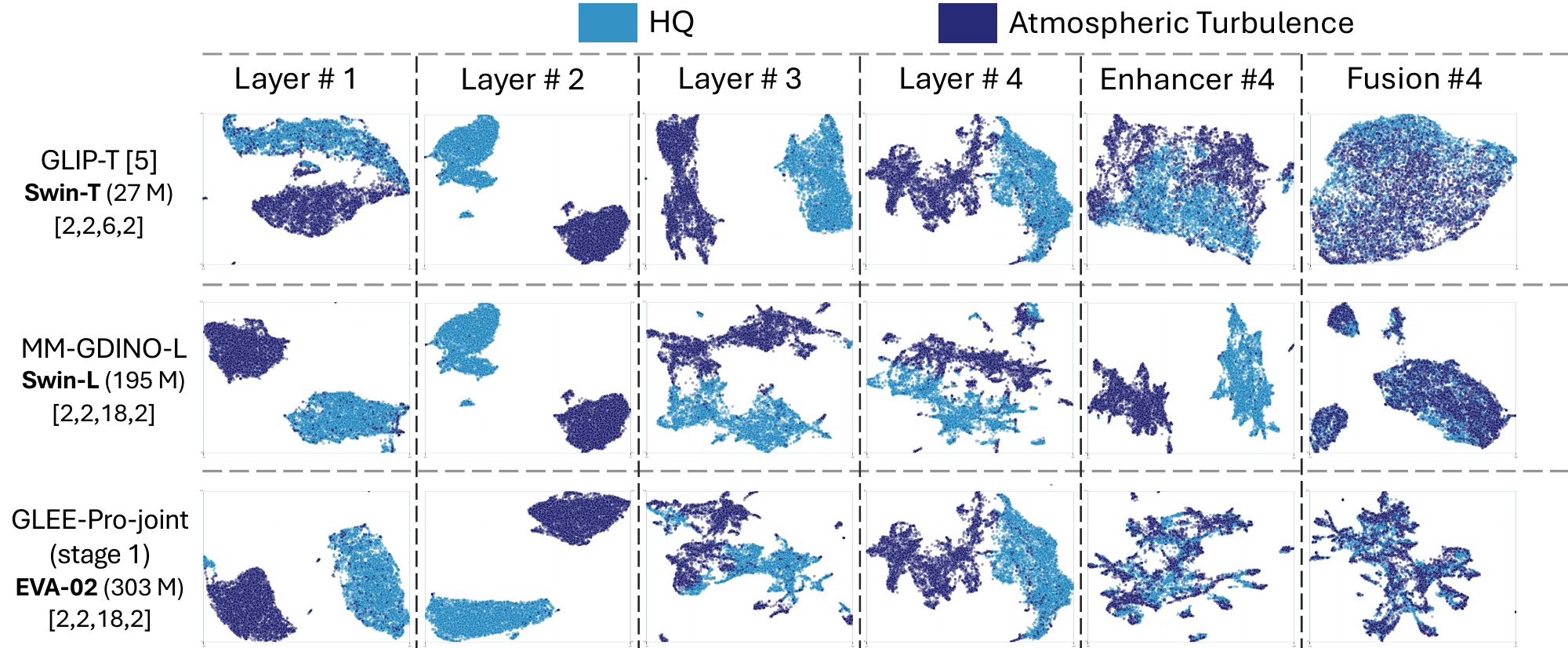}
\caption{
\textbf{Features UMAP:} 
Same details as those of Fig 8, for motion blur (above) and atmospheric turbulence (bottom), with noise implemented on COCO.  Only 1 severity. 
}
\label{fig:pipeline_flow_umap_noise}
\end{figure*}

\vspace{3pt}
For \textbf{GLEE}, model EVA-02 24 layers are partitioned similarly to MM-GIDNO ViT-Large as [2,2,18,2]. While the model only uses the last layer of the backbone feature $\mathcal{B}_4$, we have plotted intermediate features $\mathcal{B}_1$, $\mathcal{B}_2$, $\mathcal{B}_3$, and $\mathcal{B}_4$ for fair comparison.
Neck in this model is actually part of Spatial pyramid transformer structure with 
$
\mathcal{N}_1 = \mathcal{H}_1\large(\mathcal{C}_1 \large(\mathcal{B}_4\large)\large) 
\parallel
\mathcal{N}_2 = \mathcal{H}_2\large(\mathcal{C}_2 \large(\mathcal{B}_4\large)\large) 
\parallel
\mathcal{N}_3 = \mathcal{H}_3\large(\mathcal{C}_3 \large(\mathcal{B}_4\large)\large) 
\parallel
\mathcal{N}_4 = MaxPool \large(
\mathcal{H}_3\large(\mathcal{C}_3 \large(\mathcal{B}_4\large)\large) \large) 
$ 
creating 4 neck features, called `p3', `p4', `p5', and `p6', in the original code. Similar to MMGIDNO there is an encoder-decoder structure in fusion network, with "early fusion".
We only use encoder to show effect of fusion, generating  generating `$\mathcal{F}$' features as $\mathcal{F}_1$, $\mathcal{F}_2$, $\mathcal{F}_3$, $\mathcal{F}_4$ for $\mathcal{N}_1$, $\mathcal{N}_2$, $\mathcal{N}_3$, $\mathcal{N}_4$, respectively. 
For plotting, we use 
$\boldsymbol{\mathcal{B}_1, \mathcal{B}_2, \mathcal{B}_3, \mathcal{B}_4, \mathcal{N}_4, \& \mathcal{F}_4}$.

\begin{figure*}[!ht]
\centering
\includegraphics[width=\linewidth]{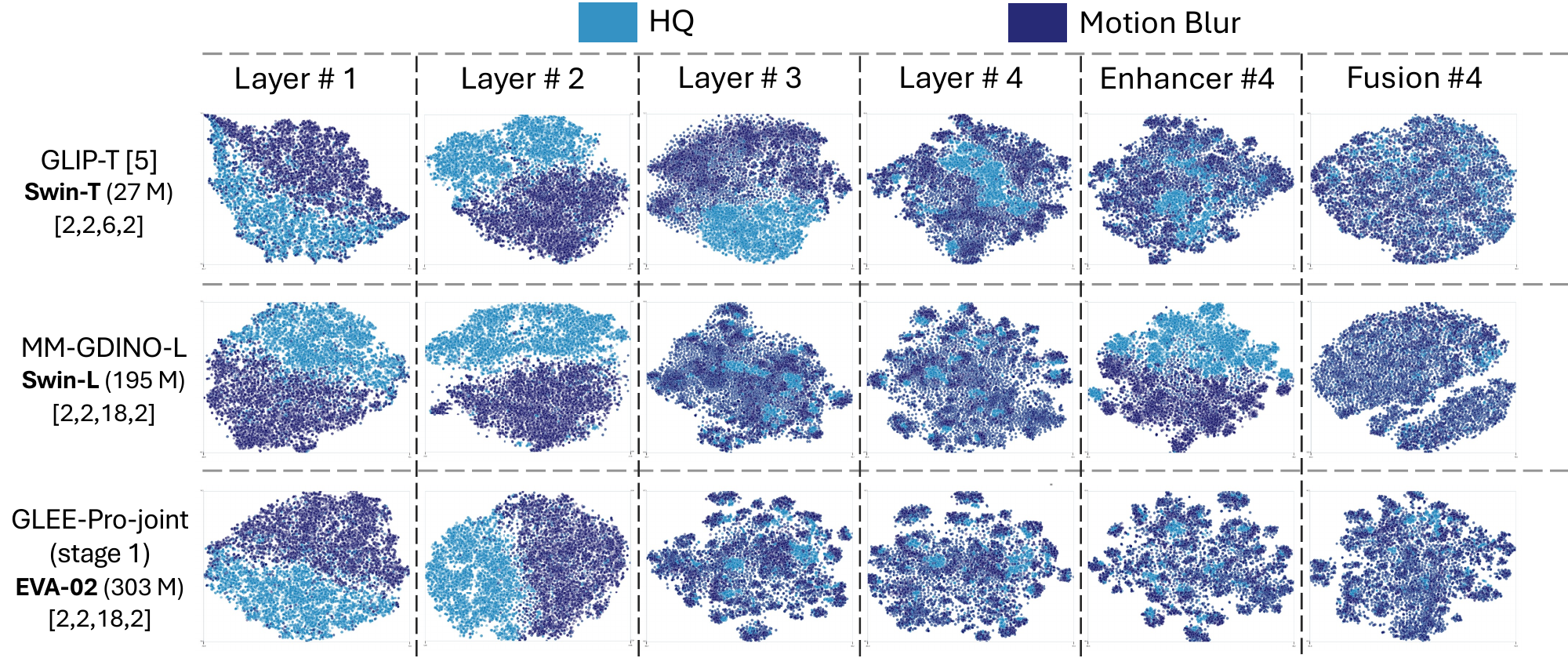}
\includegraphics[width=\linewidth]
{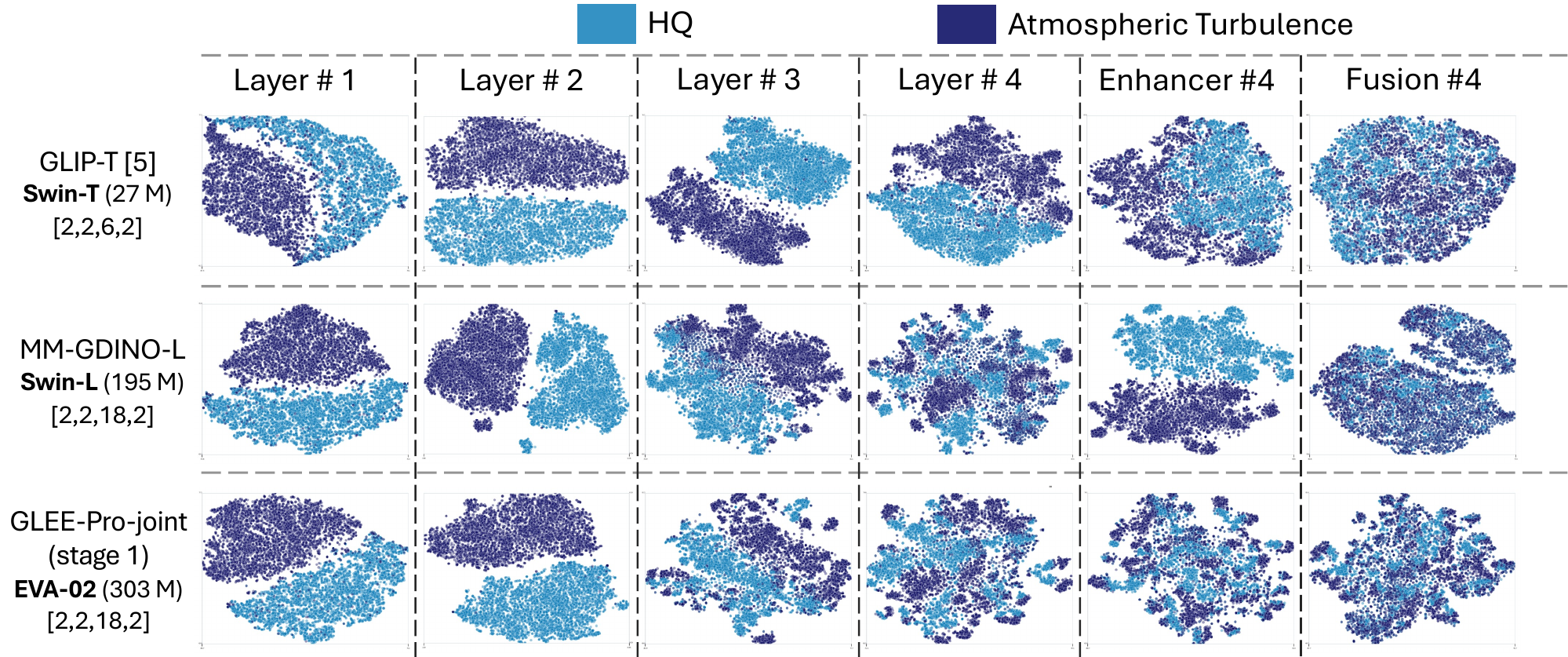}
\caption{
\textbf{Features t-SNE:} 
Same details as those of Fig 8, for motion blur (above) and atmospheric turbulence (bottom), with noise implemented on COCO, only 1 severity. 
}
\label{fig:pipeline_flow_tsne_noise}
\end{figure*}

\subsection{Fig 9(a)}
\begin{figure*}[!h]
\centering
\includegraphics[width=0.9\linewidth]{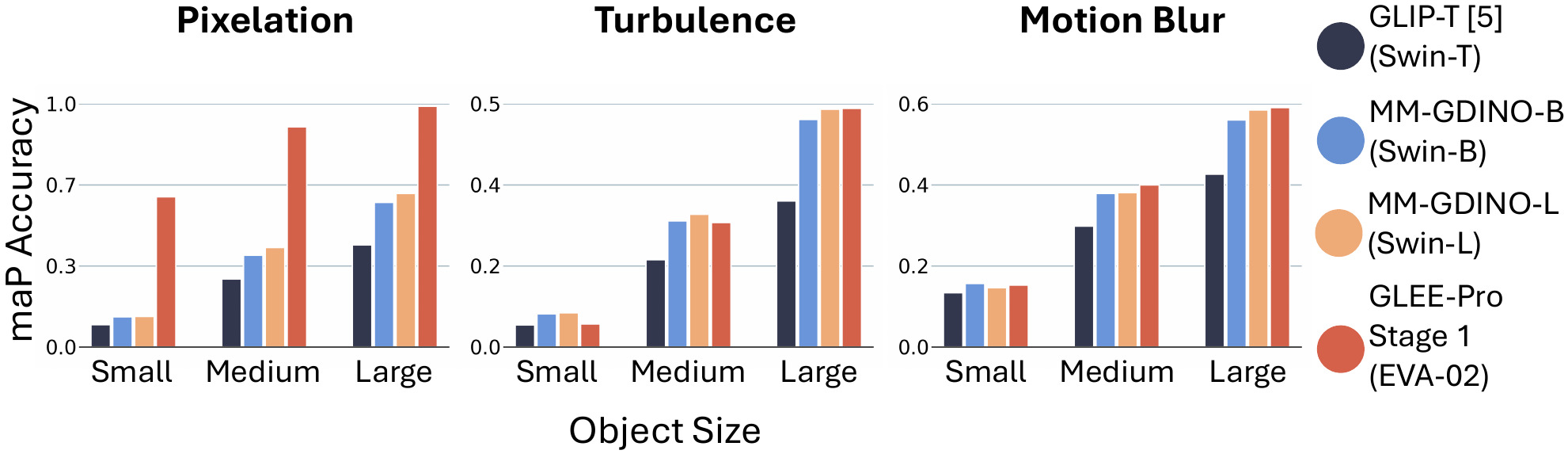}
\caption{
\textbf{Robustness vs Object Size}
All perturbations, extension of Fig 9(a).
}
\label{fig:object_size_noise}
\end{figure*}

\begin{figure*}[!h]
\centering
\includegraphics[width=0.6\linewidth]{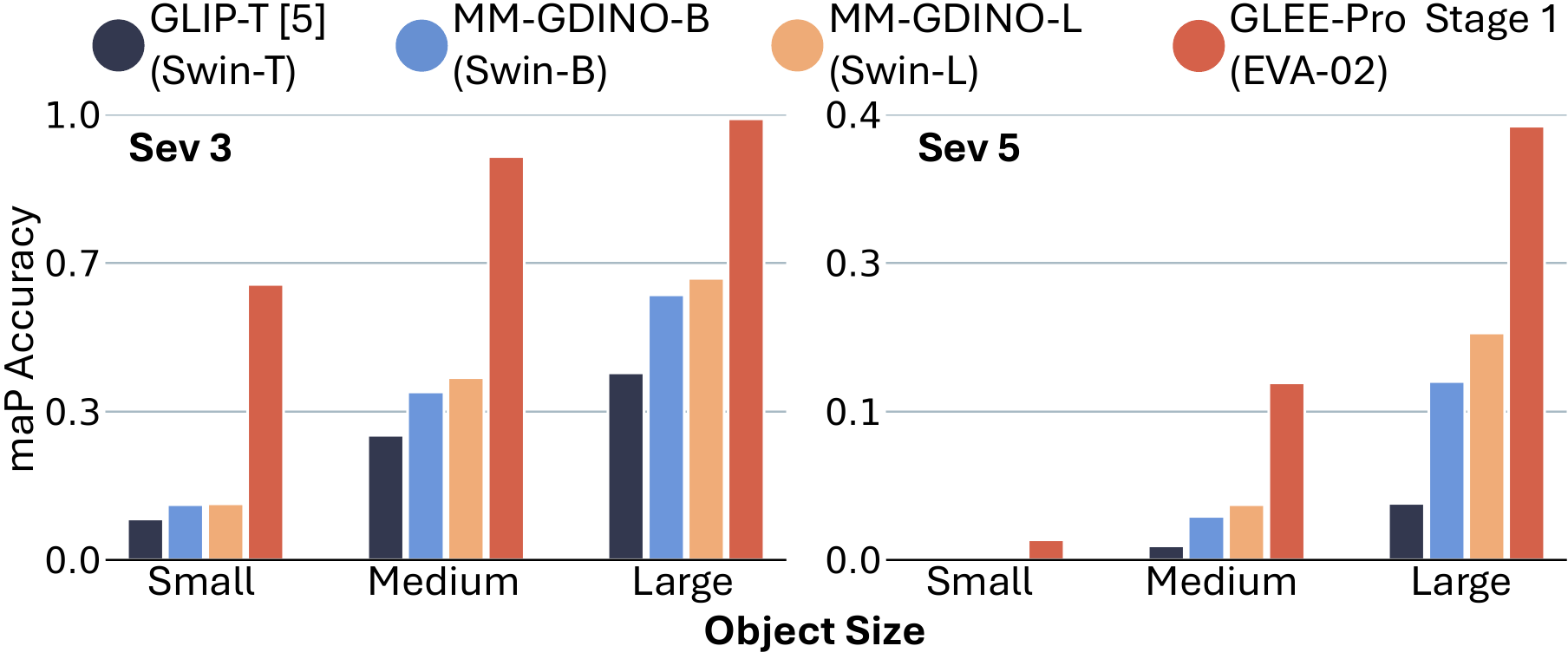}
\caption{
\textbf{Robustness vs Object Size}
Pixelation at sev 3 \& 5, extension of Fig 9(a). 
}
\label{fig:object_size_sev5}
\end{figure*}

 Fig 9(a) showed results for COCO at sev 3. Here we show results at sev 5 as well in \Cref{fig:object_size_sev5} and results for real world noises in \Cref{fig:object_size_noise}. The size of objects was determined by official annotation in COCO. The objects are categorized into three size bins: small, medium, and large, based on the area of their bounding boxes in $pixels^2$. These bins are defined as: small for areas in the range (0 , $32^2$], medium for ($32^2$, $96^2$], and large for ($96^2$, $(1e^5)^2$].

%% file: Supplementary/figures_9_all.tex
\subsection{Fig 9(b)}
\begin{figure*}[!h]
\centering
\includegraphics[width=0.8\linewidth]{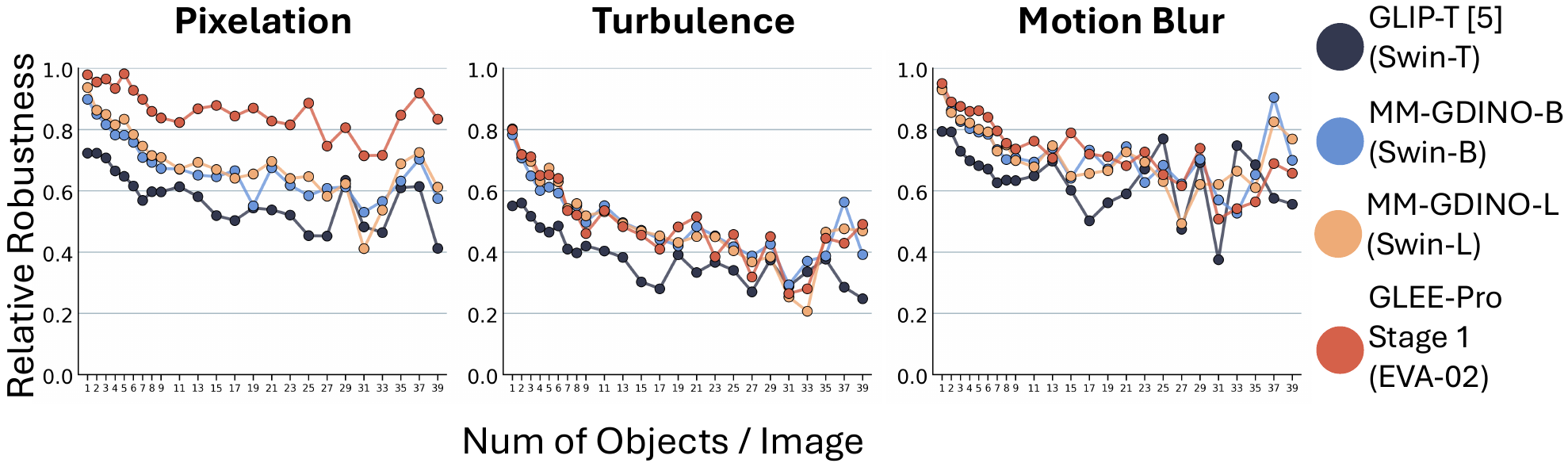}
\caption{
\textbf{Robustness vs num of objects/image for all noise perturbations}. 
Same details as those of Fig 9(b)
}
\label{fig:no_obj_noise}
\end{figure*}

\begin{figure*}[!h]
\centering
\includegraphics[width=0.6\linewidth]{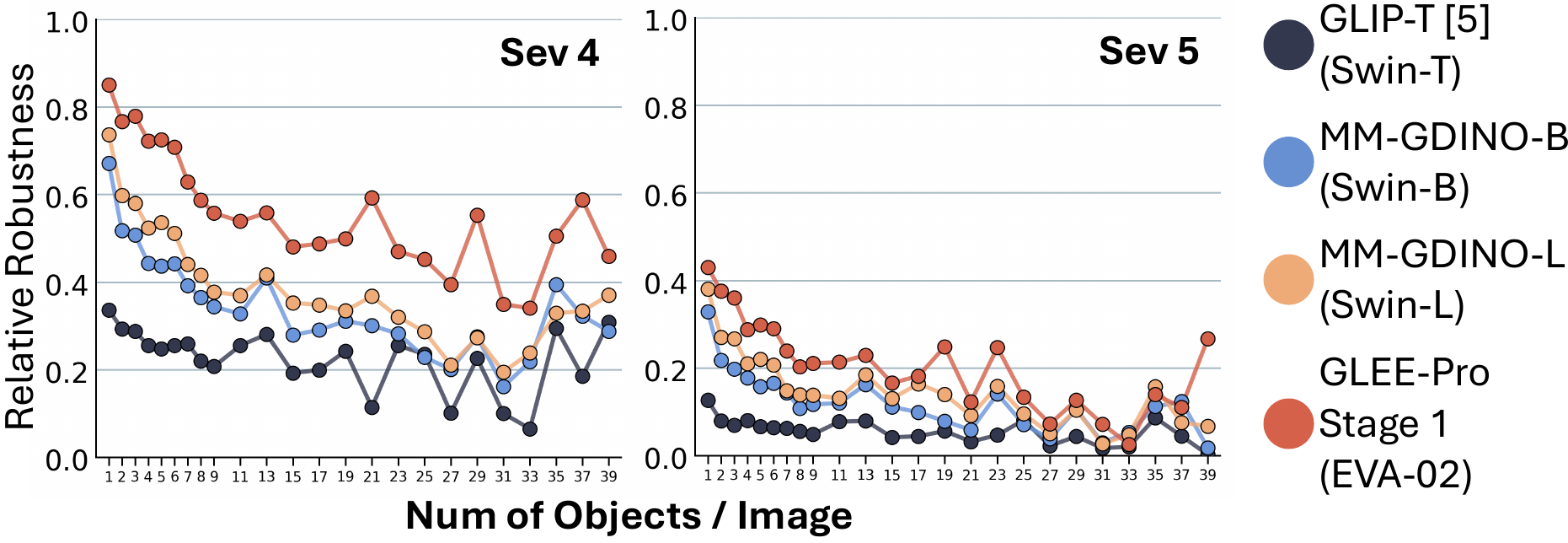}
\caption{
\textbf{Robustness vs num of objects/image at sev 4 and 5}. 
Same details as those of Fig 9(b) 
}
\label{fig:no_of_objects_sev4}
\end{figure*}

 Fig 9(b) showed results for COCO at sev 3. Here we show results at sev 4 and sev 5 in (\cref{fig:no_of_objects_sev4}) and results for real world noises in \Cref{fig:no_obj_noise}. The image was divided by the number of ground truth boxes per image. Buckets with number of images $>$ 20 were retained. After applying the filter, we got around 39 buckets. For each visualization, we have shown only the odd-number bucket after the 10th bucket.

\subsection{Fig 9(c)}
\begin{figure*}[!h]
\centering
\includegraphics[width=0.95\linewidth]{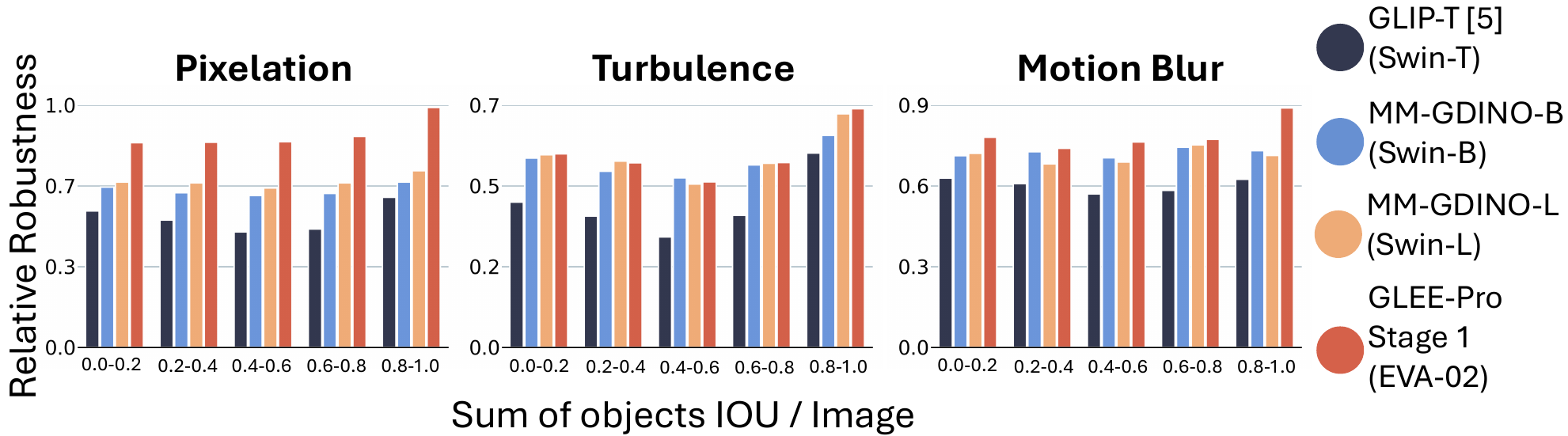}
\caption{
\textbf{Robustness vs occlusion} Real world perturbation on COCO, like Fig 9(c).
}
\label{fig:occlusion_noise}
\end{figure*}

\begin{figure*}[!h]
\centering
\includegraphics[width=0.75\linewidth]{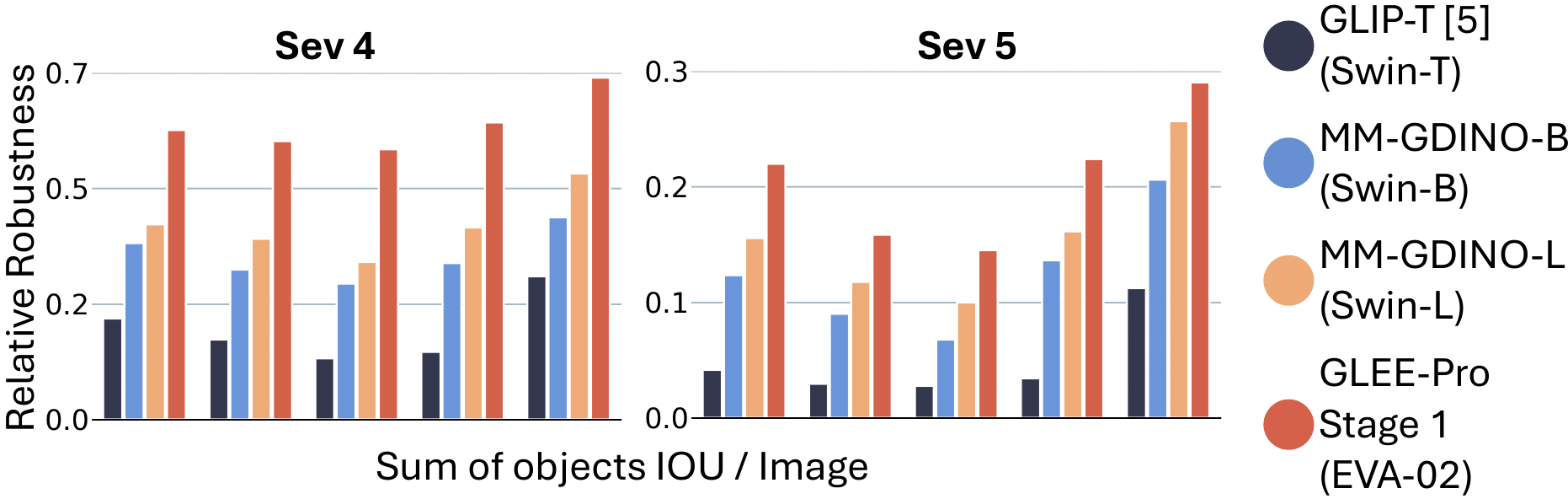}
\caption{
\textbf{Robustness vs occlusion} Pixelation for Sev 4 \& 5, like Fig 9(c) on COCO. 
}
\label{fig:occlusion_sev4}
\end{figure*}

 Fig 9(c) showed results for COCO at sev 3. Here we show results at sev 4 and sev 5 in \cref{fig:occlusion_sev4} and real world noises in \Cref{fig:occlusion_noise}. Images was divided by the summations of IOUS per image. The images are binned based upon various occlusion IOU ranges as mentioned in Fig 9(c). The normalized per-image occlusion IOU are computed as follows:
\begin{equation}
\text{IoU}_{\text{image}} = \frac{\sum\limits_{(i,j) \in \mathcal{O}} \text{IoU}(B_i, B_j)}{\left| \bigcup\limits_{(i,j) \in \mathcal{O}} B_i \cup B_j \right|}
\end{equation}
where, $\mathcal{O}$ is the set of all pairs of overlapping bounding boxes $(B_i, B_j)$. Further, the IOU bins with fewer than 50 images are removed to reduce the noise during the robustness evaluation process. 

 Buckets with number of images > 50 were kept. After applying the filter, we got 5 bins, which are shown on X-axis.

\subsection{Fig 10 (left)}

\begin{figure}[!h]
\centering
\begin{subfigure}[t]{0.47\textwidth}
\centering
\includegraphics[width=\linewidth]{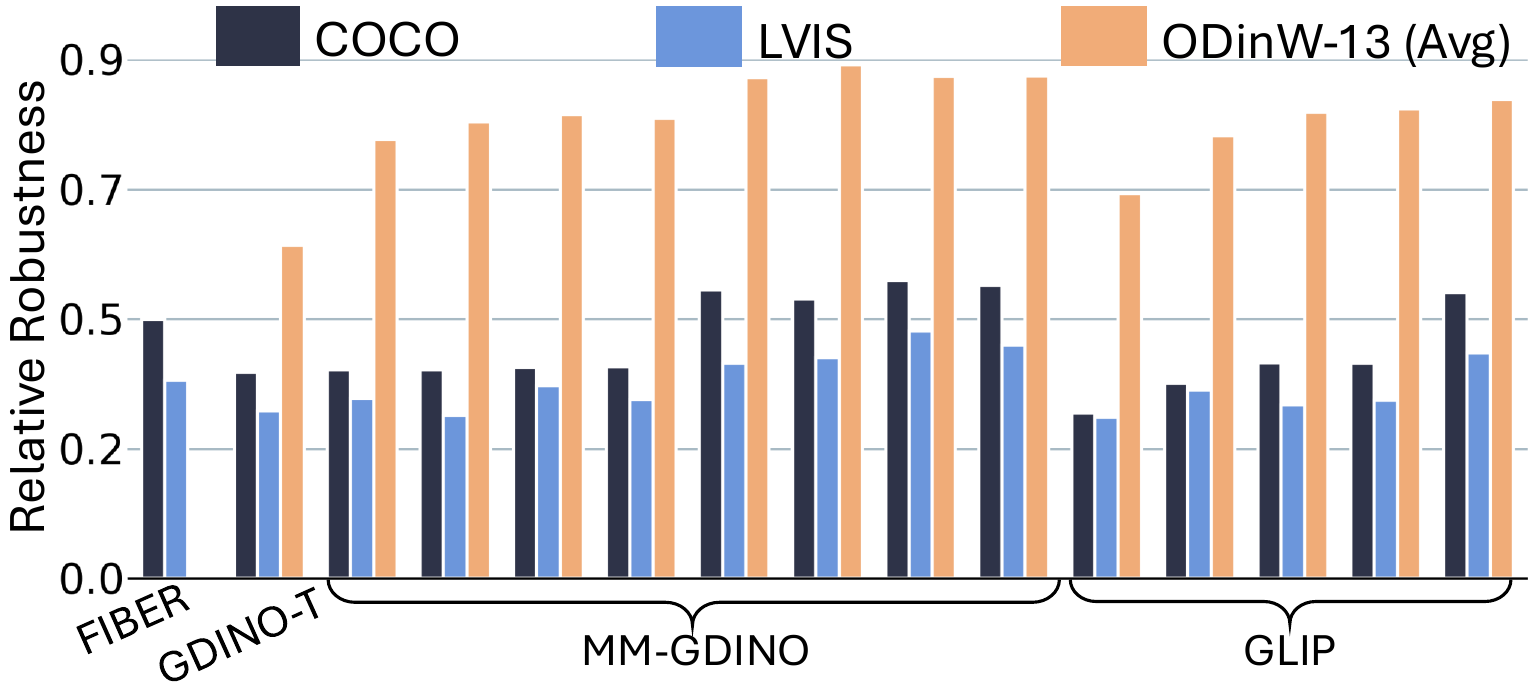}

\label{fig:robustness_data_turb}
\end{subfigure}
\hfill
\begin{subfigure}[t]{0.47\textwidth}
\centering
\includegraphics[width=\linewidth]{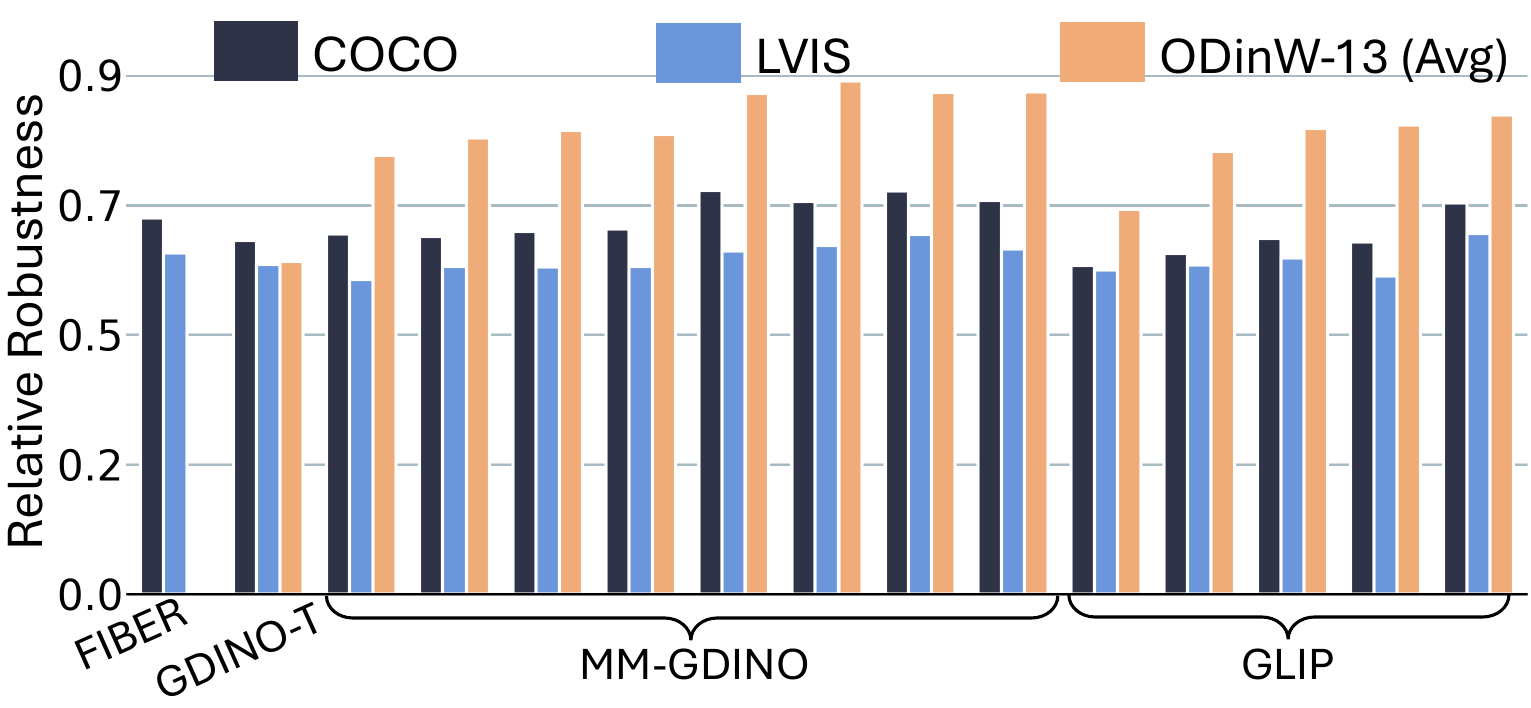}

\label{fig:robustness_data_motion}
\end{subfigure}
\caption{\textbf{Robustness of dataset under real world perturbations} \figno{(left)} Robustness of dataset under turbulence perturbation \figno{(right)} Robustness of dataset under motion blur perturbation. Same details as that of Fig 10 (Left)}

\label{fig:dataset_rob_realworld}
\end{figure}
Even with real world perturbations, the ODinW-13 maintains high robustness scores of $\simeq0.7$ across models. However, in motion blur the overall robustness is higher across models over all the datasets due to minimal perturbation effect of motion blur as shown in \Cref{fig:dataset_rob_realworld}.

\subsection{Fig 10 (right)}
\begin{figure*}[!ht]
\centering
\includegraphics[width=0.9\linewidth]{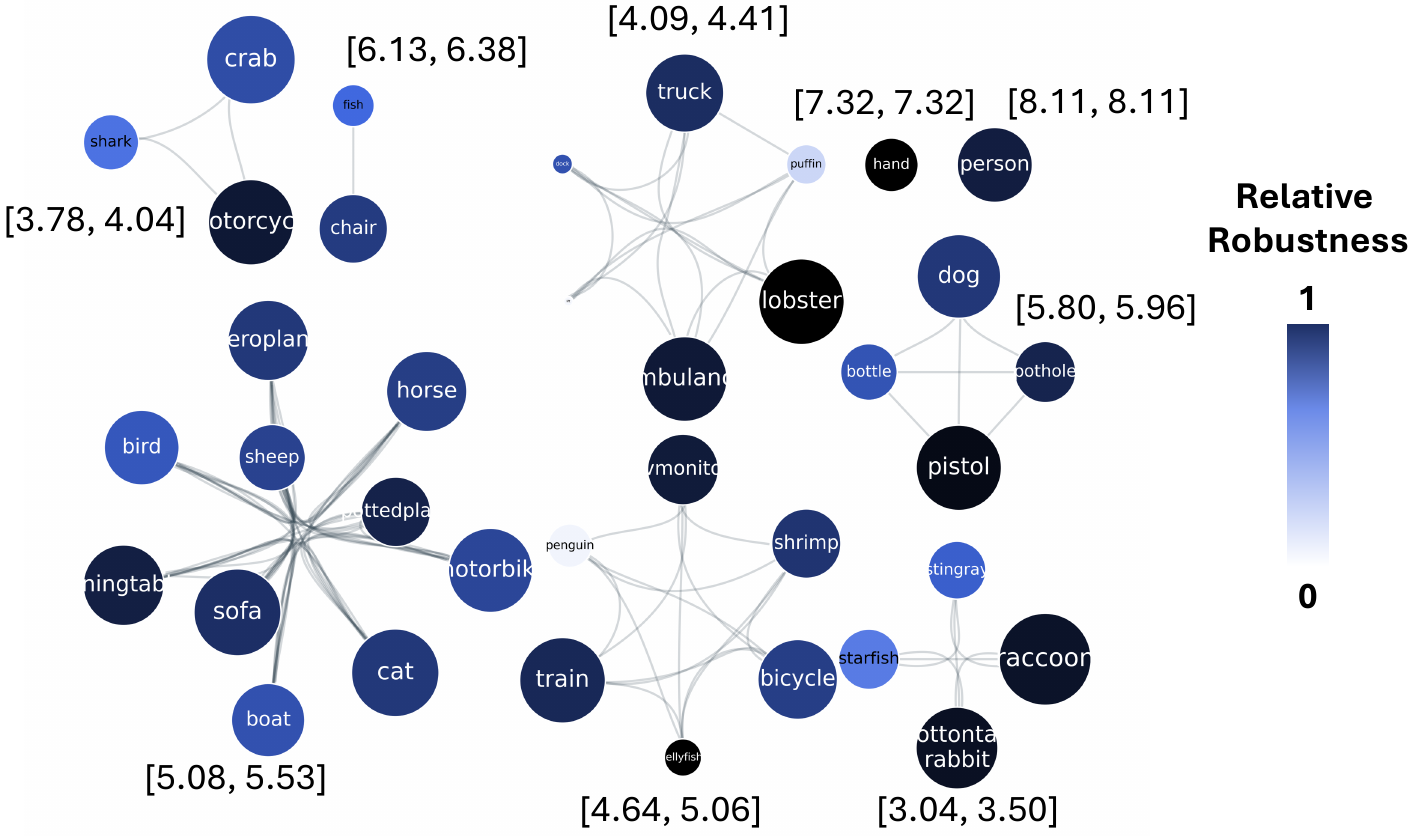}
\caption{
\textbf{Robustness vs categories for ODinW-13 at sev 3}. 
Extension of Fig 10 (right). 
}
\label{fig:no_of_objects_odinw}
\end{figure*}

 Fig 10 (right) showed results for COCO classes at sev 3. Here we consider robustness vs. class for ODinW-13 as shown in \Cref{fig:no_of_objects_odinw}. 
 For each category, we have three metrics: 1) accuracy of GLIP-T model, 2) average size of the objects (computed via mean IOU), 3) frequency of classes (\# of times certain classes appear). 
 We first cluster the categories based on the log of frequency of classes (the values indicated are the log range in each cluster). For COCO, we cluster with K=6, while for ODinW-13 clustering size is 10. 
 Color indicates the robustness, a darker shade indicates higher robustness. The size of nodes indicates the mean IOU. 
 We have only shown classes with \# of instances for that category $>$100 for COCO and $>$10 for ODinW-13.
Additionally, a lot of OdinW-13 classes overlap across classes, hence, we chose only the first occurrence of such classes.

\subsection{Fig 12}
 Fig 12 shows only 8 datasets out of 13 OdinW-13 datasets. This is because either fine-grained or superclass evaluation accuracy at sev 3 is so close to random prediction that it can't be reliably used to make any inference. Near random prediction models achieve abnormal robustness scores~\cite{pathak2025lrfm}.
The super class annotation was obtained from the official annotation of OdinW-13. The superclass for OdinW-13 datasets are as follows (super category written in brackets): 

AerialMaritimeDrone (movable-objects), Aquarium (creatures), CottontailRabbits (Cottontail-Rabbit), EgoHands (hands), NorthAmericaMushrooms (mushroom), Packages (packages), PascalVOC (VOC), pistols (Guns), pothole (potholes), Raccoon (raccoons), Shellfish (shellfish), thermalDogsAndPeople (dog-person), VehiclesOpenImages (vehicles) 

Some datasets don't have meaningful supercategories; hence, they were removed during evaluation. The datasets like AerialMaritimeDrone, PascalVOC, and Aquarium have supercategories that do not align with their class labels. Others, like Egohands, Packages, Raccoons, Pistols, and Cottontail Rabbits, have matching/similar supercategories and class labels because they have have only one class.

\begin{figure}[!h]
\centering
\begin{subfigure}[t]{0.47\textwidth}
\centering
\includegraphics[width=\linewidth]{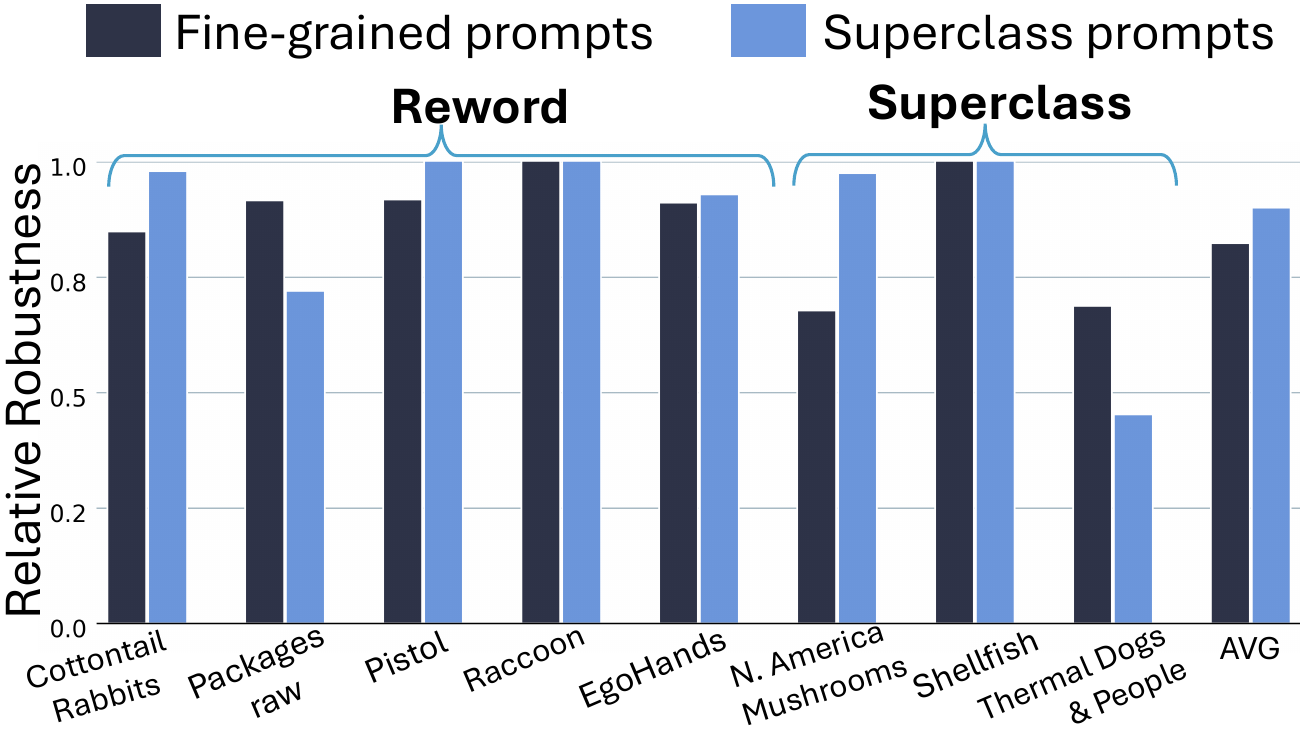}
\caption{
\textbf{Superclass vs finegrained on motion blur perturbation} 
}
\label{fig:superclass_motion}
\end{subfigure}
\hfill
\begin{subfigure}[t]{0.47\textwidth}
\centering
\includegraphics[width=\linewidth]{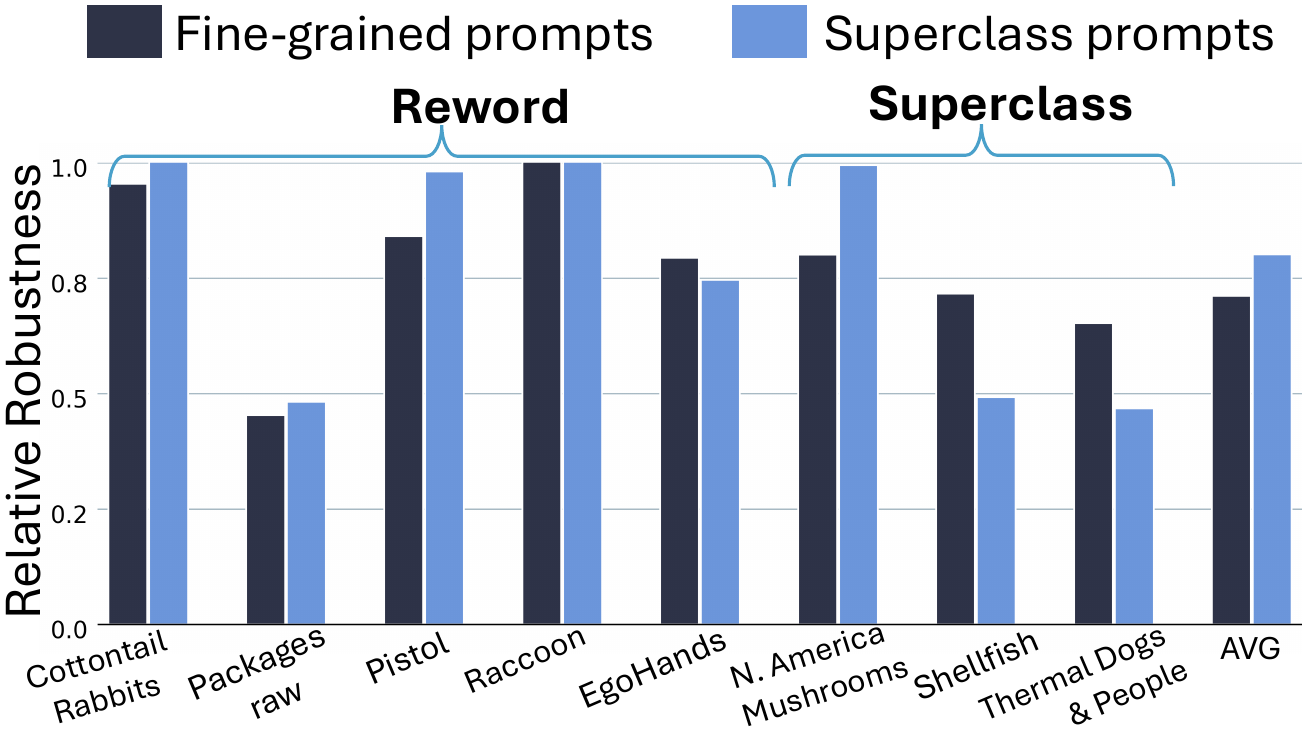}
\caption{
\textbf{Superclass vs finegrained on turbulence perturbation} 
}
\label{fig:superclass_turbulence}
\end{subfigure}
\caption{
\figno{\textit{(a)}} shows the superclass prompting performance with the motion blur perturbation \figno{\textit{(b)}} shows the superclass prompting performance with the turbulence perturbation. This follows the same trend as the pixelation perturbation, where the superclass/finegrained prompting doesn't vary the performance. 
}
\label{fig:superclass_realworld}
\end{figure}





%% file: Supplementary/dataset_analysis.tex
\section{Dataset Analysis}
\label{sec:dataset_analysis}

\subsection{COCO / LVIS vs ODinW-13}

This section presents a detailed comparison between the COCO / LVIS and ODinW-13 datasets, highlighting key differences between the datasets that might be contributing to  the difference between performance in the model.

\begin{figure}[ht]
\centering
\begin{subfigure}[b]{0.45\textwidth}
\includegraphics[width=\textwidth]{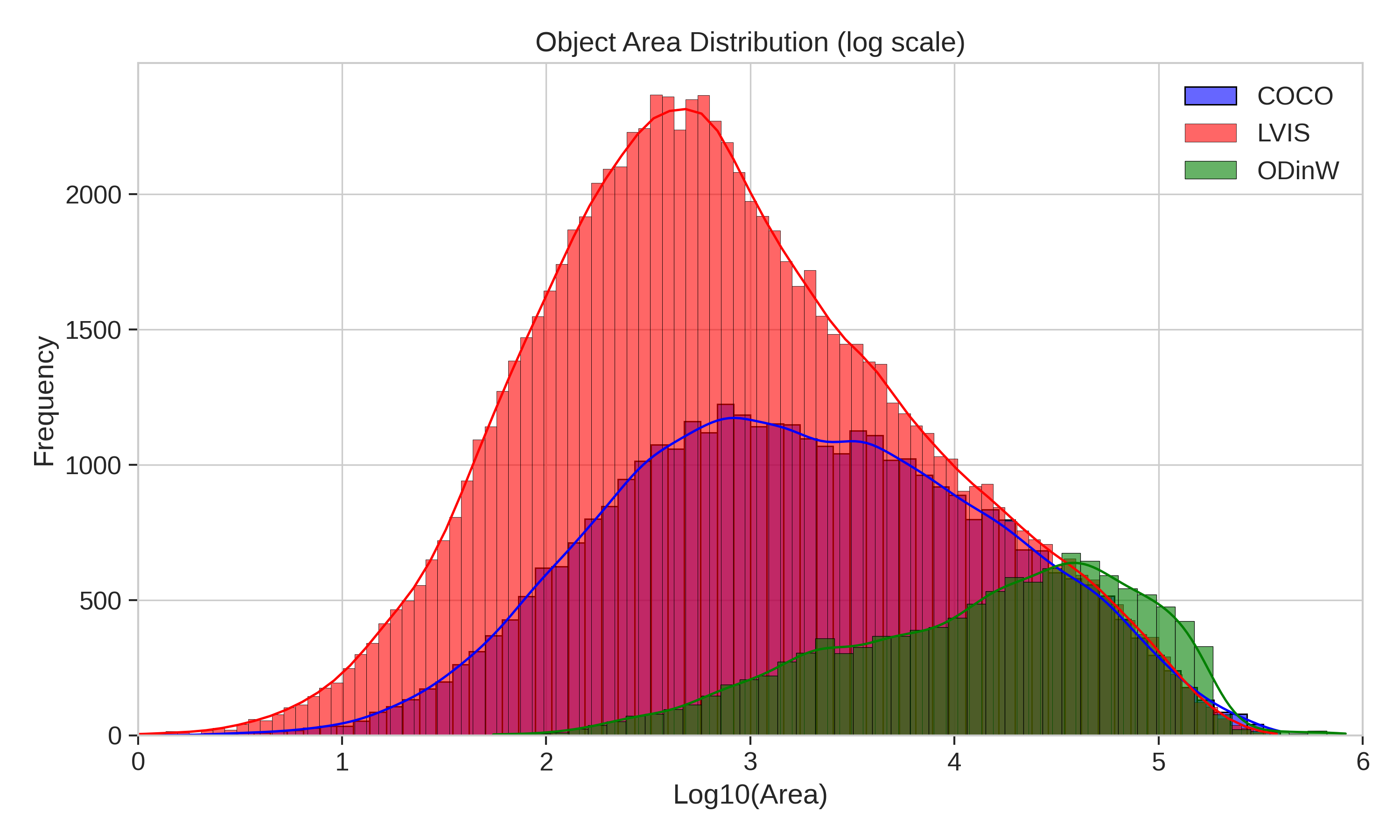}
\caption{\textbf{Object size distributions across dataset}}
\label{fig:size_distribution}
\end{subfigure}
\hfill
\begin{subfigure}[b]{0.45\textwidth}
\includegraphics[width=\textwidth]{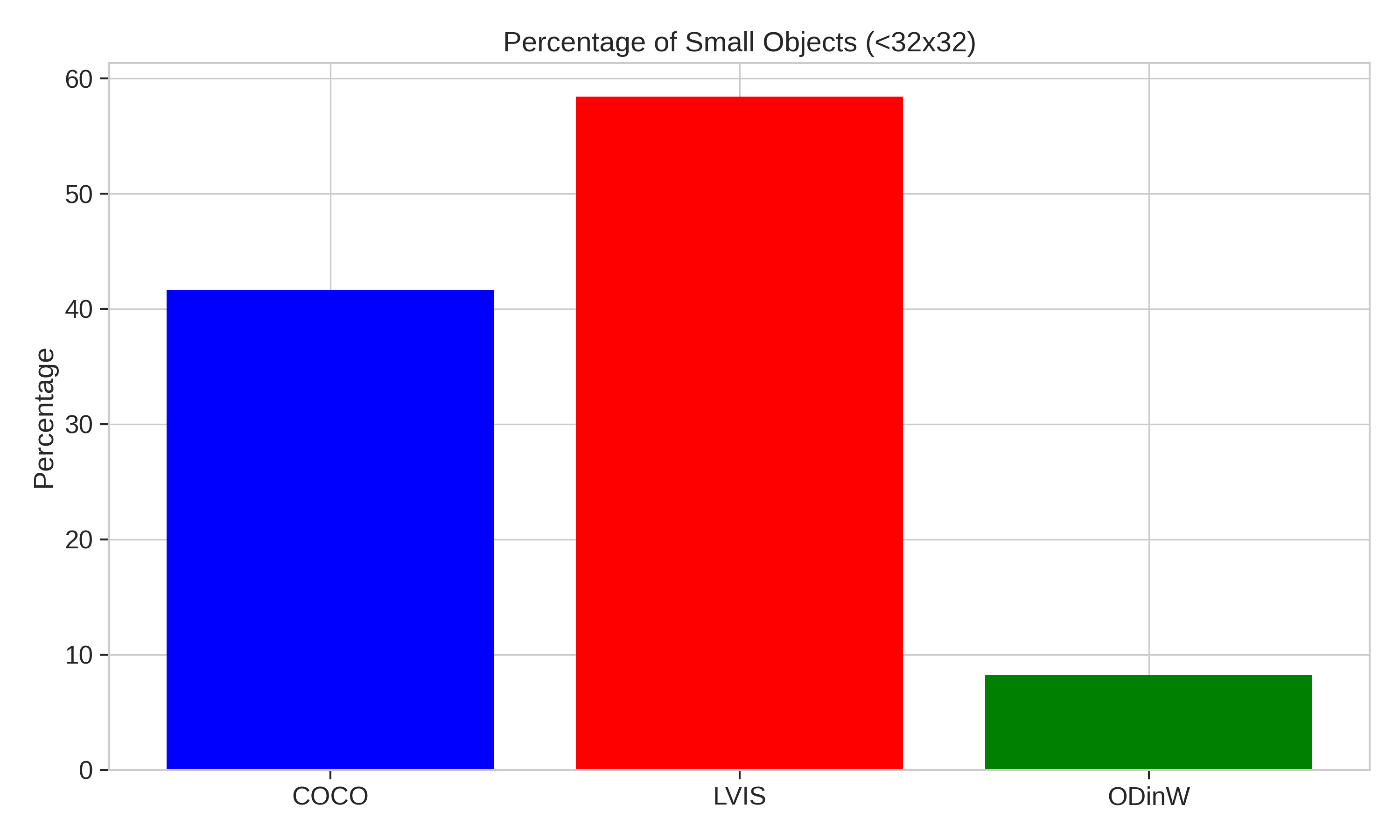}
\caption{\textbf{Small object size percentage across datasets}}
\label{fig:small_obj_dist}
\end{subfigure}
\hfill
\begin{subfigure}[b]{0.45\textwidth}
\includegraphics[width=\textwidth]{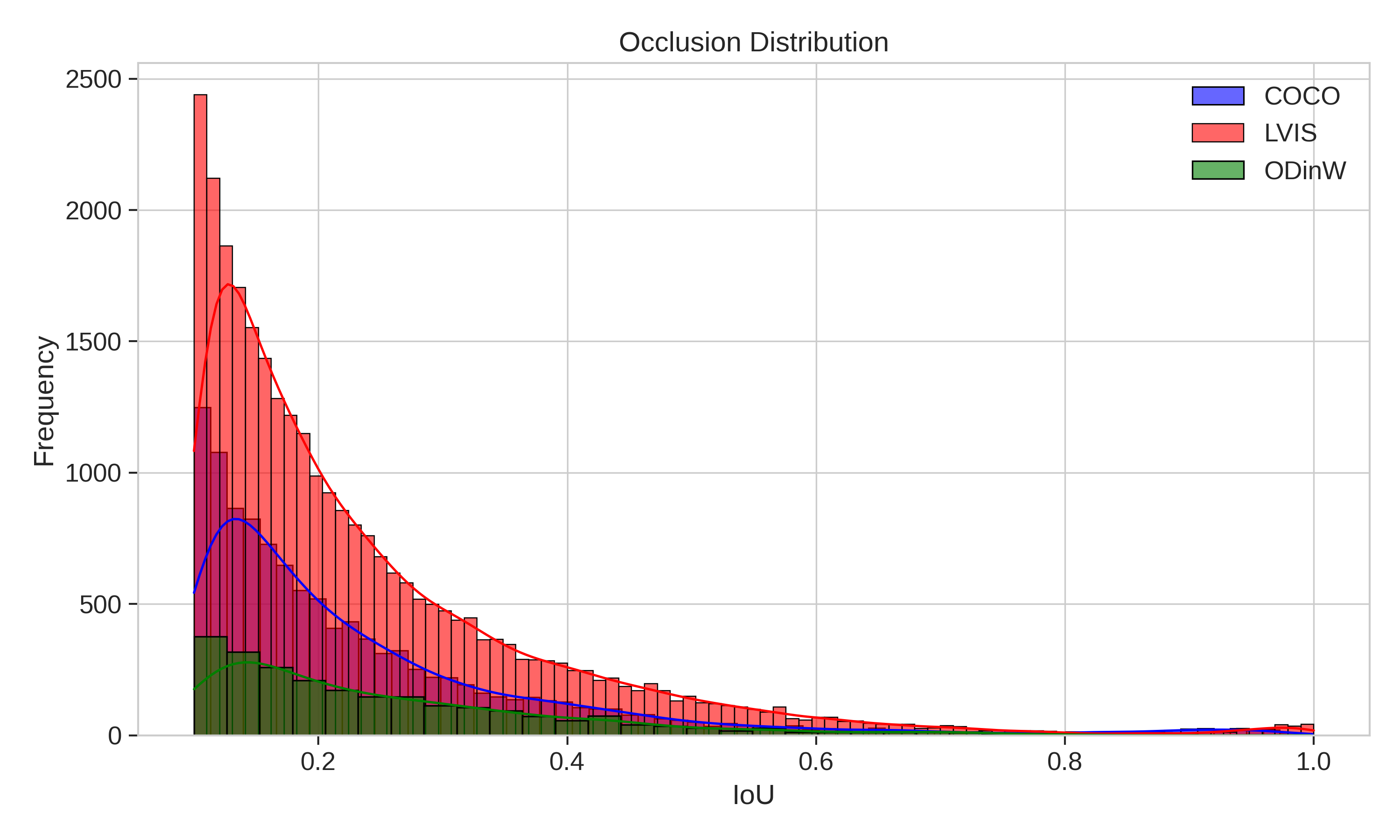}
\caption{\textbf{Occlusion rate of objects across datasets}}
\label{fig:occl_dist}
\end{subfigure}
\hfill
\begin{subfigure}[b]{0.45\textwidth}
\includegraphics[width=\textwidth]{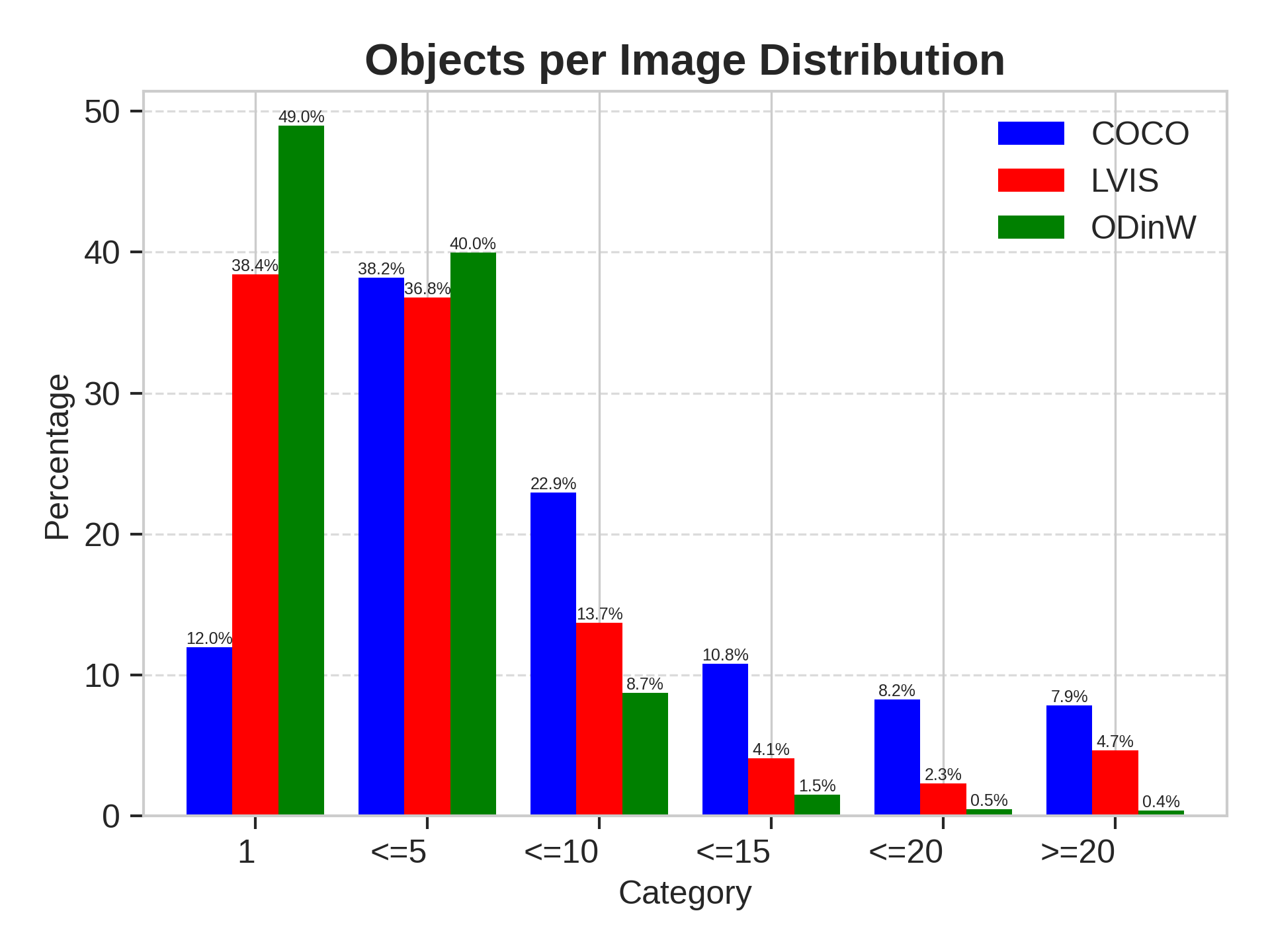}
\caption{\textbf{Number of Objects per image distribution}}
\label{fig:num_obj_dist}
\end{subfigure}
\caption{
\figno{(a)} LVIS shows a higher proportion of small objects compared to COCO, contributing to its greater vulnerability to resolution degradation. On the other hand ODinW-13 has much larger objects.
\figno{(b)} Number of small objects are more common in LVIS dataset and least common in ODinW-13 dataset.
\figno{(c)} Occlusion patterns reveals denser objects per image in LVIS, lowering the detection performance overall. 
\figno{(d)} Shows the distribution of number of objects per image across dataset. ODinW-13 has the least number of objects per image across all the images.
}
\label{fig:spatial_characteristics}
\end{figure}

\noindent \textbf{Object Size Distribution:}
 As shown in \cref{fig:small_obj_dist} \& \cref{fig:size_distribution} , LVIS has approximately 60\% of objects below $32^2$ pixels compared to 40\% in COCO. Small objects lose distinguishing features rapidly when we perturb the images.

\noindent \textbf{Spatial Characteristics:}
LVIS exhibits higher object density, occlusion rates (\cref{fig:occl_dist} 
\& \cref{fig:num_obj_dist}), and boundary complexity, exacerbating feature ambiguity at lower resolutions.

%% file: Supplementary/proposed.tex
\section{NN \& TKO (Sec 5.1 Validating Model Design) }
\label{sec:validation_viz}
\begin{figure*}[!h]
\centering
\includegraphics[width=0.8\linewidth]{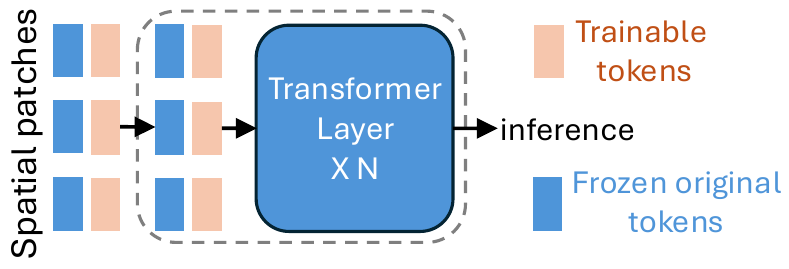}
\captionof{figure}{\textbf{TKO} \orangetext{Trainable prompts} added at every \figno{frozen layer} of transformer.}
\label{fig:lrtk0}
\end{figure*}

\begin{figure*}[!h]
\centering
\includegraphics[width=\linewidth]{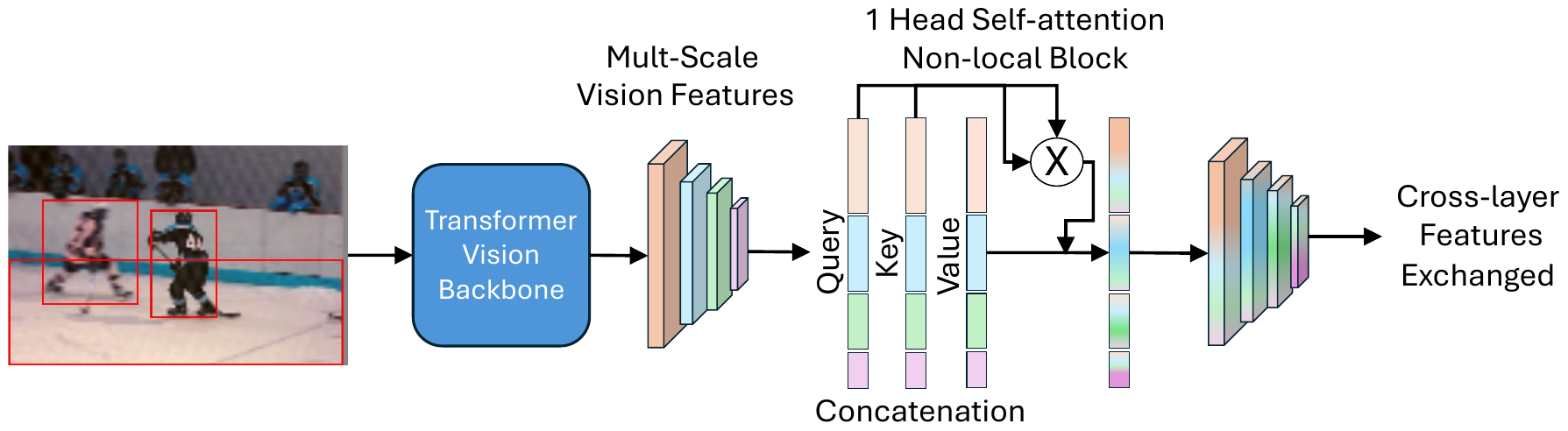}
\captionof{figure}{\textbf{NN:} Non-local block or 1-headed self-attention cross exchanges features across layers, making every layer aware of one another. Different layers (shallow and deeper) are shown in different colors. }
\label{fig:nn}
\end{figure*}

Visual Detection of TK0 and NN implementation in Sec. 5.1 shown in \cref{fig:lrtk0}, and \cref{fig:nn}

%% file: Supplementary/ethical_implications.tex
\section{Ethical Considerations \& Limitation}
\label{sec:ethical}

The diagnostic nature of our study does not obviate its ethical ramifications. We summarize the principal concerns and corresponding mitigation strategies.

\textbf{Dual use and surveillance amplification.}
While valuable for public-safety tasks (e.g., disaster response, wildlife monitoring), the improved robustness of deep learning models reduces technical barrier to pervasive or covert tracking like security cameras (CCTV), remote sensing, or mobile and aerial surveillance, reducing the 
Practitioners should adopt privacy-preserving measures and obtain explicit consent before deployment.



\textbf{Environmental footprint.}
We demonstrate that larger transformer backbones (e.g., EVA-02) confer superior robustness. However, training and inference at this scale incur substantial energy and carbon costs. We argue that future work should explore parameter-efficient robustness techniques, such as our TK0 and NN approach, which improve robustness with $~96\times$ fewer parameters.

\textbf{Limitations and future safeguards.}
Our analysis incorporates various types of synthetic noise, such as pixelation, turbulence, and Motion Blur. However, validating our analysis in a noise-agnostic setting prevents us from significantly improving robustness. For example, Foggy CitiScape would substantially benefit from fog-based pretraining. 

For the \textbf{language-based analysis}, retraining detectors with noise-aware captions e.g., \textit{“car on 
a foggy road”}, would help solidify our findings of language playing a minimal role in robustness. However, 
such noise-based caption datasets do not exist in research.
For \textbf{fine-tuned detectors analysis}, we dont really know if they are able to detect an object because they are robust
to noise or just familiar with distribution; e.g. \textit{“detecting a cat in snow”} vs. \textit{“seen cat images”}.


We advocate an expanded robustness-and-ethics benchmark that integrates fairness diagnostics, privacy-leakage assays, and real footage, all collected under informed consent.
By foregrounding these issues, we aim to e sdnsure that advances in robust zero-shot detection progress hand-in-hand with proactive mitigation of societal risks.